\documentclass{article}

\usepackage{arxiv}

\usepackage[utf8]{inputenc} 
\usepackage[T1]{fontenc}    
\usepackage{hyperref}       
\usepackage{url}            
\usepackage{booktabs}       
\usepackage{amsfonts}       
\usepackage{nicefrac}       
\usepackage{microtype}      
\usepackage{xcolor}         
\usepackage{multirow}
\usepackage{graphicx} 
\usepackage{float}
\usepackage{tabularray}
\usepackage{array}
\usepackage{amsmath}
\usepackage[numbers]{natbib}
\title{GaussianFusionOcc: A Seamless Sensor Fusion Approach for 3D Occupancy Prediction Using 3D Gaussians}

\author{
Tomislav Pavković\\
Technical University of Munich\\
BMW Group\\
\texttt{Tomislav.Pavkovic@tum.de}
\And
Mohammad-Ali Nikouei Mahani\\
BMW Group\\
\texttt{Mohammad-Ali.Nikouei-Mahani@bmw.de}
\And
Johannes Niedermayer\\
BMW Group\\
\texttt{Johannes.Niedermayer@bmw.de}
\And
Johannes Betz\\
Technical University of Munich\\
Munich Institute of Robotics and Machine Intelligence\\
\texttt{Johannes.Betz@tum.de}
}

\begin{document}
\maketitle
\begin{figure}[h]
    \centering
    \includegraphics[width=\textwidth]{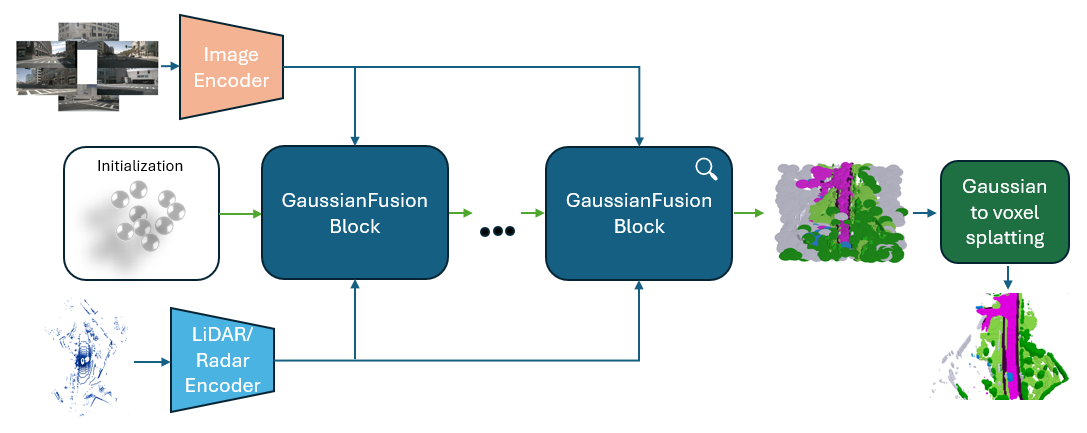}
    \caption{\textbf{GaussianFusionOcc pipeline:} Our method utilizes modality-specific encoders to extract feature maps from the input data. Extracted features are then fed to GaussianFusion blocks along with Gaussians and queries from the previous block. The GaussianFusion block, visualized in Figure \ref{fig:gaussianfusionblock}, extracts per-Gaussian features from each modality, fuses them into a unified feature vector, which is then used to refine the input Gaussians. Resulting Gaussians from the last block are splatted to a voxelized representation using the Gaussian-to-voxel splatting module.}
    \label{fig:gfoccpipeline}
\end{figure}

\begin{abstract}
3D semantic occupancy prediction is one of the crucial tasks of autonomous driving. It enables precise and safe interpretation and navigation in complex environments. Reliable predictions rely on effective sensor fusion, as different modalities can contain complementary information. Unlike conventional methods that depend on dense grid representations, our approach, GaussianFusionOcc, uses semantic 3D Gaussians alongside an innovative sensor fusion mechanism. Seamless integration of data from camera, LiDAR, and radar sensors enables more precise and scalable occupancy prediction, while 3D Gaussian representation significantly improves memory efficiency and inference speed. GaussianFusionOcc employs modality-agnostic deformable attention to extract essential features from each sensor type, which are then used to refine Gaussian properties, resulting in a more accurate representation of the environment. Extensive testing with various sensor combinations demonstrates the versatility of our approach. By leveraging the robustness of multi-modal fusion and the efficiency of Gaussian representation, GaussianFusionOcc outperforms current state-of-the-art models.
\end{abstract}

\section{Introduction}
\label{sec:intro}

Accurate 3D semantic occupancy prediction is a foundational task for safe autonomous navigation. Reliable perception of the surrounding environment enables precise situation3al awareness necessary for informed decision-making, efficient route planning, and collision avoidance \citep{ming2024occfusion, lu2024octreeocc, caesar2020nuscenes, wang2023openoccupancy}. While recent advances in voxel-based methods \citep{li2023voxformer, cao2022monoscene, jiang2024symphonize, miao2023occdepth, wei2023surroundocc, zhang2023occformer} have pushed performance on benchmarks like nuScenes \citep{caesar2020nuscenes, wei2023surroundocc, wang2023openoccupancy}, their reliance on dense volumetric grids creates prohibitive computational and memory costs \citep{huang2023tri, huang2024gaussianformer, huang2024probabilistic}. Furthermore, these methods struggle with the inherent sparsity of relevant information in driving scenes, where most of the volume remains empty or unchanging.
Concurrently, sensor fusion remains critical for robustness and reliability in dynamic environments, as cameras, LiDAR, and radar provide complementary strengths: cameras capture rich semantics, LiDAR offers precise geometry, and radar ensures reliability in adverse conditions. However, existing fusion frameworks are constrained by their underlying representations, which lack adaptability to scene complexity and scalability across modalities \cite{liu2023bevfusion, ming2024occfusion, wang2023openoccupancy}.

Emerging 3D Gaussian splatting techniques \citep{kerbl3Dgaussians, huang2024gaussianformer, huang2024probabilistic, yang2024deformable} offer a promising alternative to voxels by modeling scenes with anisotropic, learnable Gaussians, enabling faster rendering and lower memory usage. The inherent sparsity and adaptability of Gaussian representations naturally align with the characteristics of driving scenes, where information density varies across the environment \citep{huang2024gaussianformer}. By concentrating representational capacity where it matters most, Gaussian-based methods can achieve superior memory efficiency compared to voxel grids.

We present GaussianFusionOcc, a novel framework that extends the capabilities of Gaussian-based 3D scene representations to the multi-modal setting. While existing Gaussian-based approaches \citep{huang2024gaussianformer, huang2024probabilistic} have been limited to single-modality inputs, our approach introduces a modality-agnostic Gaussian encoder, capable of extracting per-Gaussian features from any individual sensor modality, such as camera, LiDAR, or radar, using a deformable attention mechanism. This attention-based approach enables the encoder to focus computational resources on the most informative regions of each sensor’s feature map, ensuring robust performance even when certain sensors are impaired or provide conflicting information. Furthermore, we propose a fusion method to construct unified Gaussian feature vectors, based on the extracted features for each used modality. This unified representation allows the model to adaptively allocate Gaussians according to scene complexity, concentrating modeling capacity in regions with intricate geometry or semantic boundaries while maintaining sparsity elsewhere. 

The main contributions of this work are the following:

We introduce the first framework that leverages 3D Gaussian splatting for multi-modal 3D semantic occupancy prediction.

We propose a modality-agnostic Gaussian encoder based on a deformable attention mechanism that effectively extracts per-Gaussian features from diverse sensor modalities, and a fusion method that creates a unified representation based on the extracted features.

We demonstrate state-of-the-art performance on the nuScenes dataset, especially on the rainy and nighttime subsets, achieving superior performance accuracy while reducing memory requirements compared to leading voxel-based approaches.

We tested the model with various sensor combinations under different scenarios to demonstrate the performance gains.

\section{Related work} \label{sec:related_work}
Earlier 3D semantic occupancy prediction approaches \citep{li2023voxformer, cao2022monoscene, jiang2024symphonize, miao2023occdepth, wei2023surroundocc, zhang2023occformer} mostly relied on dense voxel grids to represent the 3D scene. While effective in capturing fine-grained details, these dense grid-based methods often suffer from high computational and memory overhead due to the inherent sparsity of real-world environments and the need for high-resolution grids. To mitigate these limitations, recent approaches \cite{li2024bevformer, huang2023tri, li2023fb} explored alternative scene representations like Bird's-Eye-View (BEV) and Tri-Perspective View (TPV), achieving strong performance with improved efficiency. However, these planar representations often involve a compression of 3D information, potentially leading to a loss of fine-grained geometric details necessary for accurate 3D occupancy prediction.

In pursuit of more efficient and scalable 3D scene representations, object-centric approaches \citep{huang2024gaussianformer, huang2024probabilistic, gan2024gaussianoccfullyselfsupervisedefficient, yang2024deformable, lu2024octreeocc, tang2024sparseocc, wang2024opus} have emerged as a promising alternative to dense voxel grids. These methods aim to represent the scene using a collection of primitives centered around objects or regions of interest, thus avoiding the computational redundancy associated with empty voxels. Among these, 3D Semantic Gaussians have recently gained traction as a flexible and efficient representation for 3D scenes \citep{huang2024gaussianformer, huang2024probabilistic, gan2024gaussianoccfullyselfsupervisedefficient, yang2024deformable}. Each Gaussian primitive can capture local geometric and semantic information, allowing for a sparse yet detailed representation of the environment. GaussianFormer \citep{huang2024gaussianformer} introduced an object-centric approach using 3D semantic Gaussians for vision-based 3D semantic occupancy prediction, demonstrating comparable performance to state-of-the-art methods with significantly reduced memory consumption. By representing the scene with a set of learnable 3D Gaussians and employing efficient Gaussian-to-voxel splatting for occupancy prediction, GaussianFormer showcases the potential of object-centric representations for achieving both accuracy and efficiency. Probabilistic extensions of this idea, such as the probabilistic Gaussian superposition model \citep{huang2024probabilistic}, further aim to improve the utilization and efficiency of 3D Gaussian representations.

Recognizing the limitations of relying solely on a single sensor, multi-sensor fusion has become a critical direction for robust 3D occupancy prediction. Integrating LiDAR data with camera images has been shown to improve depth estimation and overall perception accuracy. BEVFusion \citep{liu2023bevfusion} proposed fusing LiDAR and camera features in the BEV space for multi-task perception. SparseFusion \citep{xie2023sparsefusion} further refined the feature fusion module for improved efficiency. While these methods primarily focus on 3D object detection \citep{liu2023bevfusion, xie2023sparsefusion, chen2023futr3d}, there's a growing interest in multi-sensor fusion for 3D semantic occupancy prediction. Camera-radar fusion has also been explored for tasks like object detection and tracking \citep{kim2023craft, nabati2021centerfusion, chen2023futr3d}, but dedicated approaches for 3D semantic occupancy prediction are scarce due to the sparsity of radar data. To address the need for robust occupancy prediction, OccFusion \citep{ming2024occfusion} was introduced as a novel framework to integrate features from surround-view cameras, radars, and 360-degree LiDAR through dynamic fusion modules, demonstrating superior performance in challenging conditions. These efforts highlight the benefits of combining complementary sensor data to achieve more reliable and accurate 3D scene understanding for autonomous driving. Building upon these advancements, our GaussianFusionOcc leverages the efficiency and flexibility of 3D semantic Gaussians while introducing a seamless sensor fusion mechanism to harness the complementary strengths of camera, LiDAR, and radar data for robust and accurate 3D occupancy prediction.

\section{Methods}
\label{sec:dataset}

\begin{figure*}[t]
\centering
    \includegraphics[width=\textwidth]{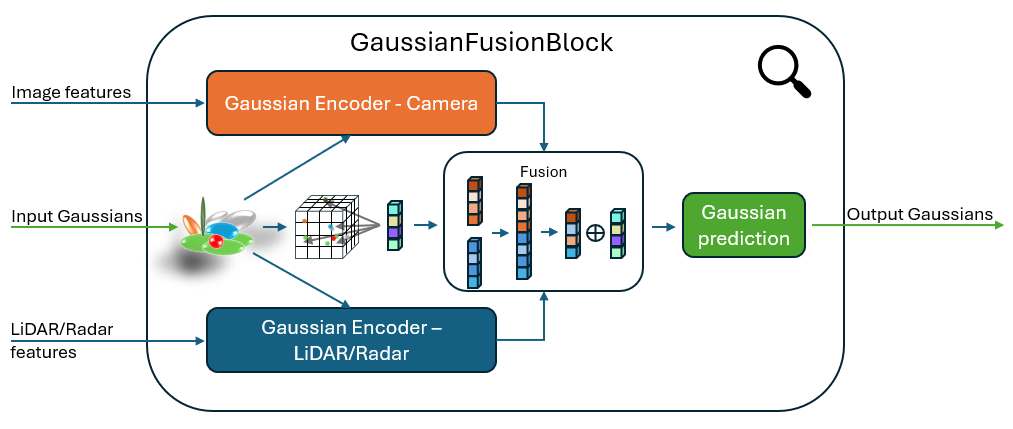}
\caption{\textbf{GaussianFusionBlock architecture.} Modality-agnostic Gaussian encoders employ a deformable attention mechanism to extract relevant features for each initial Gaussian. These per-Gaussian features are then fed to the fusion module along with the resulting features of the sparse convolution layer applied to voxelized Gaussians. Fusion module concatenates Gaussian encoder features, applies an MLP fuser, and adds sparse convolution features to create unified Gaussian features. These features are then used in the Gaussian prediction module to refine initial Gaussians by predicting mean offset, scale, rotation, opacity, and semantic class.}
\label{fig:gaussianfusionblock}
\end{figure*}
In this section, we present our seamless sensor fusion approach for 3D semantic occupancy prediction using 3D Gaussians. Our approach extracts features from sensors using sensor-specific encoders. Sensor-specific features are then fed into the Gaussian encoder blocks, where modality-agnostic deformable attention \citep{zhu2020deformable} is utilized to extract Gaussian-centric features. These features are later fused and used to refine the properties of probabilistic semantic 3D Gaussians. Finally, these refined Gaussians are splatted to generate a 3D voxel grid representation.

Sensor feature extraction:

We employ tailored feature extraction pipelines for each modality to effectively encode information from our diverse set of sensors.

For the surrounding camera images \( I=\{I_i\in \mathbb{R}^{3 \times H \times W}\}_{i=1}^{N} \) where H, W, N are height, width, and number of cameras, we utilize a combination of ResNet101-DCN \citep{he2016deep, dai2017dcn} as our 2D backbone and a Feature Pyramid Network (FPN) \citep{lin2017feature} as the neck. This architecture allows us to generate multi-scale image features \( F^{cam}_{i}=\{F^{cam}_{i,j}\in \mathbb{R}^{C_i \times H_i \times W_i}\}_{j=1}^{M} \) for \(i\)-th image where M is the number of scales.

To process the LiDAR point cloud \( P^{lidar} \in  \mathbb{R}^{C \times H \times W}\), we first perform voxelization of the input data. Following this, we employ a VoxelNet \citep{zhou2018voxelnet} encoder along with an FPN \citep{lin2017feature} to produce multi-scale Bird's-Eye-View (BEV) features \( F^{lidar}=\{F^{lidar}_{i}\in \mathbb{R}^{C_i \times H_i \times W_i}\}_{i=1}^{M} \).

Our radar encoder first pillarizes the input point cloud \( P^{radar} \in \mathbb{R}^{C \times H \times W}\). The pillarization is a form of voxelization where the 3D space is divided into vertical pillars. It then encodes the radar features using a PointPillars \citep{lang2019pointpillars} voxel encoder and a middle encoder. Radar data, while providing valuable information about object velocity, typically yields sparse features \(F^{radar}\in \mathbb{R}^{C \times H \times W}\) compared to camera and LiDAR.

Inspired by the advancements in GaussianFormer \citep{huang2024gaussianformer} and GaussianFormer-2 \citep{huang2024probabilistic}, our method reconstructs semantic probabilistic 3D Gaussians to represent the scene occupancy. This object-centric approach aims to overcome the limitations of dense grid-based representations by sparsely modeling the scene with learnable Gaussians. To initialize the Gaussian properties, we randomly sample a desired number of Gaussians \(\mathcal{G} = \{m_i, s_i, r_i, a_i, c_i\}_{i=1}^P\), use learnable parameters for initialization, or choose the desired number of points from the input LiDAR point cloud for Gaussian means. Our model comprises three key modules: Gaussian encoder, fusion module, and Gaussian prediction module.

Gaussian encoder:

To refine initial Gaussians, our method uses independent instances of the modality-agnostic Gaussian encoder module, which extracts a feature vector for each Gaussian. The encoder samples multiple 3D reference points around the center of a Gaussian \(m\) while taking into consideration the Gaussian's shape and size determined by its scale and rotation matrices \(S, R\): 
\begin{equation}
    \Delta m = RS\Phi^{offset}(q), \ \ \
    R = \{ m + \Delta m_i \}_{i=1}^{N_R}
\end{equation}
where \(\Phi^{offset}, q\) denote the MLP for offset prediction, and the input query for the specific Gaussian, respectively. Initial queries for the first block are randomly sampled.
These 3D reference points are then projected onto the feature maps of the input sensors using the sensors' intrinsic and extrinsic parameters. This projection establishes a correspondence between the 3D Gaussian and the 2D features captured by the sensors.
To extract relevant information from the sensor feature maps at the projected reference points, the Gaussian Encoder utilizes a deformable attention function \citep{zhu2020deformable}:
\begin{equation}
    F^{GE} = \sum_{i=1}^N \sum_{j=1}^{N_R} DA(Q, P(R), F^{sensor})
\end{equation}
where \(DA(), P(), F^{sensor}, N, N_R\) denote deformable attention function, projection from world to sensor features coordinates, sensor feature maps, number of sensor inputs (number of cameras or radars), and number of reference points, respectively.
The attention mechanism computes a weighted sum of the features sampled around these refined reference points, where the weights indicate the relevance of each sampled feature. This aggregated feature vector represents the visual (or other sensory) information associated with the 3D Gaussian.

Fusion:

The fusion module is designed to effectively integrate multi-modal sensor information to generate a unified feature representation for each 3D Gaussian. Given a set of per-Gaussian feature vectors extracted from individual sensors such as cameras, LiDAR, and radar, the fusion module aims to create a comprehensive descriptor that leverages the complementary strengths of each modality.
The feature vectors derived from each available sensor modality are concatenated along the feature channel dimension. This operation results in a single, extended feature vector that aggregates information from all input sensors. If we denote the feature vector for a Gaussian from the \(i\)-th sensor as \(F_i\), the concatenation operation can be represented as: \(F^{concatenated} = [F_1, F_2, ..., F_n]\) where \(n\) is the number of sensor modalities being utilized. Following the concatenation, the combined feature vector is passed through a Multi-Layer Perceptron \(\Phi^{fusion}\).
The MLP serves to learn complex interdependencies between features originating from different sensors, weigh the contribution of different sensor features based on their relevance for representing the 3D Gaussian, and reduce the dimensionality of the concatenated feature vector to a more manageable and informative unified representation.

The feature vector for each 3D Gaussian on the output of the MLP encapsulates the integrated information from all considered sensor modalities, providing a richer and more robust descriptor of the local 3D space represented by the Gaussian. 
In parallel, input Gaussians are represented as a point cloud of their means. The voxelized representation is then fed to 3D sparse convolutions to leverage the spatial relationships between these Gaussians, inspired by the self-encoding module from GaussianFormer \citep{huang2024gaussianformer}. Sparse convolutions \(SC\) are particularly suitable for processing sparse 3D data, like our Gaussian means, as they only operate on occupied voxels. This pathway allows each Gaussian to gather contextual information from its neighboring Gaussians, understanding the local spatial arrangement and potential relationships between scene elements represented by these Gaussians. Finally, to combine the benefits of both pathways, the output vectors from the 3D sparse convolutions are added to the feature vectors generated by the MLP for each corresponding Gaussian:
\begin{equation}
    Q = F^{unified}= \Phi^{fusion}(F^{concatenated}) \oplus SC(m)
\end{equation}
This element-wise addition fuses the modality-specific information (refined by the MLP) with the spatial context information (learned by the sparse convolutions).
These unified Gaussian feature vectors are subsequently used for refining the properties of the Gaussians, ultimately contributing to the accurate prediction of 3D semantic occupancy. 

Gaussian prediction:

Leveraging the information aggregated from multi-modal sensor inputs, the Gaussian Prediction module updates the parameters of each Gaussian to better represent the surrounding 3D scene. The functionality of the Gaussian prediction module draws inspiration from the refinement module employed in GaussianFormer \citep{huang2024gaussianformer}. Specifically, for each 3D Gaussian, the unified feature vector from the Fusion module serves as input to a multi-layer perceptron \(\Phi^{refine}\). This MLP decodes intermediate Gaussian properties, including offset to the mean ($\hat{m}$), scale ($\hat{s}$), rotation ($\hat{r}$), opacity ($\hat{a}$), and semantic logits ($\hat{c}$):
\begin{equation}
    (\hat{m}, \hat{s}, \hat{r}, \hat{a}, \hat{c}) = \Phi^{refine}(Q)
\end{equation}

Following the refinement step, the updated Gaussian properties and unified Gaussian feature vectors are fed to the next Gaussian fusion block for subsequent refinement. This iterative refinement process allows for a progressively more accurate and semantically meaningful representation of the 3D scene. After the final refinement, the collection of refined 3D Gaussians is used to generate the final 3D semantic occupancy prediction. To achieve this, we employ the Gaussian-to-Voxel splatting method \(GS\) proposed in GaussianFormer-2 \citep{huang2024probabilistic}:
\begin{equation}
    o(x) = GS(m,s,r,a,c)
\end{equation}
where x denotes the voxel position.
This approach utilizes probabilistic Gaussian superposition to transform the sparse Gaussian representation into a dense voxel grid occupancy prediction.

To train GaussianFusionOcc, we use lovasz-softmax loss \citep{berman2018lovasz} $L_{lovasz}$ and binary cross entropy loss $L_{BCE}$. During the training, the model splats Gaussians to a voxel grid representation in each iteration of the refinement, so the losses could be calculated for each iteration. We can then express the total loss as:
\begin{equation}
    L^{total} = \sum^N_{i=1}(L^{lovasz}_i + L^{BCE}_i)
\end{equation}
where $N$ is the number of refinement iterations.
\section{Experiments} \label{sec:experiments}

In this section, we present a comprehensive evaluation of our proposed GaussianFusionOcc framework for 3D semantic occupancy prediction. We assess its performance on the nuScenes validation set, including challenging rainy and nighttime scenarios, and analyze the contribution of key components through ablation studies. Efficiency comparisons are also provided.

\subsection{Dataset details}
We use the nuScenes \citep{caesar2020nuscenes} dataset for training and evaluating our GaussianFusionOcc model. NuScenes is a large-scale multi-modal dataset designed for autonomous driving research, featuring 1000 scenes of approximately 20 seconds each, with keyframes annotated at 2Hz. Each keyframe is equipped with data from a full suite of sensors, including 6 surround-view cameras with a 360° horizontal field of view, 5 radars, and 1 LiDAR. The dataset also provides 3D bounding box annotations for 23 object classes and 8 attributes.
While nuScenes is rich in various annotations, dense 3D semantic occupancy ground truth is not natively provided. To address this, we leverage the ground truth occupancy annotations generated by SurroundOcc \citep{wei2023surroundocc}.
The occupancy prediction range is set to [-50m, 50m] for the X and Y axes and [-5m, 3m] for the Z axis. The voxel resolution for the occupancy grid is set at 0.5 meters, resulting in a defined 3D volume for prediction. The nuScenes dataset is officially divided into training (700 scenes), validation (150 scenes), and test (150 scenes) splits, and we conduct our training on the training split and evaluate the performance of our model on the validation split.

\subsection{Implementation details} \label{subsec:implementation_details}
GaussianFusionOcc uses a ResNet101-DCN \citep{he2016deep, dai2017dcn} as the image feature backbone, initialized with weights from the FCOS3D \citep{wang2021fcos3d} checkpoint. Resulting image feature maps are then processed using FPN \citep{lin2017feature} to create multi-scale image features with 1/4, 1/8, 1/16, and 1/32 of the original resolution. For the LiDAR encoder, it uses VoxelNet \citep{zhou2018voxelnet} with weights from the FUTR3D \citep{chen2023futr3d} checkpoint, and FPN \citep{lin2017feature} producing multi-scale image features with the same downsampling as for image features. Radar encoder uses voxel encoder and middle encoder from PointPillars \citep{lang2019pointpillars} network initialized with weights from FUTR3D \citep{chen2023futr3d} checkpoint. For the 3D Gaussian representation, we use 6400 and 25600 Gaussians in our experiments. The model consists of 4 Gaussian fusion blocks, each consisting of Gaussian encoders, a fusion module, and a Gaussian prediction module. The Gaussian encoder and fusion module produce features with 128 channels. 
We train our GaussianFusionOcc model for 20 epochs with a batch size of 8 on an NVIDIA RTX A6000 GPU. We employ the AdamW optimizer \citep{loshchilov2017decoupled} with a weight decay of 0.01. The learning rate is warmed up to 2e-4 in the first 500 iterations, and subsequently decreased following a cosine schedule. For data augmentation, we apply photometric distortions to the input images.

\subsection{Main results}
\begin{table*}[h]
\centering
\caption{3D semantic occupancy prediction results on nuScenes validation set. * marks the model supervised by dense occupancy annotations, while the original was trained with LiDAR segmentation labels. ** marks the model using LiDAR initialization, *** the random initialization, while other models use learnable initialization. Modality notation: Camera (C), LiDAR (L), Radar (R)}
\resizebox{\textwidth}{!}{%
\begin{tabular}{l|c|c|c|cccccccccccccccc}
\toprule
Method & Modality & IoU & mIoU & \rotatebox{90}{barrier} & \rotatebox{90}{bicycle} & \rotatebox{90}{bus} & \rotatebox{90}{car} & \rotatebox{90}{constr. veh.} & \rotatebox{90}{motorcycle} & \rotatebox{90}{pedestrian} & \rotatebox{90}{traffic cone} & \rotatebox{90}{trailer} & \rotatebox{90}{truck} & \rotatebox{90}{driveable surf.} & \rotatebox{90}{other flat} & \rotatebox{90}{sidewalk} & \rotatebox{90}{terrain} & \rotatebox{90}{man made} & \rotatebox{90}{vegetation} \\
\midrule
MonoScene \citep{cao2022monoscene} & C & 23.96 & 7.31 & 4.03 & 0.35 & 8.00 & 8.04 & 2.90 & 0.28 & 1.16 & 0.67 & 4.01 & 4.35 & 27.72 & 5.20 & 15.13 & 11.29 & 9.03 & 14.86 \\
Atlas \citep{murez2020atlas} & C & 28.66 & 15.00 & 10.64 & 5.68 & 19.66 & 24.94 & 8.90 & 8.84 & 6.47 & 3.28 & 10.42 & 16.21 & 34.86 & 15.46 & 21.89 & 20.95 & 11.21 & 20.54 \\
BEVFormer \citep{li2024bevformer} & C & 30.50 & 16.75 & 14.22 & 6.58 & 23.46 & 28.28 & 8.66 & 10.77 & 6.64 & 4.05 & 11.20 & 17.78 & 37.28 & 18.00 & 22.88 & 22.17 & 13.80 & 22.21 \\
TPVFormer \citep{huang2023tri} & C & 11.51 & 11.66 & 16.14 & 7.17 & 22.63 & 17.13 & 8.83 & 11.39 & 10.46 & 8.23 & 9.43 & 17.02 & 8.07 & 13.64 & 13.85 & 10.34 & 4.90 & 7.37 \\
TPVFormer* \citep{huang2023tri} & C & 30.86 & 17.10 & 15.96 & 5.31 & 23.86 & 27.32 & 9.79 & 8.74 & 7.09 & 5.20 & 10.97 & 19.22 & 38.87 & 21.25 & 24.26 & 23.15 & 11.73 & 20.81 \\
OccFormer \citep{zhang2023occformer} & C & 31.39 & 19.03 & 18.65 & 10.41 & 23.92 & 30.29 & 10.31 & 14.19 & 13.59 & 10.13 & 12.49 & 20.77 & 38.78 & 19.79 & 24.19 & 22.21 & 13.48 & 21.35 \\
SurroundOcc \citep{wei2023surroundocc} & C & 31.49 & 20.30 & 20.59 & 11.68 & 28.06 & 30.86 & 10.70 & 15.14 & 14.09 & 12.06 & 14.38 & 22.26 & 37.29 & 23.70 & 24.49 & 22.77 & 14.89 & 21.86 \\
GaussianFormer \citep{huang2024gaussianformer} & C & 29.83 & 19.10 & 19.52 & 11.26 & 26.11 & 29.78 & 10.47 & 13.83 & 12.58 & 8.67 & 12.74 & 21.57 & 39.63 & 23.28 & 24.46 & 22.99 & 9.59 & 19.12 \\
GaussianFormer-2 \citep{huang2024probabilistic} & C & 31.74 & 20.82 & 21.39 & 13.44 & 28.49 & 30.82 & 10.92 & 15.84 & 13.55 & 10.53 & 14.04 & 22.92 & 40.61 & 24.36 & 26.08 & 24.27 & 13.83 & 21.98 \\
\midrule
L-CONet \citep{wang2023openoccupancy} & L & 39.40 & 17.70 & 19.20 & 4.00 & 15.10 & 26.90 & 6.20 & 3.80 & 6.80 & 6.00 & 14.10 & 13.10 & 39.70 & 19.10 & 24.00 & 23.90 & 25.10 & 35.70 \\
M-CONet \citep{wang2023openoccupancy} & C+L & 39.20 & 24.70 & 24.80 & 13.00 & 31.60 & 34.80 & 14.60 & 18.00 & 20.00 & 14.70 & 20.00 & 26.60 & 39.20 & 22.80 & 26.10 & 26.00 & 26.00 & 37.10 \\
OccFusion \citep{ming2024occfusion} & C+R & 33.97 & 20.73 & 20.46 & 13.98 & 27.99 & 31.52 & 13.68 & 18.45 & 15.79 & 13.05 & 13.94 & 23.84 & 37.85 & 19.60 & 22.41 & 21.20 & 16.16 & 21.81 \\
OccFusion \citep{ming2024occfusion} & C+L & 44.35 & 26.87 & 26.67 & 18.38 & 32.97 & 35.81 & 19.39 & 22.17 & 24.48 & 17.77 & 21.46 & 29.67 & 39.01 & 21.94 & 24.90 & 26.76 & 28.53 & 40.03 \\
OccFusion \citep{ming2024occfusion} & C+L+R & 44.66 & 27.30 & 27.09 & \textbf{19.56} & 33.68 & 36.23 & 21.66 & 24.84 & 25.29 & 16.33 & 21.81 & 30.01 & 39.53 & 19.94 & 24.94 & 26.45 & 28.93 & 40.41 \\
\midrule
GaussianFusionOcc** (Ours)& C & 37.05 & 22.43 & 22.46 & 14.19 & 28.66 & 29.93 & 15.10 & 17.08 & 18.22 & 9.71 & 18.21 & 25.12 & 37.19 & 20.59 & 23.39 & 23.65 & 22.91 & 32.55 \\
GaussianFusionOcc** (Ours) & C+R & 37.37 & 22.80 & 22.66 & 13.95 & 29.70 & 31.30 & 15.84 & 18.16 & 19.02 & 9.61 & 17.63 & 25.51 & 37.72 & 20.02 & 23.30 & 23.57 & 23.66 & 33.21 \\
GaussianFusionOcc*** (Ours)& L & \textbf{45.32} & 29.75 & 30.02 & 16.46 & 35.02 & 38.93 & 22.25 & 24.65 & 29.64 & 18.41 & \textbf{24.64} & 30.93 & \textbf{43.31} & \textbf{26.30} & 28.95 & 29.29 & 34.33 & 42.92 \\
GaussianFusionOcc (Ours)& C+L & 45.16 & 30.21 & 30.22 & 18.70 & 35.91 & 39.57 & \textbf{22.67} & \textbf{27.36} & 30.10 & 18.59 & 24.45 & 31.25 & 43.06 & 25.76 & 29.12 & 29.33 & 34.65 & 42.70 \\
GaussianFusionOcc (Ours)& C+L+R & 45.20 & \textbf{30.37} & \textbf{30.43} & 18.54 & \textbf{36.23} & \textbf{39.66} & 22.57 & 27.35 & \textbf{30.30} & \textbf{19.14} & 24.56 & \textbf{31.95} & 42.60 & 25.82 & \textbf{29.48} & \textbf{29.70} & \textbf{34.78} & \textbf{42.95} \\
\bottomrule
\end{tabular}%
}
\label{tab:nuscenes_val}
\end{table*}

The superior performance of our proposed GaussianFusionOcc framework is clearly demonstrated in Table \ref{tab:nuscenes_val}, where it outperforms a range of state-of-the-art methods across various categories of 3D semantic occupancy prediction on the nuScenes validation set \citep{caesar2020nuscenes}.

We compare our GaussianFusionOcc with several categories of existing methods:

\textbf{Comparison with planar-based models:} Our results indicate a significant improvement over planar-based methods such as BEVFormer \citep{li2024bevformer} and TPVFormer \citep{huang2023tri}. While these models lift 2D image features to a Bird's-Eye View (BEV) or Tri-Perspective View (TPV), the object-centric nature of Gaussians in GaussianFusionOcc allows for a more efficient and flexible representation. GaussianFusionOcc outperforms planar-based methods across all semantic categories, with especially pronounced gains for small or vertically complex classes like pedestrians, motorcycles, or traffic cones.

\textbf{Comparison with grid-based models:} GaussianFusionOcc demonstrates improved performance over grid-based representations like OccFormer \citep{zhang2023occformer}. This is attributed to the sparse modeling of the scene with learnable Gaussians, which better allocates representational capacity to complex object shapes and details, particularly in capturing fine-grained structures and handling varying object scales.

\textbf{Comparison with existing gaussian-based models:} Even when compared to other Gaussian-based methods, such as GaussianFormer-2 \citep{huang2024probabilistic}, GaussianFusionOcc demonstrates enhanced performance. This improvement can be attributed to the unique multi-sensor fusion mechanism employed in GaussianFusionOcc, which addresses the limitation of previous Gaussian-based approaches that rely solely on camera input. The superior results over GaussianFormer-2 highlight the effectiveness of our fusion framework in leveraging the benefits of the Gaussian representation for 3D semantic occupancy prediction with multi-sensor input.

\textbf{Comparison with sensor-fusion models:} Table \ref{tab:nuscenes_val} demonstrates that GaussianFusionOcc also surpasses sensor-fusion models such as M-CONet \citep{wang2023openoccupancy} and OccFusion \citep{ming2024occfusion}. Notably, GaussianFusionOcc achieves this superior performance with a significantly lower number of parameters (79.96M) compared to M-CONet (137M) and OccFusion (114.97M). This suggests that our novel approach, utilizing a more effective representation and processing of visual cues through 3D Gaussians, can surpass the performance of multi-sensor fusion techniques based on grid representations.

Overall, the results presented in Table \ref{tab:nuscenes_val} establish GaussianFusionOcc as a new state-of-the-art for 3D semantic occupancy prediction. Our model's superior performance across different architectural paradigms underscores the efficacy of its core design principles.

\subsection{Performance under challenging scenarios}

\begin{table*}[htb!]
\centering
\caption{3D semantic occupancy prediction results on rainy scenario subset of nuScenes validation set. All methods are trained with dense occupancy labels from \citep{wei2023surroundocc}. Modality notation: Camera (C), LiDAR (L), Radar (R).}
\resizebox{\textwidth}{!}{%
\begin{tabular}{l|c|c|c|ccccccccccccccccc}
\toprule
Method & Modality & IoU & mIoU & \rotatebox{90}{barrier} & \rotatebox{90}{bicycle} & \rotatebox{90}{bus} & \rotatebox{90}{car} & \rotatebox{90}{constr. veh.} & \rotatebox{90}{motorcycle} & \rotatebox{90}{pedestrian} & \rotatebox{90}{traffic cone} & \rotatebox{90}{trailer} & \rotatebox{90}{truck} & \rotatebox{90}{driveable surf.} & \rotatebox{90}{other flat} & \rotatebox{90}{sidewalk} & \rotatebox{90}{terrain} & \rotatebox{90}{man made} & \rotatebox{90}{vegetation} \\
\midrule
OccFusion \citep{ming2024occfusion} & C & 31.10 & 18.99 & 18.55 & 14.29 & 22.28 & 30.02 & 10.19 & 15.20 & 10.03 & 9.71 & 13.28 & 20.98 & 37.18 & 23.47 & 27.74 & 17.46 & 10.36 & 23.13 \\
SurroundOcc \citep{wei2023surroundocc} & C & 30.57 & 21.40 & 21.40 & 12.75 & 25.49 & 31.31 & 11.39 & 12.65 & 8.94 & 9.48 & 14.51 & 21.52 & 35.34 & \textbf{25.32} & 29.89 & 18.37 & 14.44 & 24.78 \\
OccFusion \citep{ming2024occfusion} & C+R & 33.75 & 20.78 & 20.14 & 16.33 & 26.37 & 32.39 & 11.56 & 17.08 & 11.14 & 10.54 & 13.61 & 22.42 & 37.50 & 22.79 & 29.50 & 17.58 & 17.06 & 26.49 \\
OccFusion \citep{ming2024occfusion} & C+L & 43.36 & 26.55 & 24.95 & 19.11 & 34.23 & 36.07 & 17.01 & 21.07 & 18.87 & 17.46 & 21.81 & 28.73 & 37.82 & 24.39 & 30.80 & 20.37 & 28.95 & 43.12 \\
OccFusion \citep{ming2024occfusion} & C+L+R & 43.50 & 26.72 & 25.30 & 18.71 & 33.58 & 36.28 & 17.76 & 22.44 & 20.80 & 15.89 & 22.63 & 28.75 & 39.28 & 22.72 & 30.78 & 20.15 & 28.99 & 43.37 \\
GaussianFusionOcc & C & 36.83 & 21.86 & 21.50 & 12.56 & 28.89 & 29.50 & 13.16 & 12.97 & 12.31 & 7.60 & 19.32 & 23.33 & 38.12 & 22.13 & 29.17 & 20.65 & 22.60 & 36.00 \\
GaussianFusionOcc & L & 43.80 & 28.40 & 27.16 & 16.79 & 36.99 & 38.09 & \textbf{21.08} & 16.09 & 25.48 & 17.60 & \textbf{26.67} & 30.37 & 40.36 & 24.30 & 33.59 & 22.47 & 33.28 & 44.16 \\
GaussianFusionOcc & C+R & 37.03 & 21.87 & 21.97 & 13.92 & 29.91 & 30.64 & 9.68 & 12.90 & 12.70 & 6.72 & 19.48 & 23.97 & 37.43 & 21.78 & 29.02 & 19.58 & 23.77 & 36.51 \\
GaussianFusionOcc & C+L & 44.28 & 29.19 & 28.10 & 19.84 & 36.28 & 38.90 & 18.11 & 21.13 & \textbf{26.14} & 17.95 & 25.79 & 29.92 & \textbf{41.72} & \textbf{27.35} & 34.99 & 22.85 & 33.84 & \textbf{44.10} \\
GaussianFusionOcc & C+L+R & \textbf{44.36} & \textbf{29.86} & \textbf{28.40} & \textbf{19.88} & \textbf{38.87} & \textbf{39.33} & 20.61 & \textbf{26.05} & 25.66 & \textbf{17.97} & 26.07 & \textbf{31.02} & 41.70 & 24.94 & \textbf{35.27} & \textbf{24.08} & \textbf{33.90} & 44.00 \\
\bottomrule
\end{tabular}%
}
\label{tab:nuscenes_rainy}
\end{table*}

\begin{table*}[htb!]
\centering
\caption{3D semantic occupancy prediction results on night scenario subset of nuScenes validation set. All methods are trained with dense occupancy labels from \citep{wei2023surroundocc}. Modality notation: Camera (C), LiDAR (L), Radar (R).}
\resizebox{\textwidth}{!}{%
\begin{tabular}{l|c|c|c|ccccccccccccccccc}
\toprule
Method & Modality & IoU & mIoU & \rotatebox{90}{barrier} & \rotatebox{90}{bicycle} & \rotatebox{90}{bus} & \rotatebox{90}{car} & \rotatebox{90}{constr. veh.} & \rotatebox{90}{motorcycle} & \rotatebox{90}{pedestrian} & \rotatebox{90}{traffic cone} & \rotatebox{90}{trailer} & \rotatebox{90}{truck} & \rotatebox{90}{driveable surf.} & \rotatebox{90}{other flat} & \rotatebox{90}{sidewalk} & \rotatebox{90}{terrain} & \rotatebox{90}{man made} & \rotatebox{90}{vegetation} \\
\midrule
OccFusion \citep{ming2024occfusion} & C & 24.49 & 9.99 & 10.40 & 12.03 & 0.00 & 29.94 & 0.00 & 9.92 & 4.88 & 0.91 & 0.00 & 17.79 & 29.10 & 2.37 & 10.80 & 9.40 & 8.68 & 13.57 \\
SurroundOcc \citep{wei2023surroundocc} & C & 24.38 & 10.80 & 10.55 & 14.60 & 0.00 & 31.05 & 0.00 & 8.26 & 5.37 & 0.58 & 0.00 & 18.75 & 30.72 & 2.74 & 12.39 & 11.53 & 10.52 & 15.77 \\
OccFusion \citep{ming2024occfusion} & C+R & 27.09 & 11.13 & 10.78 & 12.77 & 0.00 & 33.50 & 0.00 & 12.72 & 4.91 & 0.61 & 0.00 & 19.97 & 29.51 & 0.94 & 12.15 & 10.72 & 11.81 & 17.72 \\
OccFusion \citep{ming2024occfusion} & C+L & 41.38 & 15.26 & 12.74 & 13.52 & 0.00 & 35.85 & 0.00 & 15.33 & 13.19 & 0.83 & 0.00 & 23.78 & 32.49 & 0.92 & 14.24 & 20.54 & 23.57 & 37.10 \\
OccFusion \citep{ming2024occfusion} & C+L+R & 41.47 & 15.82 & 13.27 & \textbf{13.53} & 0.00 & 36.41 & 0.00 & 19.71 & 12.16 & \textbf{2.04} & 0.00 & 25.90 & 32.44 & 0.80 & 14.30 & 21.06 & 24.49 & 37.00 \\
GaussianFusionOcc & C & 32.05 & 11.28 & 6.13 & 4.41 & 0.00 & 30.40 & 0.00 & 12.51 & 3.06 & 0.43 & 0.00 & 21.55 & 28.91 & \textbf{3.54} & 13.10 & 13.71 & 16.12 & 26.58 \\
GaussianFusionOcc & L & \textbf{43.00} & 18.76 & \textbf{19.62} & 8.32 & 0.00 & 39.74 & 0.00 & 26.85 & 13.28 & 0.21 & 0.00 & \textbf{39.11} & 39.01 & 2.02 & \textbf{20.04} & \textbf{22.81} & 29.03 & \textbf{40.12} \\
GaussianFusionOcc & C+R & 33.08 & 12.31 & 9.18 & 7.68 & 0.00 & 32.05 & 0.00 & 14.14 & 5.75 & 0.00 & 0.00 & 20.89 & 31.17 & 2.52 & 12.87 & 15.10 & 17.29 & 28.35 \\
GaussianFusionOcc & C+L & 42.78 & \textbf{18.66} & 16.09 & 12.27 & 0.00 & \textbf{39.82} & 0.00 & 27.66 & 13.68 & 0.07 & 0.00 & 38.25 & \textbf{40.10} & 2.07 & 19.64 & 19.82 & \textbf{29.55} & 39.61 \\
GaussianFusionOcc & C+L+R & 42.51 & 18.45 & 12.15 & 11.47 & 0.00 & 39.77 & 0.00 & \textbf{29.58} & \textbf{15.27} & 0.04 & 0.00 & 37.08 & 37.13 & 2.58 & 19.94 & 20.63 & 29.42 & 40.10 \\
\bottomrule
\end{tabular}%
}
\label{tab:nuscenes_night}
\end{table*}

We further evaluate the performance of GaussianFusionOcc under challenging weather and lighting conditions on subsets of the nuScenes validation set, specifically focusing on rainy and nighttime scenarios. The results are presented in Table \ref{tab:nuscenes_rainy} and Table \ref{tab:nuscenes_night}.

\textbf{Rainy scenario:} As shown in Table \ref{tab:nuscenes_rainy}, GaussianFusionOcc (C+L+R) achieves an mIoU of 29.86\% and IoU of 44.36\%. Compared to single-modality variants, the fusion of camera, LiDAR, and radar consistently improves performance in rainy conditions. These results highlight the robustness of our multi-sensor fusion strategy to adverse weather, where cameras may struggle.
 
\textbf{Night scenario:} Table \ref{tab:nuscenes_night} shows the performance in nighttime scenarios. GaussianFusionOcc (C+L+R) obtains an mIoU of 18.45\% and an IoU of 42.51\%. The inclusion of LiDAR and radar significantly boosts performance in low-light conditions compared to camera-only methods, demonstrating enhanced long-distance sensing capabilities. GaussianFusionOcc achieves higher scores with the addition of radar data, especially for dynamic object categories such as cars, bicycles, and motorcycles, due to radar's robustness to illumination changes and its capacity to provide velocity information.

The experimental results strongly suggest that GaussianFusionOcc demonstrates significant performance gains in 3D semantic occupancy prediction, particularly in adverse weather (rainy) and low-light (night) scenarios.

\subsection{Efficiency analysis}

\begin{table*}[h]
\centering
\caption{Efficiency comparison of multi-modal 3D semantic occupancy prediction on nuScenes validation set. Latency results marked with * were taken from the paper that introduced the model and were measured on a different GPU.}
\resizebox{\textwidth}{!}{%
\begin{tabular}{l|c|c|c|c|c|c}
\toprule
Method & Modality & IoU & mIoU & Params & Memory (GB) & Latency (ms) \\
\midrule
L-CONet \citep{wang2023openoccupancy} & L & 39.40 & 17.70 & - & 8.5 & - \\
M-CONet \citep{wang2023openoccupancy} & C+L & 39.20 & 24.70 & 137M & 24 & - \\
OccFusion \citep{ming2024occfusion} & C+R & 33.97 & 20.73 & 92.71M & 5.56 & 588* \\
OccFusion \citep{ming2024occfusion} & C+L & 44.35 & 26.87 & 92.71M & 5.56 & 591* \\
OccFusion \citep{ming2024occfusion} & C+L+R & 44.66 & 27.30 & 114.97M & 5.78 & 601* \\
\midrule
GaussianFusionOcc & C & 37.05 & 22.43 & 71.39M & 2.44 & 282\\
GaussianFusionOcc & C+R & 37.37 & 22.80 & 71.86M & 2.58 & 315 \\
GaussianFusionOcc & L & \textbf{45.32} & 29.75 & 34.14M & 0.49 & 179\\
GaussianFusionOcc & C+L & 45.16 & 30.21 & 79.63M & 2.61 & 460 \\
GaussianFusionOcc & C+L+R & 45.20 & \textbf{30.37} & 79.96M & 2.90 & 480 \\
\bottomrule
\end{tabular}%
}
\label{tab:nuscenes_val_efficiency}
\end{table*}

Table \ref{tab:nuscenes_val_efficiency} presents the efficiency comparison of multi-modal 3D semantic occupancy prediction methods, reporting the number of learnable parameters, memory usage, and latency.

GaussianFusionOcc achieves state-of-the-art accuracy while substantially reducing computational and memory requirements compared to grid-based baselines.
In the LiDAR-only setting, GaussianFusionOcc achieves 45.32\% IoU and 29.75\% mIoU, while requiring only 34.14 million parameters and 0.49GB of memory with inference time 179 ms. The GaussianFusionOcc further improves prediction performance in multi-modal settings, while staying significantly more efficient than comparable multi-modal methods. The memory and parameter savings of GaussianFusionOcc can be attributed to its adaptive, object-centric 3D Gaussian representation. This design enables the model to maintain a low memory footprint even as the number of sensor modalities increases. Latency measurements further highlight the practical advantages of GaussianFusionOcc, showing significantly faster inference than compared models. 
We report an additional ablation study on the number of Gaussians, the number of channels for feature representation, and the initialization strategy, in the appendix  \ref{sec:ablation}.

\subsection{Qualitative analysis}
\begin{figure*}[htb!]
    \centering
    \footnotesize
    \setlength{\fboxsep}{0pt}%
    \setlength{\fboxrule}{0.1pt}%
    \setlength{\tabcolsep}{1pt}
    \renewcommand{\arraystretch}{0.1}
    \begin{tabular}{>{\centering\arraybackslash}m{0.34\textwidth} >{\centering\arraybackslash}m{0.22\textwidth} >{\centering\arraybackslash}m{0.22\textwidth} >{\centering\arraybackslash}m{0.22\textwidth}}
        \textbf{Input Images} & \textbf{Ground Truth} & \textbf{GaussianFusionOcc (C)} & \textbf{GaussianFusionOcc (C+L+R)} \\
        \setlength{\tabcolsep}{0pt}
        \renewcommand{\arraystretch}{0}
        \vspace*{\fill}
        \begin{tabular}{ccc}
            \includegraphics[width=0.33\linewidth]{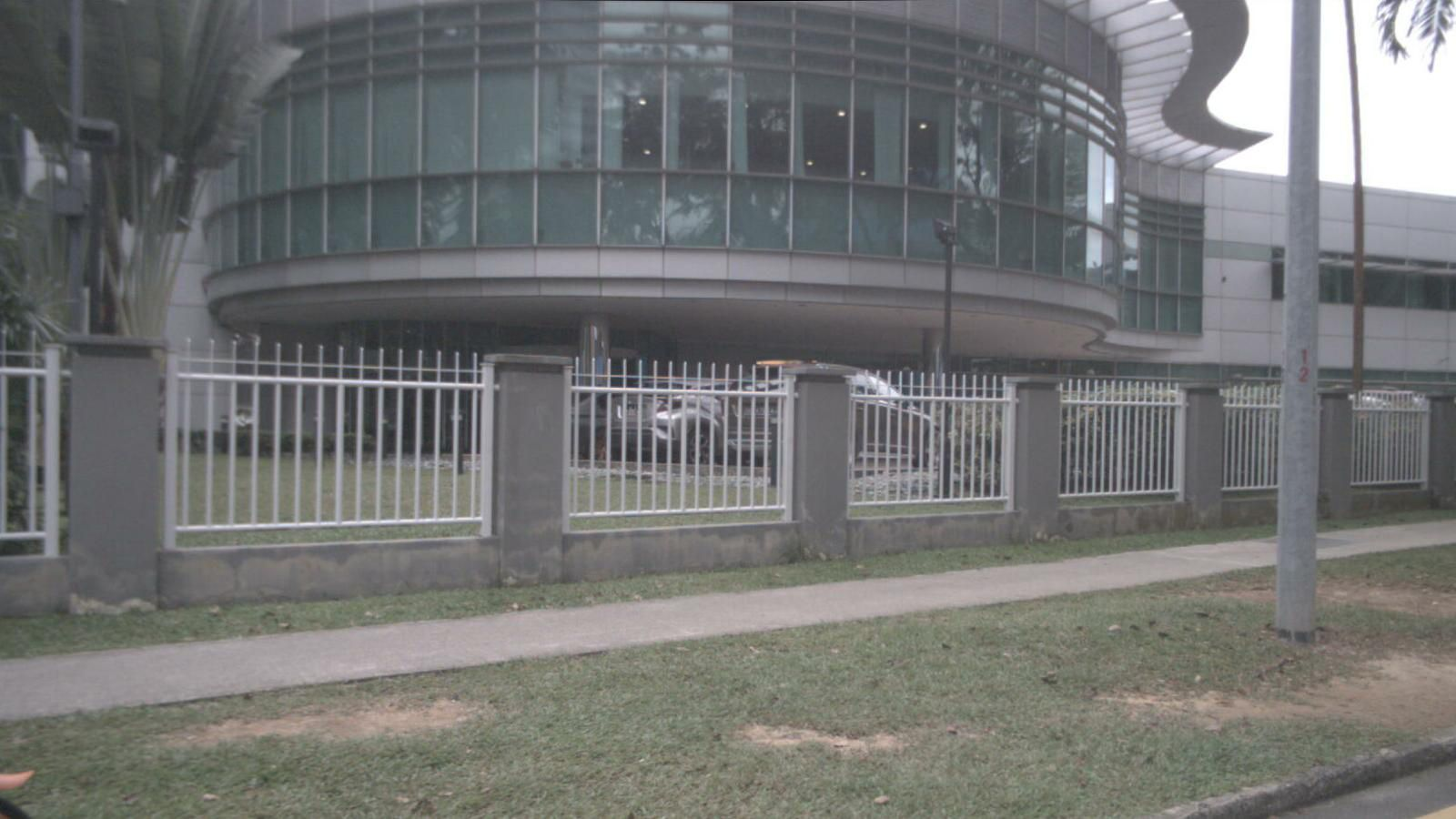} & \includegraphics[width=0.33\linewidth]{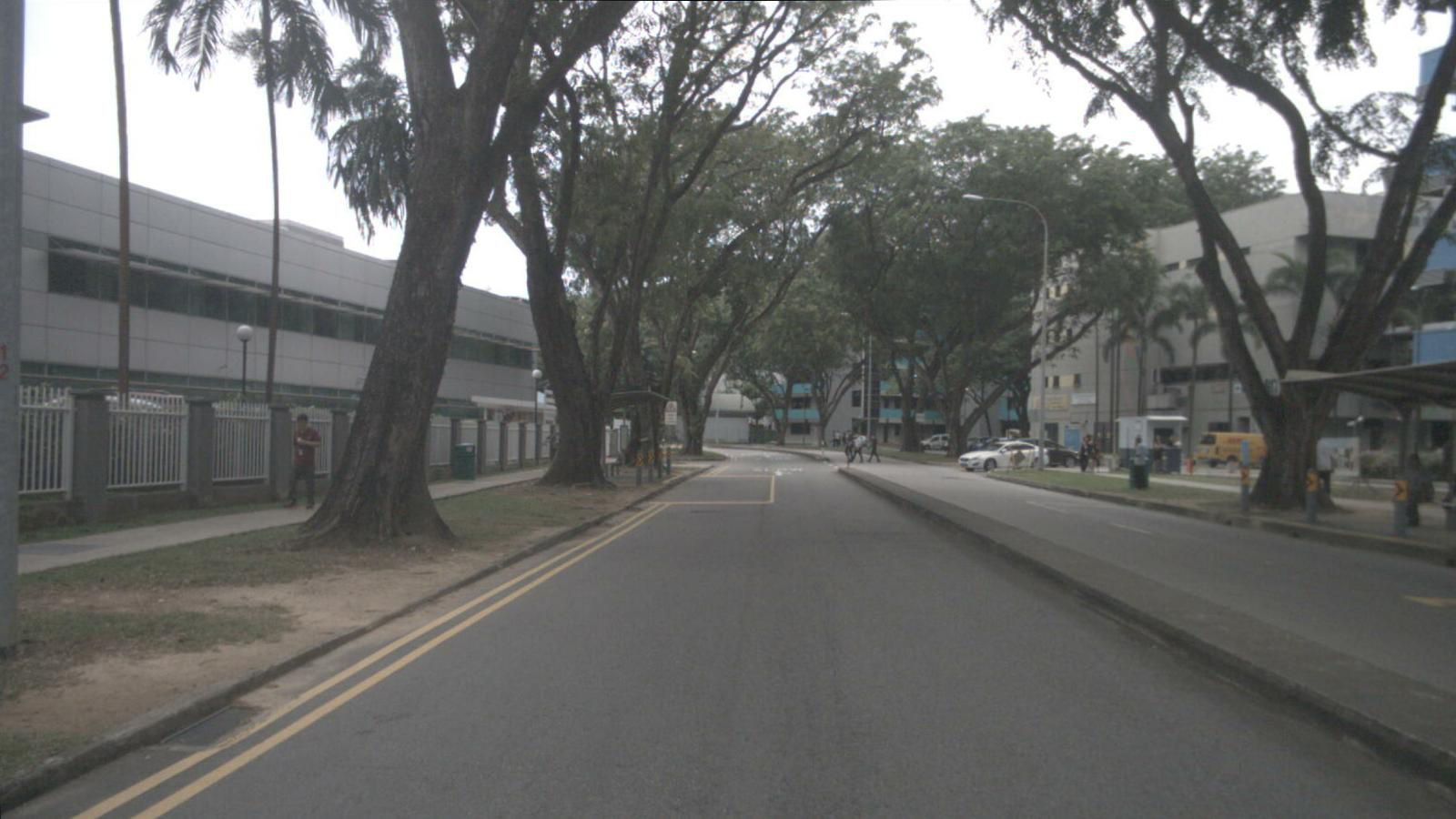} & \includegraphics[width=0.33\linewidth]{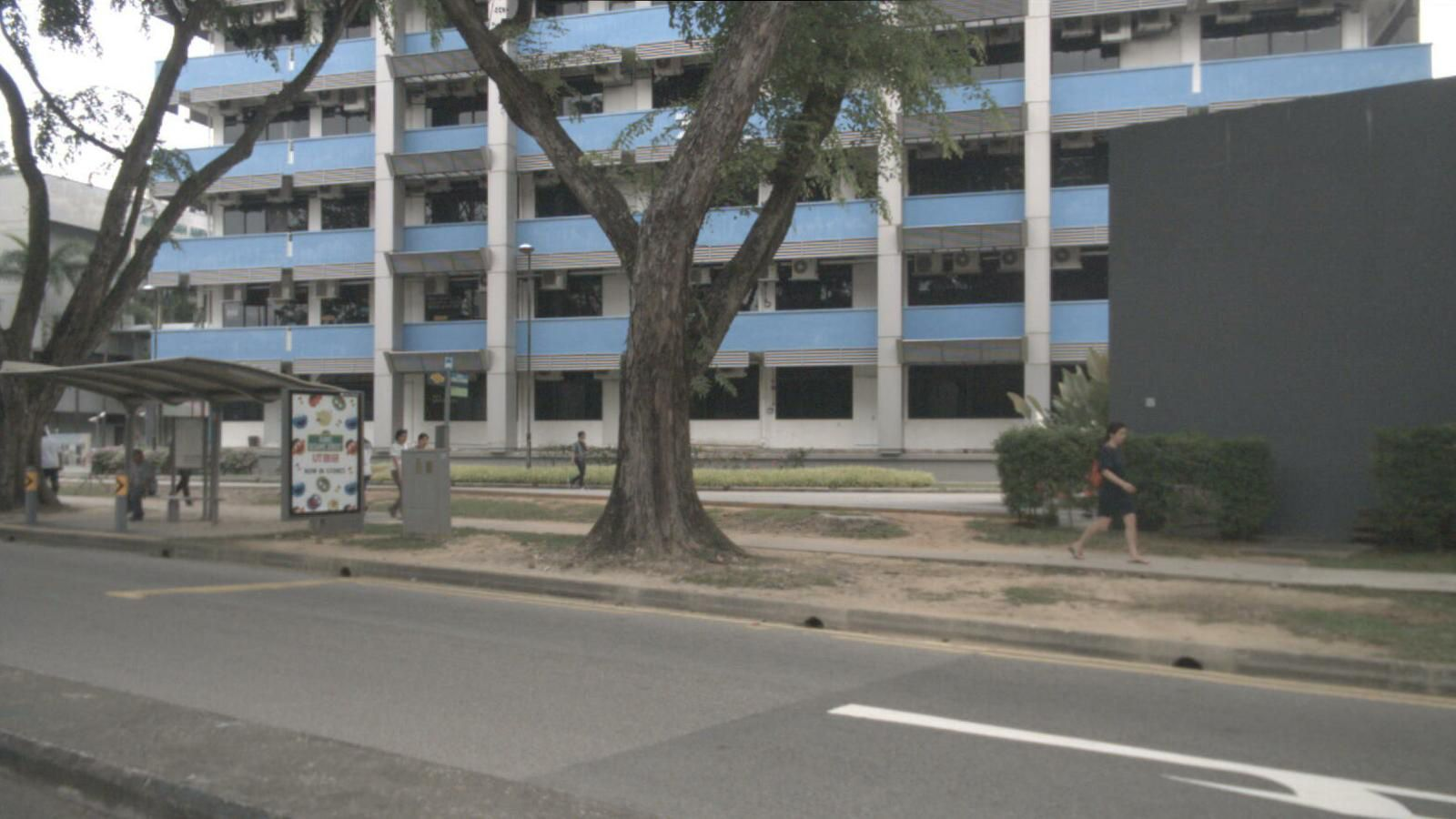} \\
            \includegraphics[width=0.33\linewidth]{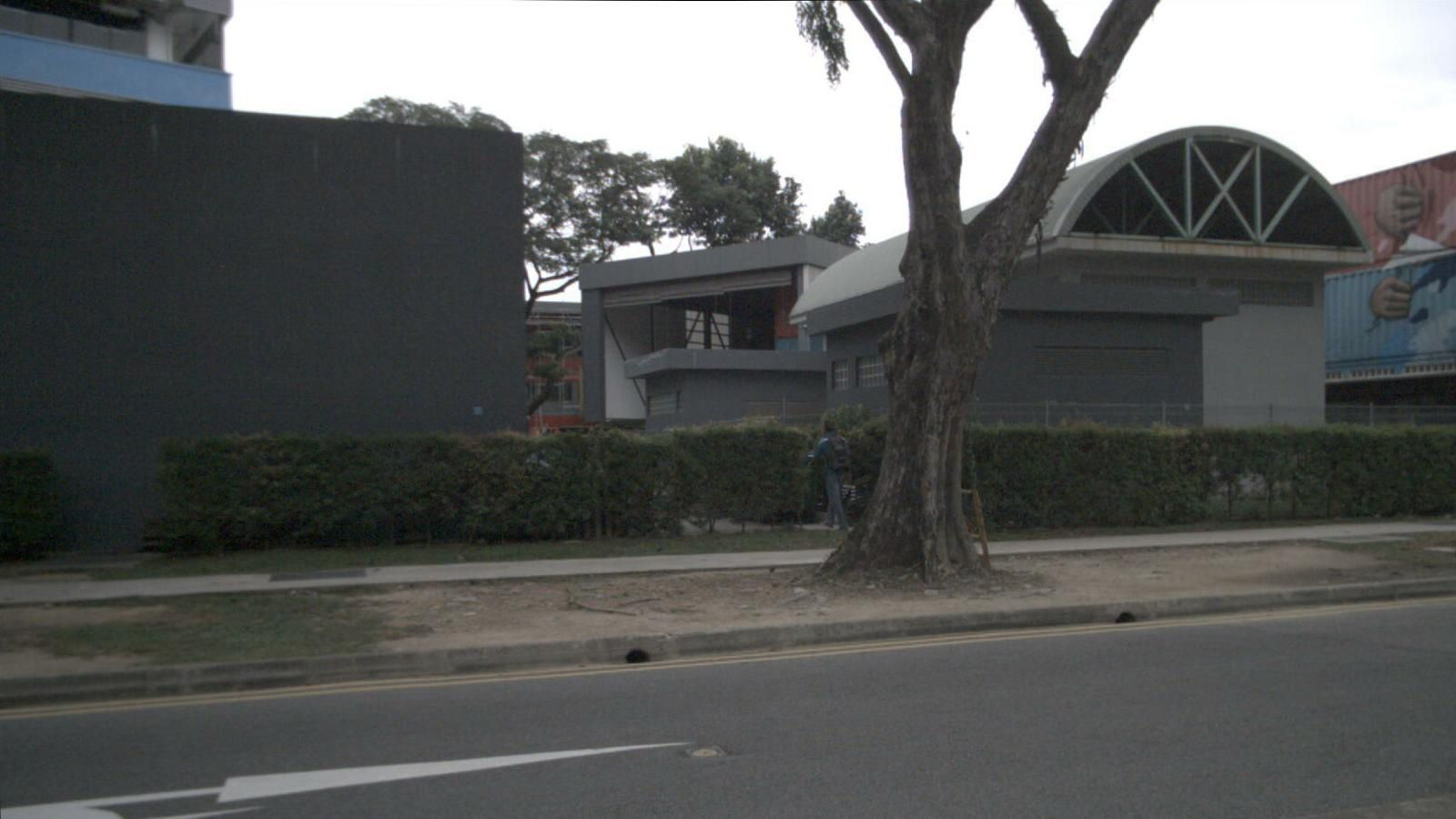} & \includegraphics[width=0.33\linewidth]{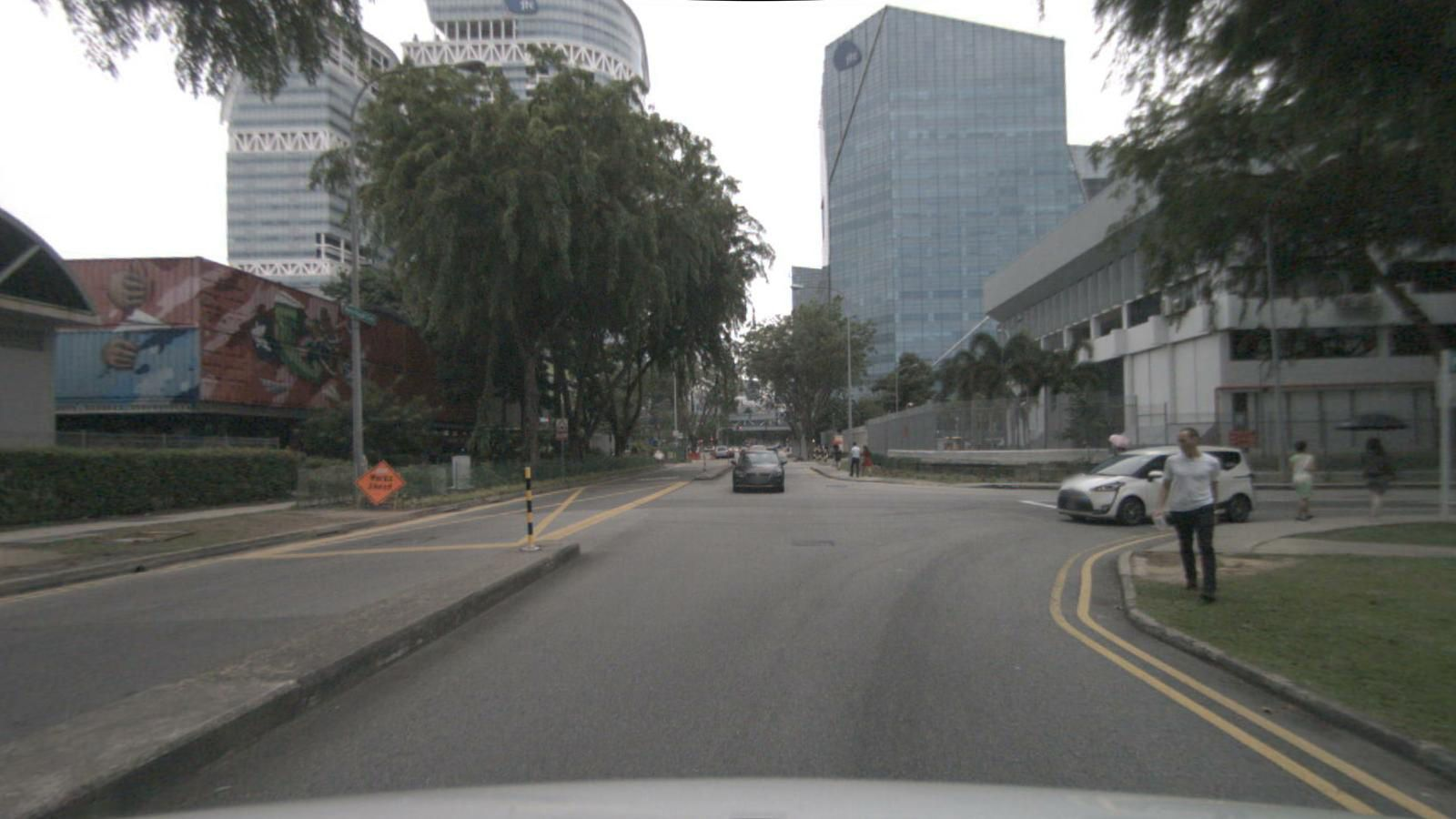} & \includegraphics[width=0.33\linewidth]{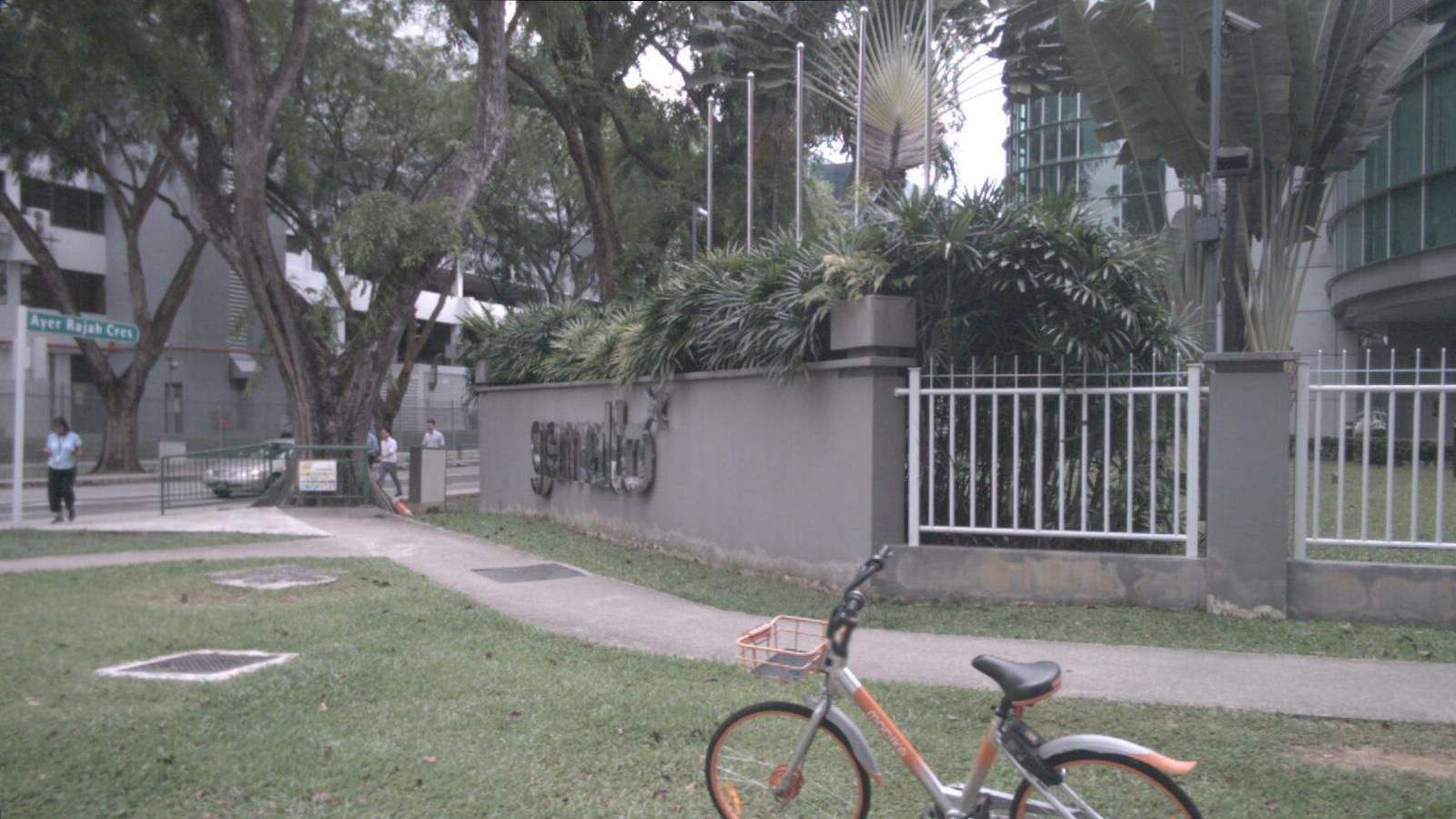} \\
        \end{tabular}
        \vspace*{\fill} & \fbox{\includegraphics[width=1\linewidth]{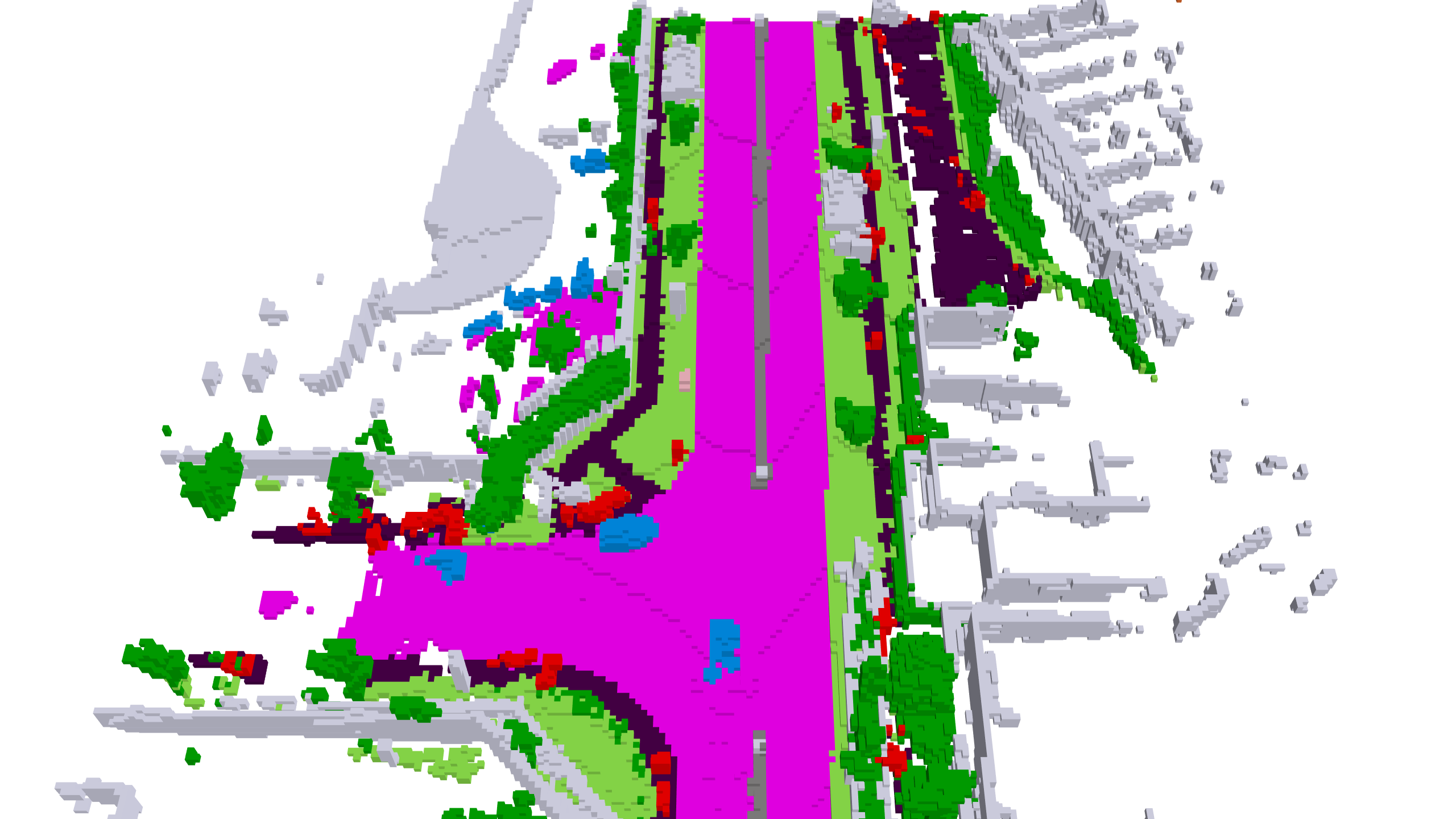}} & \fbox{\includegraphics[width=1\linewidth]{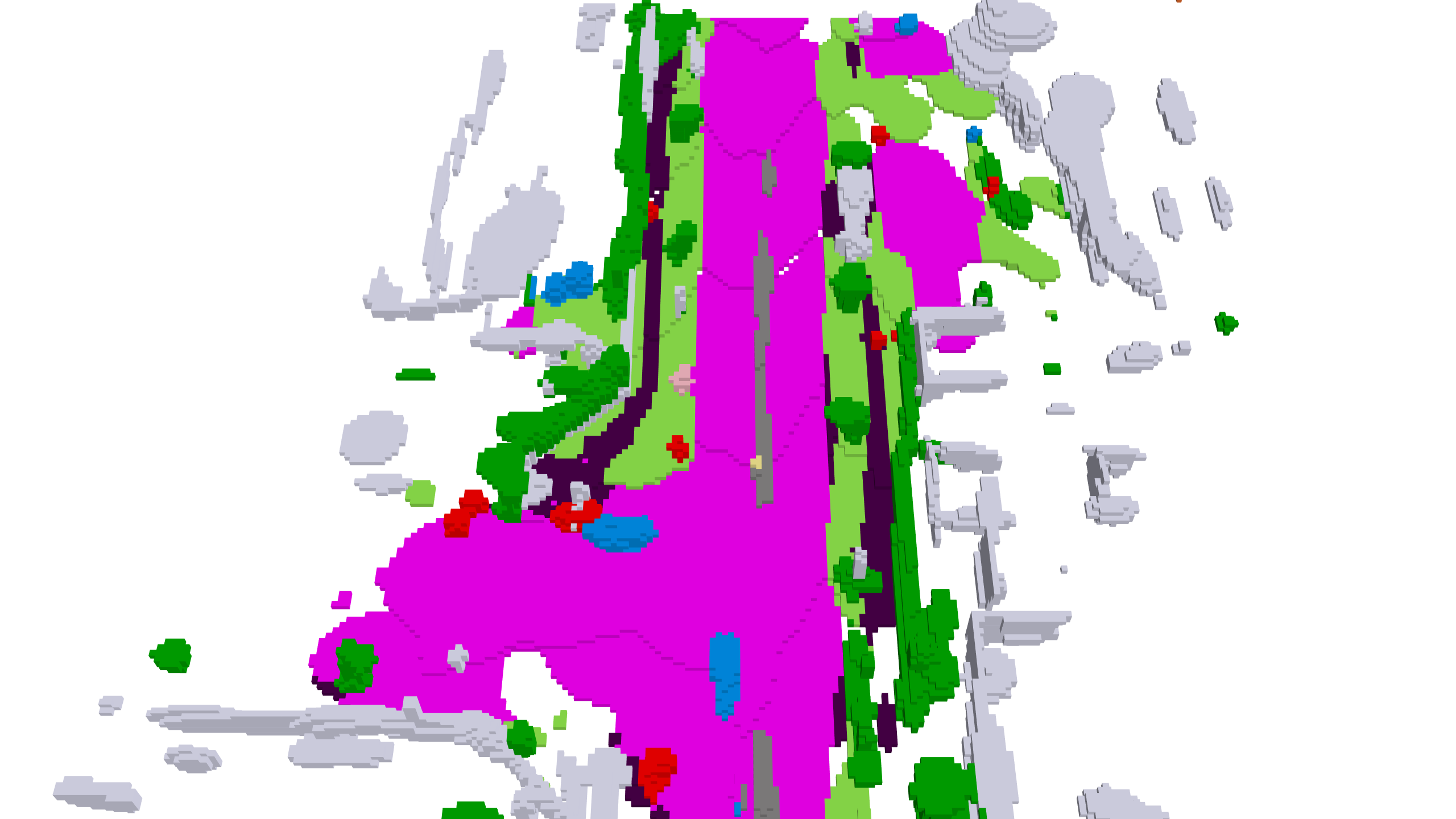}} & \fbox{\includegraphics[width=1\linewidth]{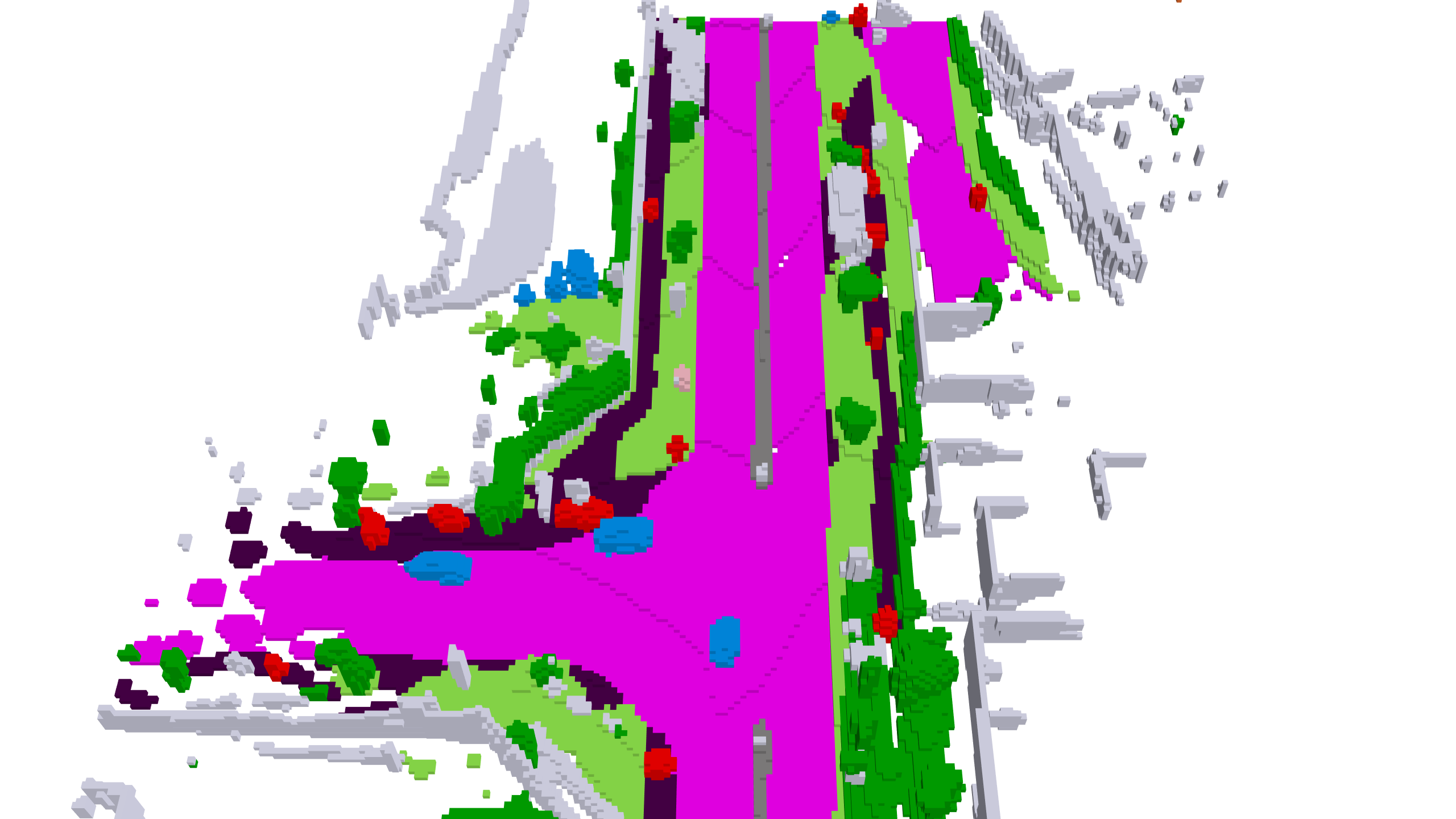}} \\
        \setlength{\tabcolsep}{0pt}
        \renewcommand{\arraystretch}{0}
        \vspace*{\fill}
        \begin{tabular}{ccc}
            \includegraphics[width=0.33\linewidth]{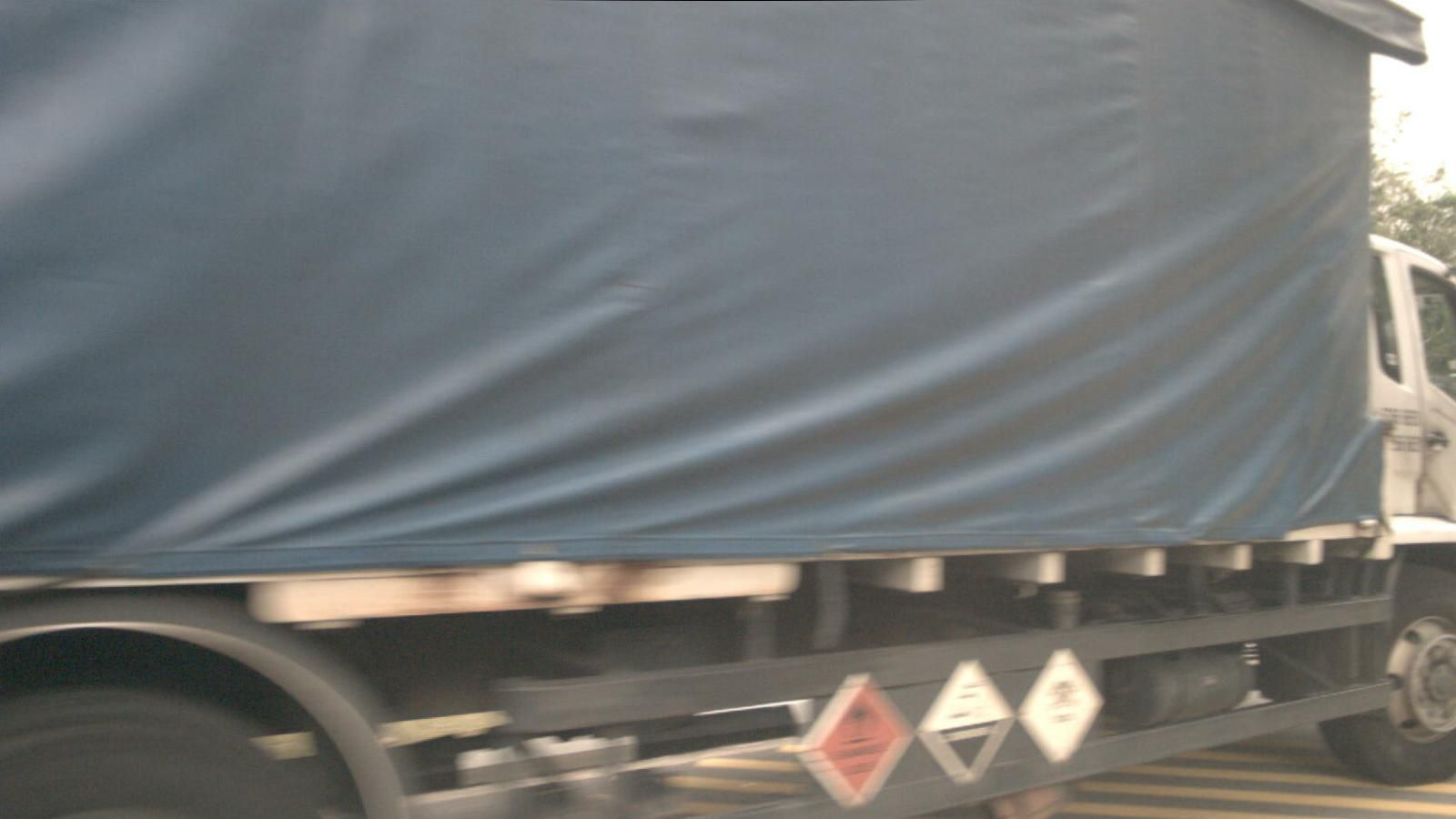} & \includegraphics[width=0.33\linewidth]{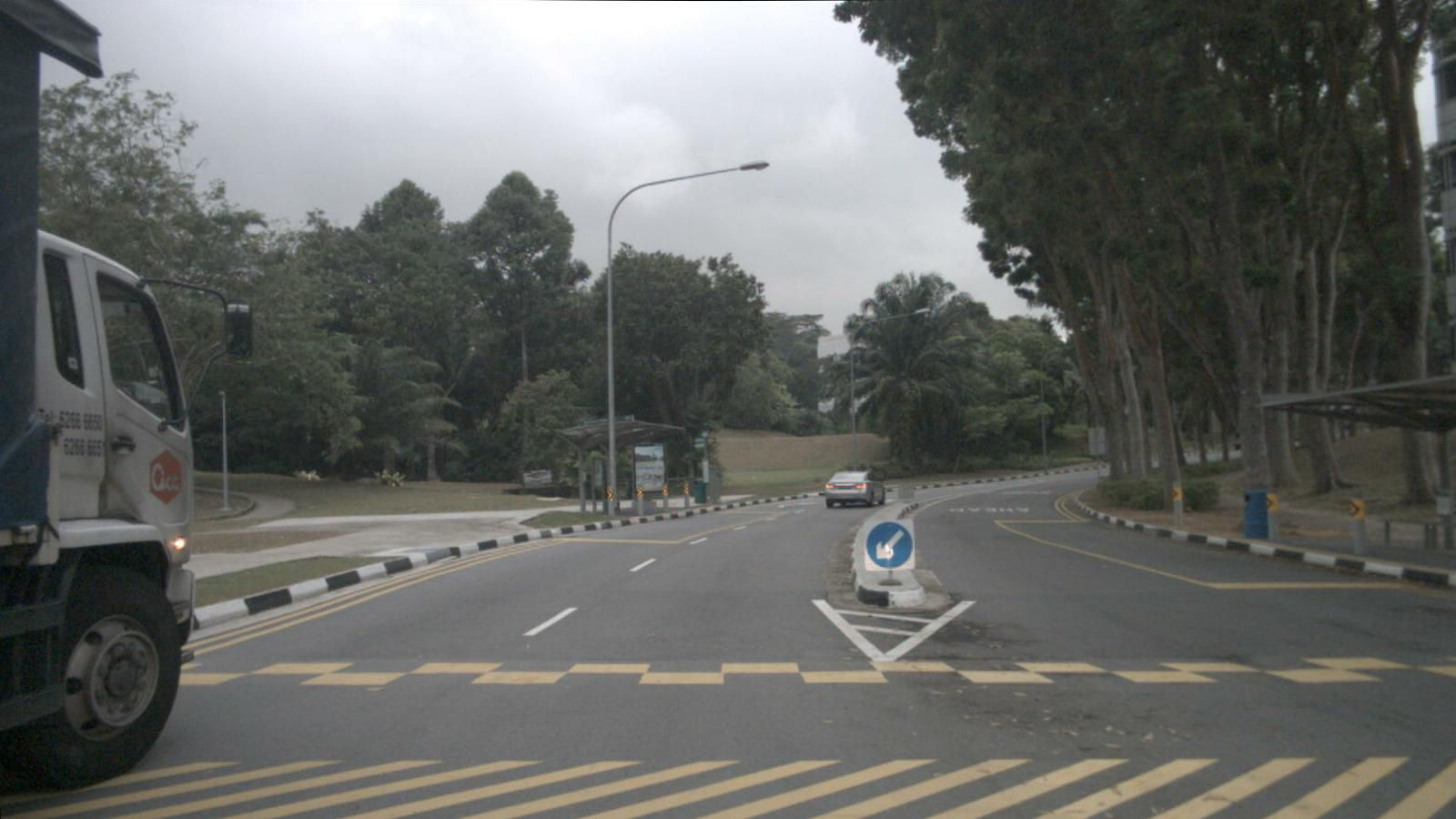} & \includegraphics[width=0.33\linewidth]{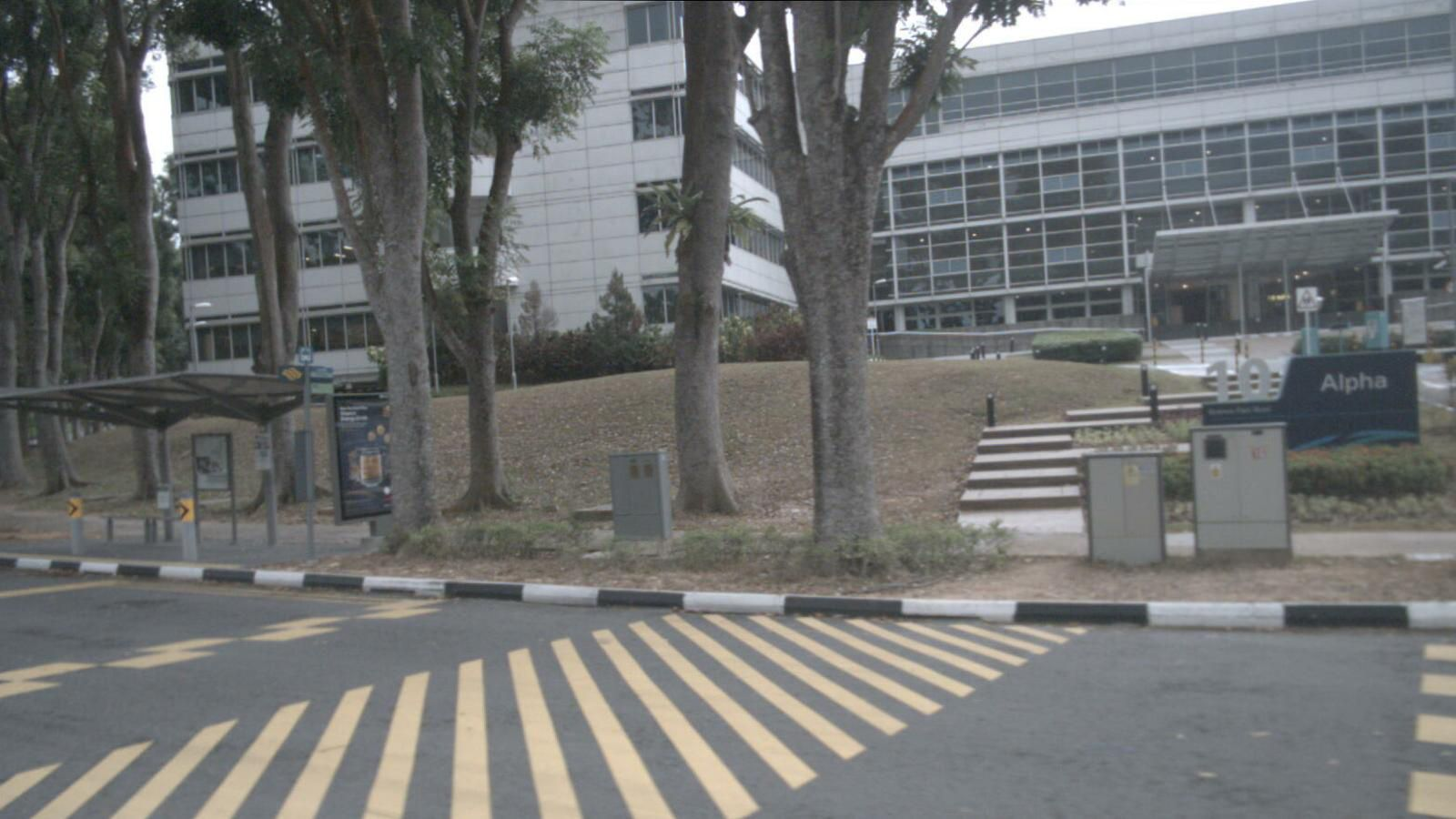} \\
            \includegraphics[width=0.33\linewidth]{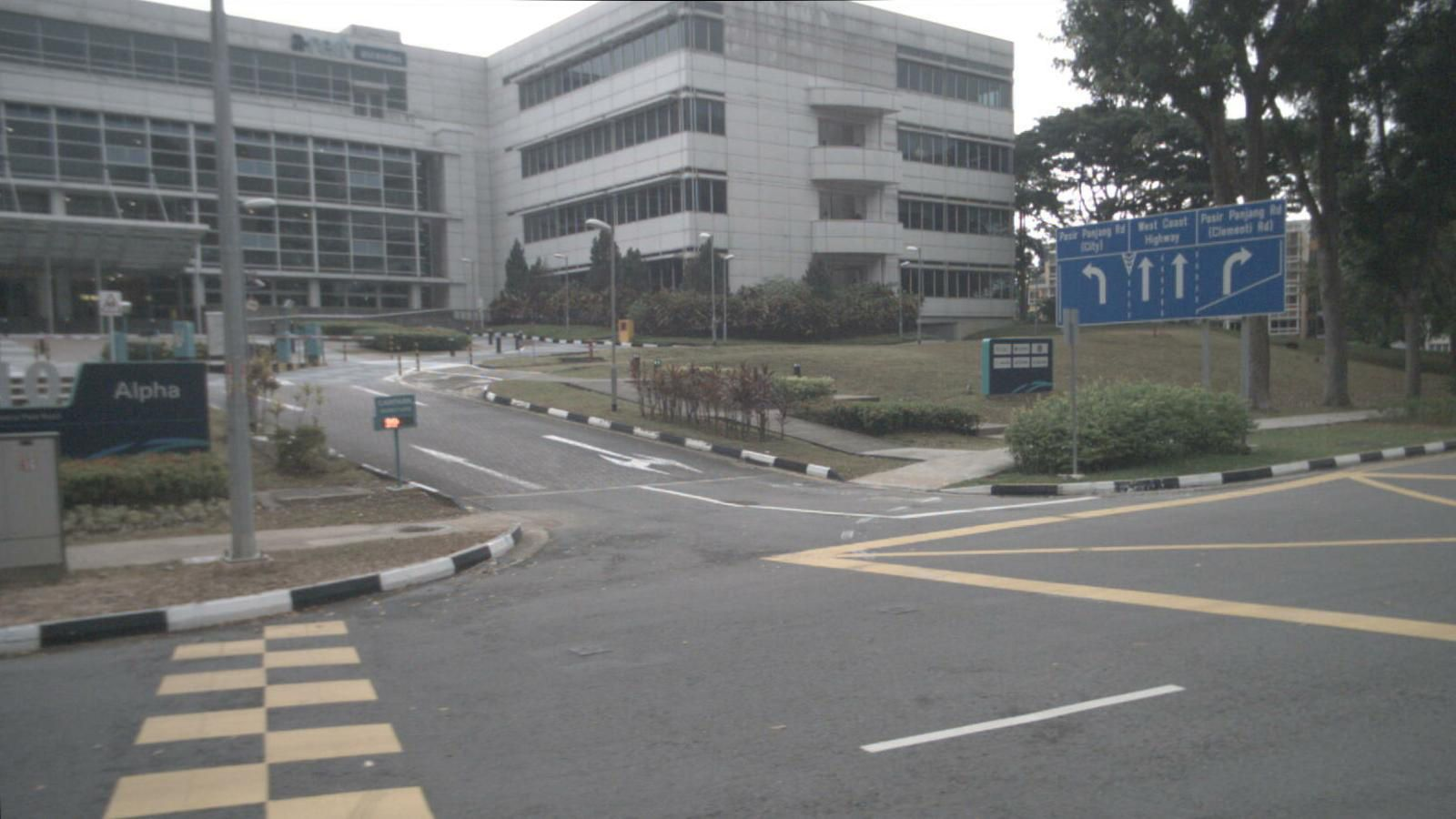} & \includegraphics[width=0.33\linewidth]{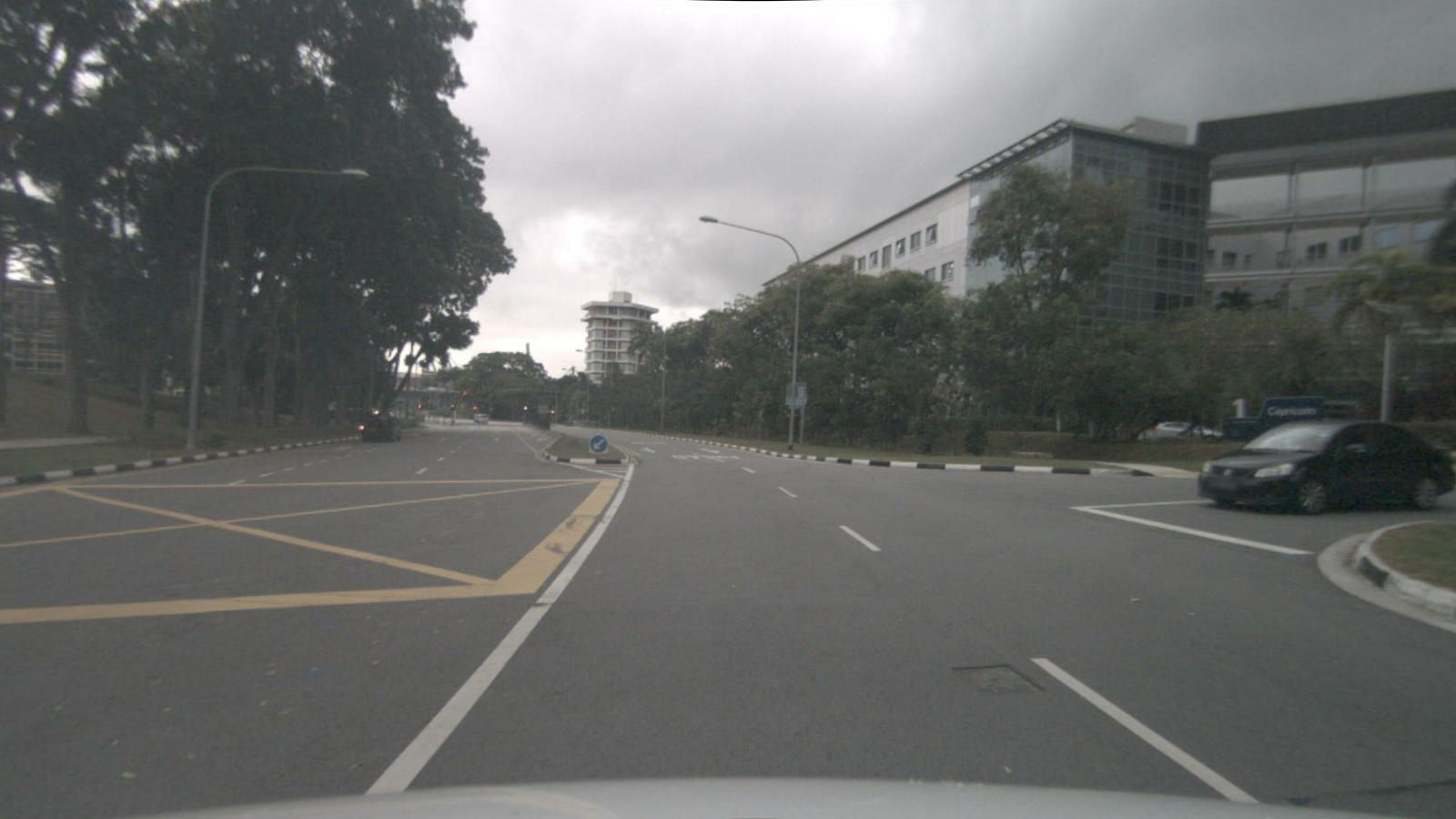} & \includegraphics[width=0.33\linewidth]{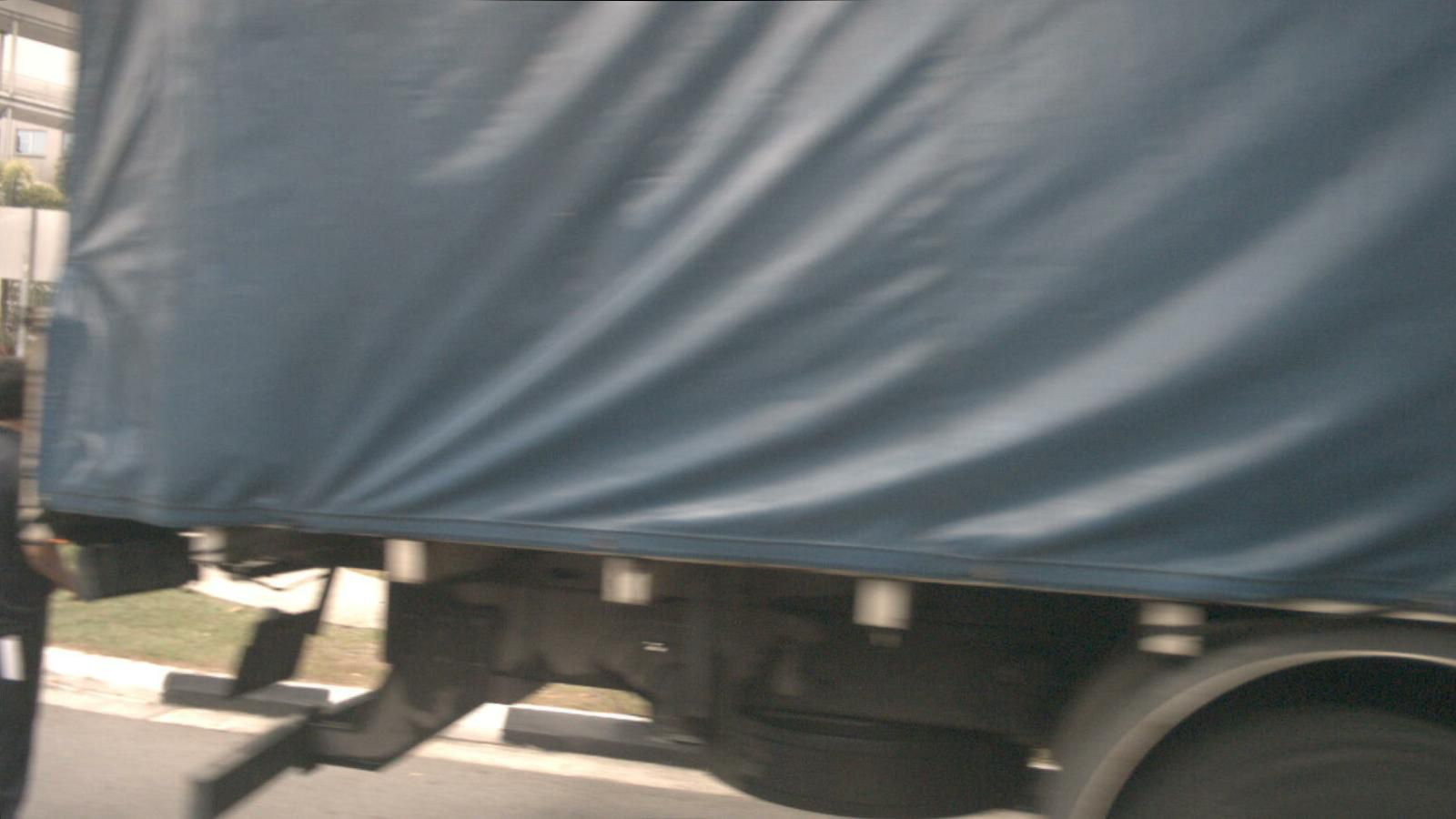} \\
        \end{tabular}
        \vspace*{\fill} & \fbox{\includegraphics[width=1\linewidth]{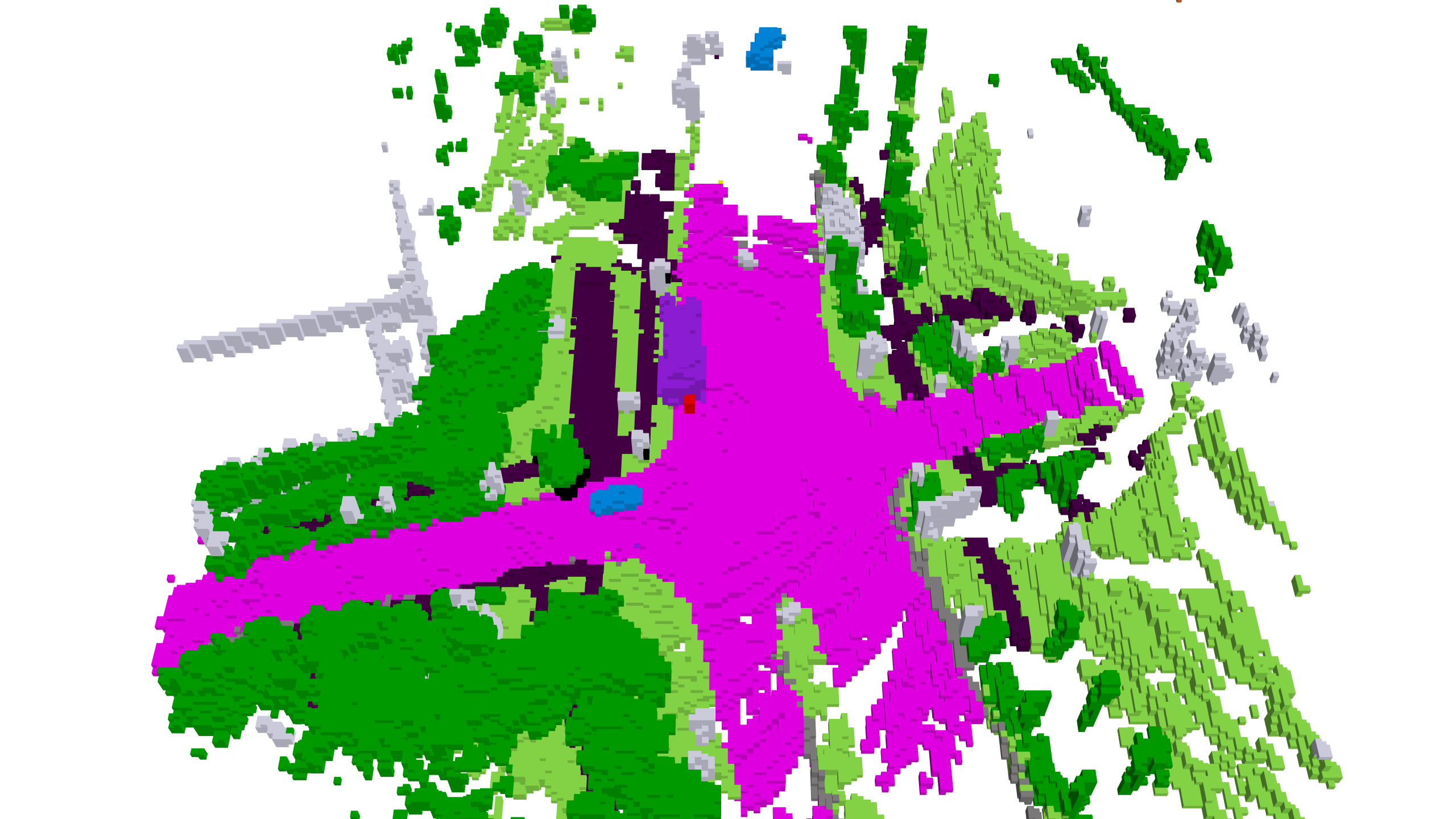}} & \fbox{\includegraphics[width=1\linewidth]{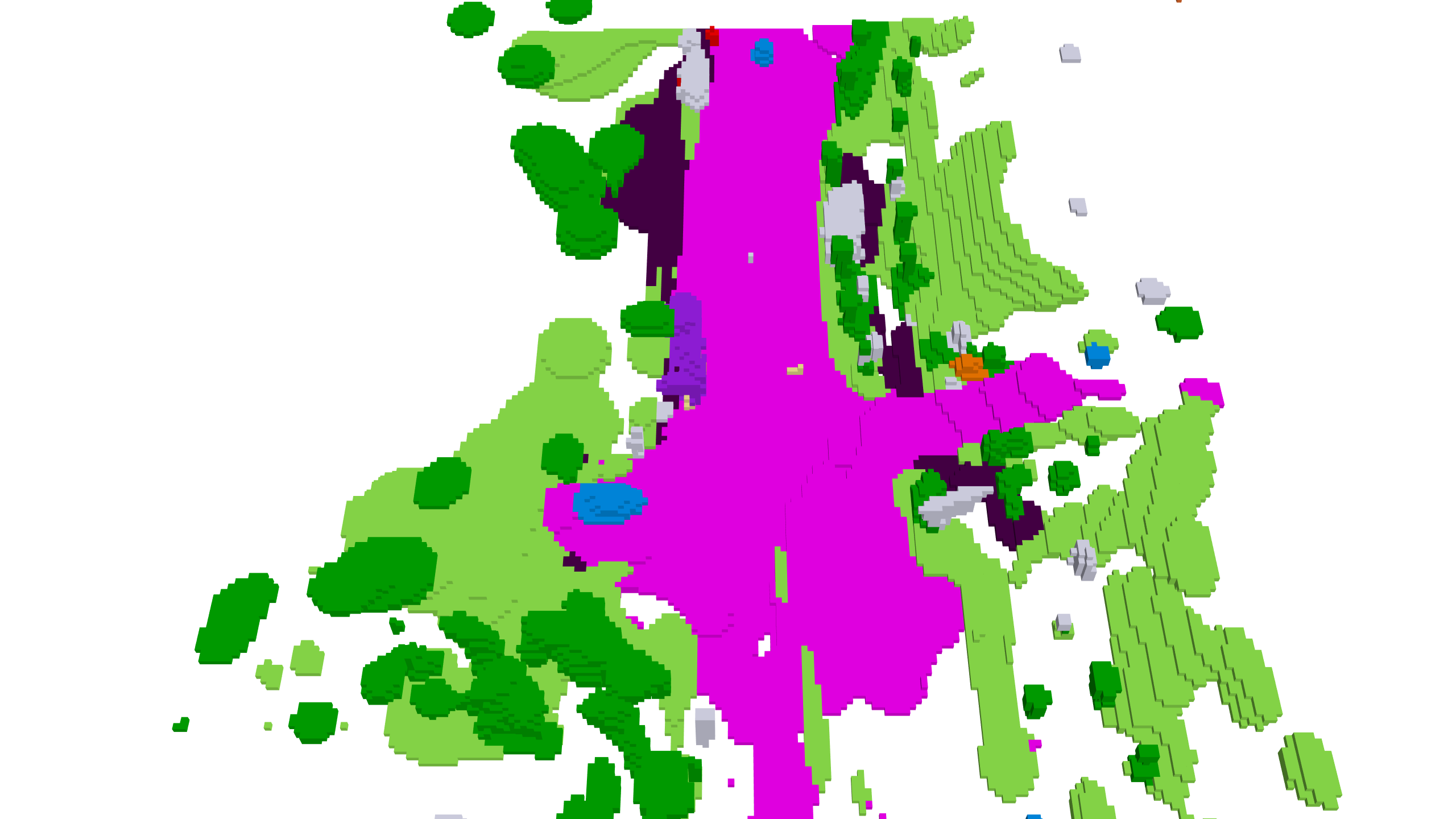}} & \fbox{\includegraphics[width=1\linewidth]{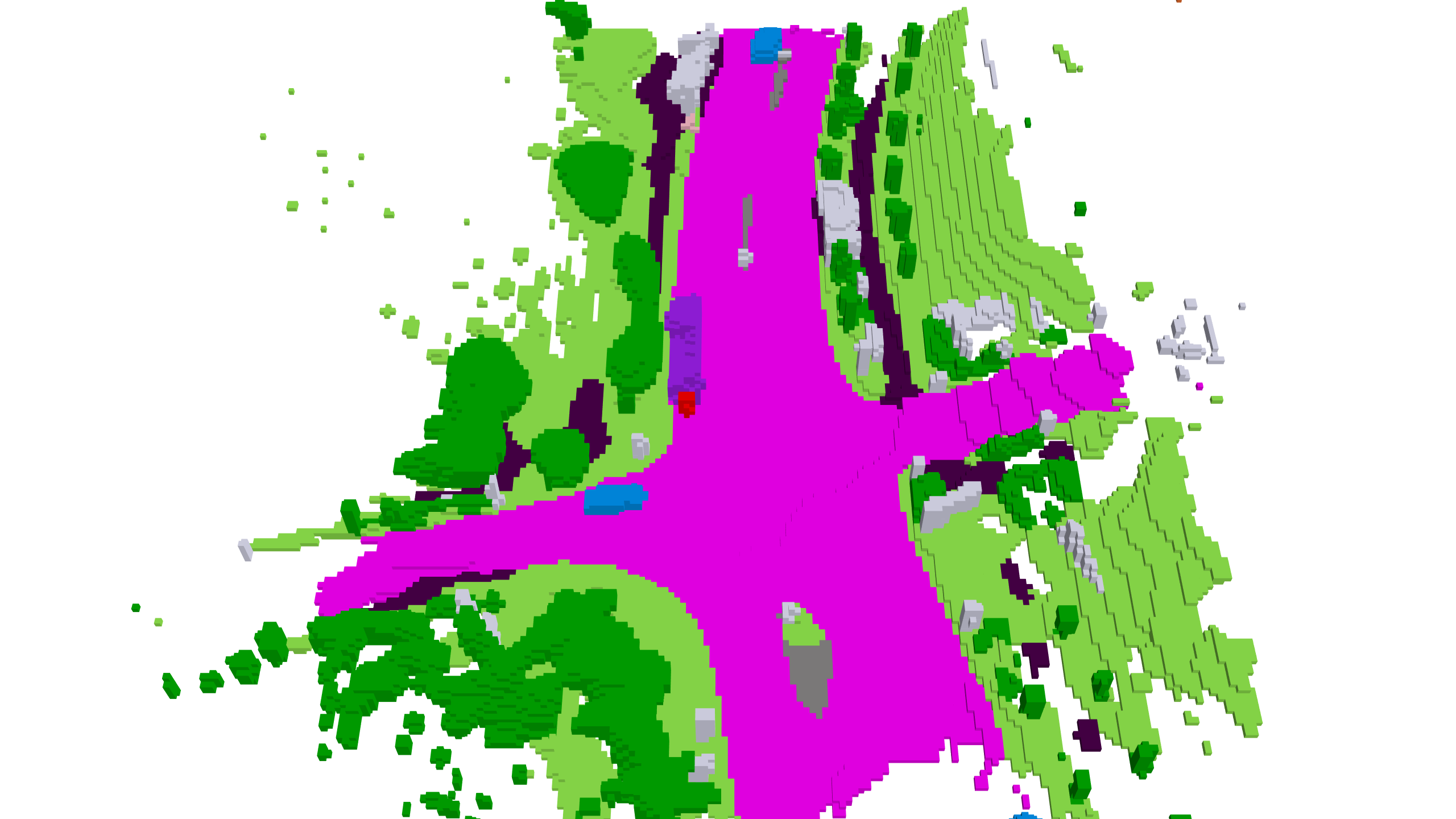}} \\
    \end{tabular}
    \caption{\textbf{Occupancy visualizations on nuScenes.} Our model is able to predict both comprehensive and realistic semantic 3D occupancy.}
    \label{fig:occupancy_visualization}
\end{figure*}

The analysis of the visualization results further underscores the benefits of multi-sensor fusion for 3D semantic occupancy prediction. In the Figure \ref{fig:occupancy_visualization}, the model incorporating camera, LiDAR, and radar data exhibits significantly better prediction results, particularly in regions further away from the ego vehicle, compared to the camera-only model. The improved performance of the model using all sensors in distant regions suggests that the LiDAR and radar data effectively compensate for the limitations of the camera-only approach in these areas, leading to a more comprehensive and accurate understanding of the 3D scene geometry. Figure \ref{fig:occupancy_visualization_all} demonstrates the improvement in the allocation of the Gaussians when using the full sensor suite.
Figure \ref{fig:occupancy_visualization_night}, comparing the performance of camera-only, camera+radar, and camera+LiDAR models in nighttime scenarios, clearly illustrates the critical role of additional sensors in overcoming the challenges posed by low-light conditions. The inclusion of both radar and LiDAR data leads to substantial improvements in the model's ability to perceive the environment under these conditions, as vision-centric approaches are known to perform poorly in nighttime scenarios due to the sensitivity of cameras to varying and limited illumination. The addition of LiDAR's 3D spatial awareness or radar's ability to detect dynamic objects and measure velocity even in low visibility contributes to a more robust perception system for autonomous vehicles operating at night. The most noticeable improvement is in predicting distant dynamic objects and reducing hallucinations of close objects.
\section{Conclusion}
\label{sec:conclusion}

In this paper, we introduced GaussianFusionOcc, a novel and seamless sensor fusion framework for 3D semantic occupancy prediction that leverages the efficiency and flexibility of the 3D Gaussian representation. As the first framework to extend Gaussian-based modeling to multi-sensor input, it addresses critical limitations of grid-based methods, achieving state-of-the-art performance (30.37 mIoU and 45.20 IoU) while significantly reducing memory usage, number of parameters, and latency. The proposed modality-agnostic Gaussian encoder and fusion mechanism enable efficient integration of camera, LiDAR, and radar data, exploiting their complementary strengths. The experiments demonstrate the model's resilience in adverse weather and nighttime conditions, achieving significantly better results compared to previous state-of-the-art methods.

Limitations:

While the framework excels in dynamic urban environments, its reliance on predefined number of Gaussians limits adaptability in extremly sparse scenarios where some of the allocated Gaussians may remain unused and opaque. The framework might benefit from dynamic pruning and densification by removing underutilized Gaussians and introuducing new ones in regions requiring higher detail. Another limitation can arise from the inherent sparsity of radar data which might not always provide substantial improvements for dense prediction tasks, requiring further research to effectively fuse its features.

{
    \small
    \bibliographystyle{unsrt}
    \bibliography{main}

\begin{thebibliography}{10}

\bibitem{ming2024occfusion}
Zhenxing Ming, Julie~Stephany Berrio, Mao Shan, and Stewart Worrall.
\newblock Occfusion: Multi-sensor fusion framework for 3d semantic occupancy prediction.
\newblock {\em IEEE Transactions on Intelligent Vehicles}, 2024.

\bibitem{lu2024octreeocc}
Yuhang Lu, Xinge Zhu, Tai Wang, and Yuexin Ma.
\newblock Octreeocc: Efficient and multi-granularity occupancy prediction using octree queries.
\newblock {\em Advances in Neural Information Processing Systems}, 37:79618--79641, 2024.

\bibitem{caesar2020nuscenes}
Holger Caesar, Varun Bankiti, Alex~H Lang, Sourabh Vora, Venice~Erin Liong, Qiang Xu, Anush Krishnan, Yu~Pan, Giancarlo Baldan, and Oscar Beijbom.
\newblock nuscenes: A multimodal dataset for autonomous driving.
\newblock In {\em Proceedings of the IEEE/CVF conference on computer vision and pattern recognition}, pages 11621--11631, 2020.

\bibitem{wang2023openoccupancy}
Xiaofeng Wang, Zheng Zhu, Wenbo Xu, Yunpeng Zhang, Yi~Wei, Xu~Chi, Yun Ye, Dalong Du, Jiwen Lu, and Xingang Wang.
\newblock Openoccupancy: A large scale benchmark for surrounding semantic occupancy perception.
\newblock In {\em Proceedings of the IEEE/CVF International Conference on Computer Vision}, pages 17850--17859, 2023.

\bibitem{li2023voxformer}
Yiming Li, Zhiding Yu, Christopher Choy, Chaowei Xiao, Jose~M Alvarez, Sanja Fidler, Chen Feng, and Anima Anandkumar.
\newblock Voxformer: Sparse voxel transformer for camera-based 3d semantic scene completion.
\newblock In {\em Proceedings of the IEEE/CVF conference on computer vision and pattern recognition}, pages 9087--9098, 2023.

\bibitem{cao2022monoscene}
Anh-Quan Cao and Raoul De~Charette.
\newblock Monoscene: Monocular 3d semantic scene completion.
\newblock In {\em Proceedings of the IEEE/CVF Conference on Computer Vision and Pattern Recognition}, pages 3991--4001, 2022.

\bibitem{jiang2024symphonize}
Haoyi Jiang, Tianheng Cheng, Naiyu Gao, Haoyang Zhang, Tianwei Lin, Wenyu Liu, and Xinggang Wang.
\newblock Symphonize 3d semantic scene completion with contextual instance queries.
\newblock In {\em Proceedings of the IEEE/CVF Conference on Computer Vision and Pattern Recognition}, pages 20258--20267, 2024.

\bibitem{miao2023occdepth}
Ruihang Miao, Weizhou Liu, Mingrui Chen, Zheng Gong, Weixin Xu, Chen Hu, and Shuchang Zhou.
\newblock Occdepth: A depth-aware method for 3d semantic scene completion.
\newblock {\em arXiv preprint arXiv:2302.13540}, 2023.

\bibitem{wei2023surroundocc}
Yi~Wei, Linqing Zhao, Wenzhao Zheng, Zheng Zhu, Jie Zhou, and Jiwen Lu.
\newblock Surroundocc: Multi-camera 3d occupancy prediction for autonomous driving.
\newblock In {\em Proceedings of the IEEE/CVF International Conference on Computer Vision}, pages 21729--21740, 2023.

\bibitem{zhang2023occformer}
Yunpeng Zhang, Zheng Zhu, and Dalong Du.
\newblock Occformer: Dual-path transformer for vision-based 3d semantic occupancy prediction.
\newblock In {\em Proceedings of the IEEE/CVF International Conference on Computer Vision}, pages 9433--9443, 2023.

\bibitem{huang2023tri}
Yuanhui Huang, Wenzhao Zheng, Yunpeng Zhang, Jie Zhou, and Jiwen Lu.
\newblock Tri-perspective view for vision-based 3d semantic occupancy prediction.
\newblock In {\em Proceedings of the IEEE/CVF conference on computer vision and pattern recognition}, pages 9223--9232, 2023.

\bibitem{huang2024gaussianformer}
Yuanhui Huang, Wenzhao Zheng, Yunpeng Zhang, Jie Zhou, and Jiwen Lu.
\newblock Gaussianformer: Scene as gaussians for vision-based 3d semantic occupancy prediction.
\newblock In {\em European Conference on Computer Vision}, pages 376--393. Springer, 2024.

\bibitem{huang2024probabilistic}
Yuanhui Huang, Amonnut Thammatadatrakoon, Wenzhao Zheng, Yunpeng Zhang, Dalong Du, and Jiwen Lu.
\newblock Probabilistic gaussian superposition for efficient 3d occupancy prediction.
\newblock {\em arXiv preprint arXiv:2412.04384}, 2024.

\bibitem{liu2023bevfusion}
Zhijian Liu, Haotian Tang, Alexander Amini, Xinyu Yang, Huizi Mao, Daniela~L Rus, and Song Han.
\newblock Bevfusion: Multi-task multi-sensor fusion with unified bird's-eye view representation.
\newblock In {\em 2023 IEEE international conference on robotics and automation (ICRA)}, pages 2774--2781. IEEE, 2023.

\bibitem{kerbl3Dgaussians}
Bernhard Kerbl, Georgios Kopanas, Thomas Leimk{\"u}hler, and George Drettakis.
\newblock 3d gaussian splatting for real-time radiance field rendering.
\newblock {\em ACM Transactions on Graphics}, 42(4), July 2023.

\bibitem{yang2024deformable}
Ziyi Yang, Xinyu Gao, Wen Zhou, Shaohui Jiao, Yuqing Zhang, and Xiaogang Jin.
\newblock Deformable 3d gaussians for high-fidelity monocular dynamic scene reconstruction.
\newblock In {\em Proceedings of the IEEE/CVF conference on computer vision and pattern recognition}, pages 20331--20341, 2024.

\bibitem{li2024bevformer}
Zhiqi Li, Wenhai Wang, Hongyang Li, Enze Xie, Chonghao Sima, Tong Lu, Qiao Yu, and Jifeng Dai.
\newblock Bevformer: learning bird's-eye-view representation from lidar-camera via spatiotemporal transformers.
\newblock {\em IEEE Transactions on Pattern Analysis and Machine Intelligence}, 2024.

\bibitem{li2023fb}
Zhiqi Li, Zhiding Yu, David Austin, Mingsheng Fang, Shiyi Lan, Jan Kautz, and Jose~M Alvarez.
\newblock Fb-occ: 3d occupancy prediction based on forward-backward view transformation.
\newblock {\em arXiv preprint arXiv:2307.01492}, 2023.

\bibitem{gan2024gaussianoccfullyselfsupervisedefficient}
Wanshui Gan, Fang Liu, Hongbin Xu, Ningkai Mo, and Naoto Yokoya.
\newblock Gaussianocc: Fully self-supervised and efficient 3d occupancy estimation with gaussian splatting, 2024.

\bibitem{tang2024sparseocc}
Pin Tang, Zhongdao Wang, Guoqing Wang, Jilai Zheng, Xiangxuan Ren, Bailan Feng, and Chao Ma.
\newblock Sparseocc: Rethinking sparse latent representation for vision-based semantic occupancy prediction.
\newblock In {\em Proceedings of the IEEE/CVF Conference on Computer Vision and Pattern Recognition}, pages 15035--15044, 2024.

\bibitem{wang2024opus}
Jiabao Wang, Zhaojiang Liu, Qiang Meng, Liujiang Yan, Ke~Wang, Jie Yang, Wei Liu, Qibin Hou, and Ming-Ming Cheng.
\newblock Opus: occupancy prediction using a sparse set.
\newblock {\em arXiv preprint arXiv:2409.09350}, 2024.

\bibitem{xie2023sparsefusion}
Yichen Xie, Chenfeng Xu, Marie-Julie Rakotosaona, Patrick Rim, Federico Tombari, Kurt Keutzer, Masayoshi Tomizuka, and Wei Zhan.
\newblock Sparsefusion: Fusing multi-modal sparse representations for multi-sensor 3d object detection.
\newblock In {\em Proceedings of the IEEE/CVF International Conference on Computer Vision}, pages 17591--17602, 2023.

\bibitem{chen2023futr3d}
Xuanyao Chen, Tianyuan Zhang, Yue Wang, Yilun Wang, and Hang Zhao.
\newblock Futr3d: A unified sensor fusion framework for 3d detection.
\newblock In {\em proceedings of the IEEE/CVF conference on computer vision and pattern recognition}, pages 172--181, 2023.

\bibitem{kim2023craft}
Youngseok Kim, Sanmin Kim, Jun~Won Choi, and Dongsuk Kum.
\newblock Craft: Camera-radar 3d object detection with spatio-contextual fusion transformer.
\newblock In {\em Proceedings of the AAAI Conference on Artificial Intelligence}, volume~37, pages 1160--1168, 2023.

\bibitem{nabati2021centerfusion}
Ramin Nabati and Hairong Qi.
\newblock Centerfusion: Center-based radar and camera fusion for 3d object detection.
\newblock In {\em Proceedings of the IEEE/CVF winter conference on applications of computer vision}, pages 1527--1536, 2021.

\bibitem{zhu2020deformable}
Xizhou Zhu, Weijie Su, Lewei Lu, Bin Li, Xiaogang Wang, and Jifeng Dai.
\newblock Deformable detr: Deformable transformers for end-to-end object detection.
\newblock {\em arXiv preprint arXiv:2010.04159}, 2020.

\bibitem{he2016deep}
Kaiming He, Xiangyu Zhang, Shaoqing Ren, and Jian Sun.
\newblock Deep residual learning for image recognition.
\newblock In {\em Proceedings of the IEEE conference on computer vision and pattern recognition}, pages 770--778, 2016.

\bibitem{dai2017dcn}
Jifeng Dai, Haozhi Qi, Yuwen Xiong, Yi~Li, Guodong Zhang, Han Hu, and Yichen Wei.
\newblock Deformable convolutional networks.
\newblock In {\em Proceedings of the IEEE international conference on computer vision}, pages 764--773, 2017.

\bibitem{lin2017feature}
Tsung-Yi Lin, Piotr Doll{\'a}r, Ross Girshick, Kaiming He, Bharath Hariharan, and Serge Belongie.
\newblock Feature pyramid networks for object detection.
\newblock In {\em Proceedings of the IEEE conference on computer vision and pattern recognition}, pages 2117--2125, 2017.

\bibitem{zhou2018voxelnet}
Yin Zhou and Oncel Tuzel.
\newblock Voxelnet: End-to-end learning for point cloud based 3d object detection.
\newblock In {\em Proceedings of the IEEE conference on computer vision and pattern recognition}, pages 4490--4499, 2018.

\bibitem{lang2019pointpillars}
Alex~H Lang, Sourabh Vora, Holger Caesar, Lubing Zhou, Jiong Yang, and Oscar Beijbom.
\newblock Pointpillars: Fast encoders for object detection from point clouds.
\newblock In {\em Proceedings of the IEEE/CVF conference on computer vision and pattern recognition}, pages 12697--12705, 2019.

\bibitem{berman2018lovasz}
Maxim Berman, Amal~Rannen Triki, and Matthew~B Blaschko.
\newblock The lov{\'a}sz-softmax loss: A tractable surrogate for the optimization of the intersection-over-union measure in neural networks.
\newblock In {\em Proceedings of the IEEE conference on computer vision and pattern recognition}, pages 4413--4421, 2018.

\bibitem{wang2021fcos3d}
Tai Wang, Xinge Zhu, Jiangmiao Pang, and Dahua Lin.
\newblock Fcos3d: Fully convolutional one-stage monocular 3d object detection.
\newblock In {\em Proceedings of the IEEE/CVF international conference on computer vision}, pages 913--922, 2021.

\bibitem{loshchilov2017decoupled}
Ilya Loshchilov and Frank Hutter.
\newblock Decoupled weight decay regularization.
\newblock {\em arXiv preprint arXiv:1711.05101}, 2017.

\bibitem{murez2020atlas}
Zak Murez, Tarrence Van~As, James Bartolozzi, Ayan Sinha, Vijay Badrinarayanan, and Andrew Rabinovich.
\newblock Atlas: End-to-end 3d scene reconstruction from posed images.
\newblock In {\em European conference on computer vision}, pages 414--431. Springer, 2020.

\end{thebibliography}
}

\appendix
\section{Additional visualizations}  \label{sec:add_visualizations}
\begin{figure*}[htb!]
    \centering
    \centering
    \footnotesize
    \setlength{\fboxsep}{0pt}%
    \setlength{\fboxrule}{0.1pt}%
    \setlength{\tabcolsep}{1pt}
    \renewcommand{\arraystretch}{0.2}
    \begin{tabular}{>{\centering\arraybackslash}p{0.24\textwidth} >{\centering\arraybackslash}p{0.24\textwidth} >{\centering\arraybackslash}p{0.24\textwidth} >{\centering\arraybackslash}p{0.24\textwidth}}
        \textbf{Ground Truth} & \textbf{GaussianFusionOcc (C)} & \textbf{GaussianFusionOcc (C+R)} & \textbf{GaussianFusionOcc (C+L)} \\
        \fbox{\includegraphics[width=1\linewidth]{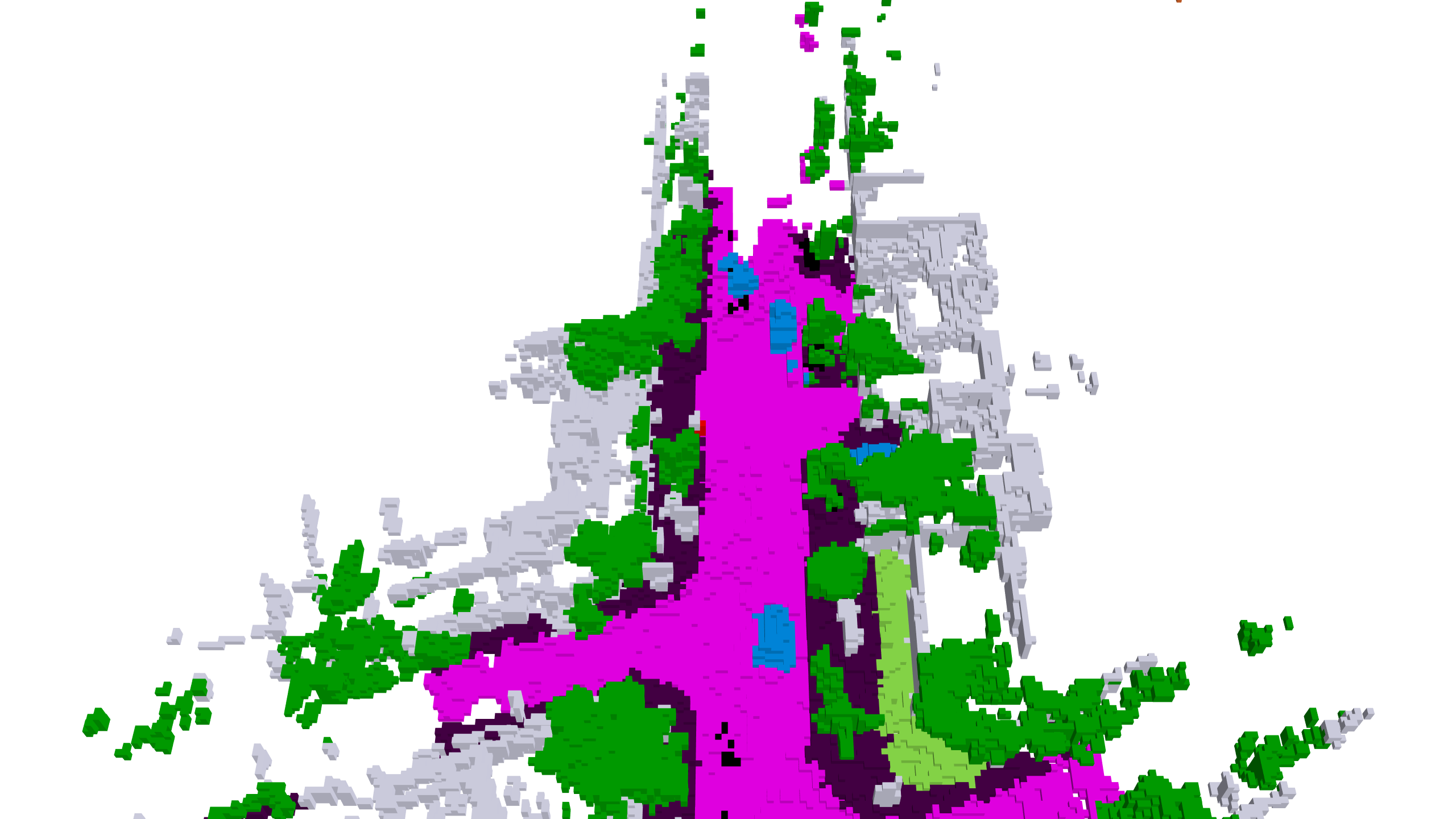}} & \fbox{\includegraphics[width=1\linewidth]{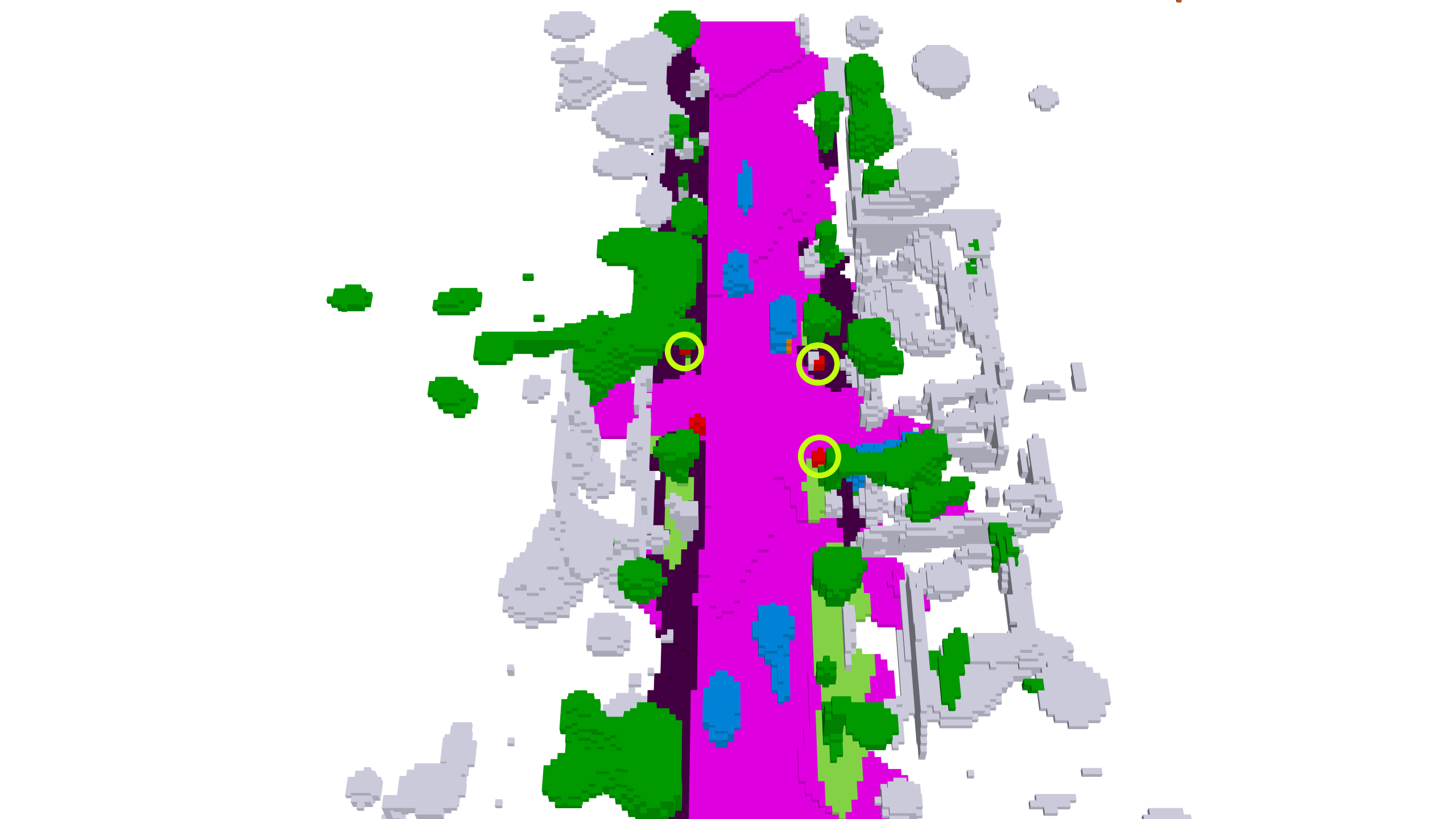}} & \fbox{\includegraphics[width=1\linewidth]{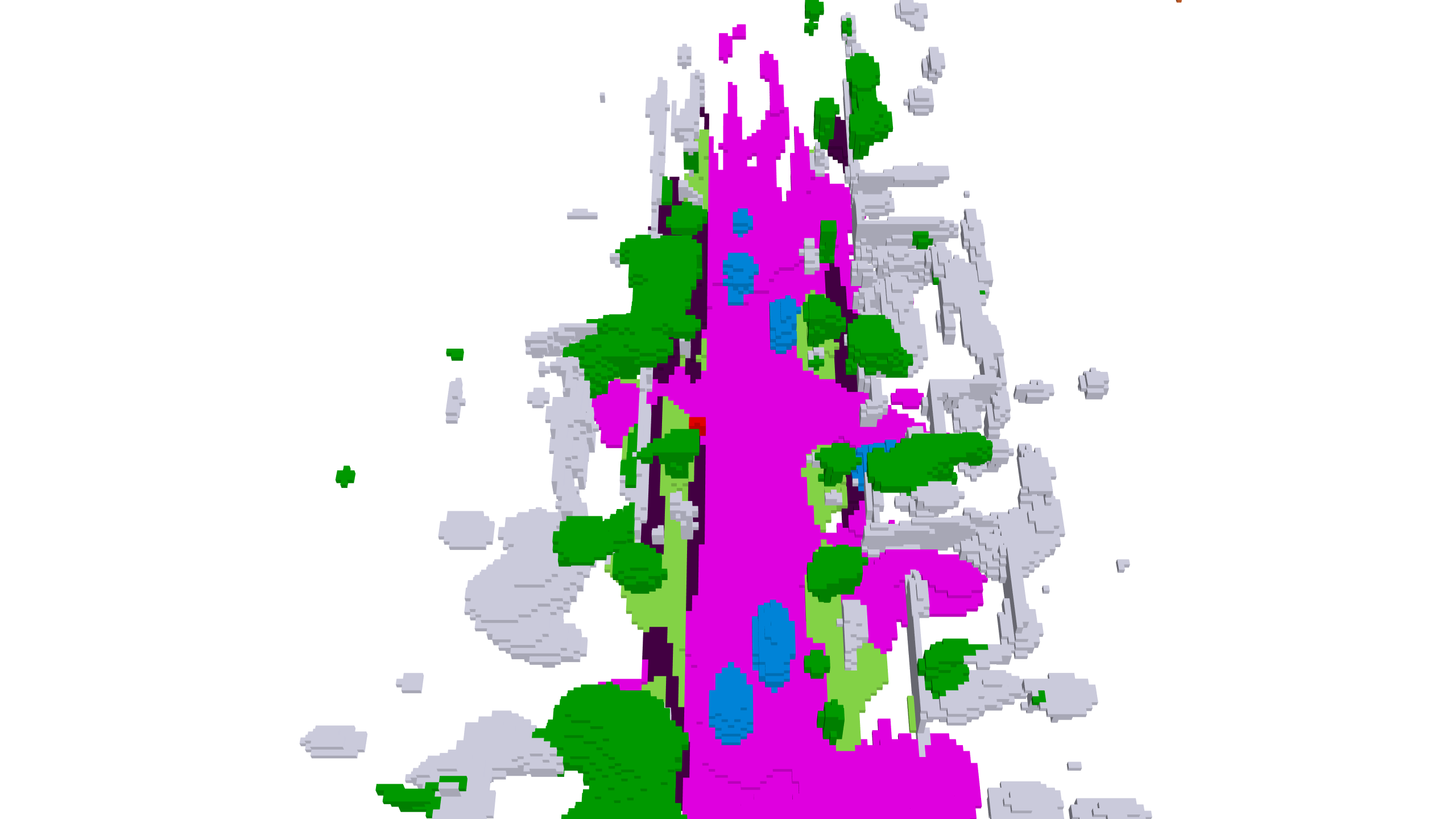}} & \fbox{\includegraphics[width=1\linewidth]{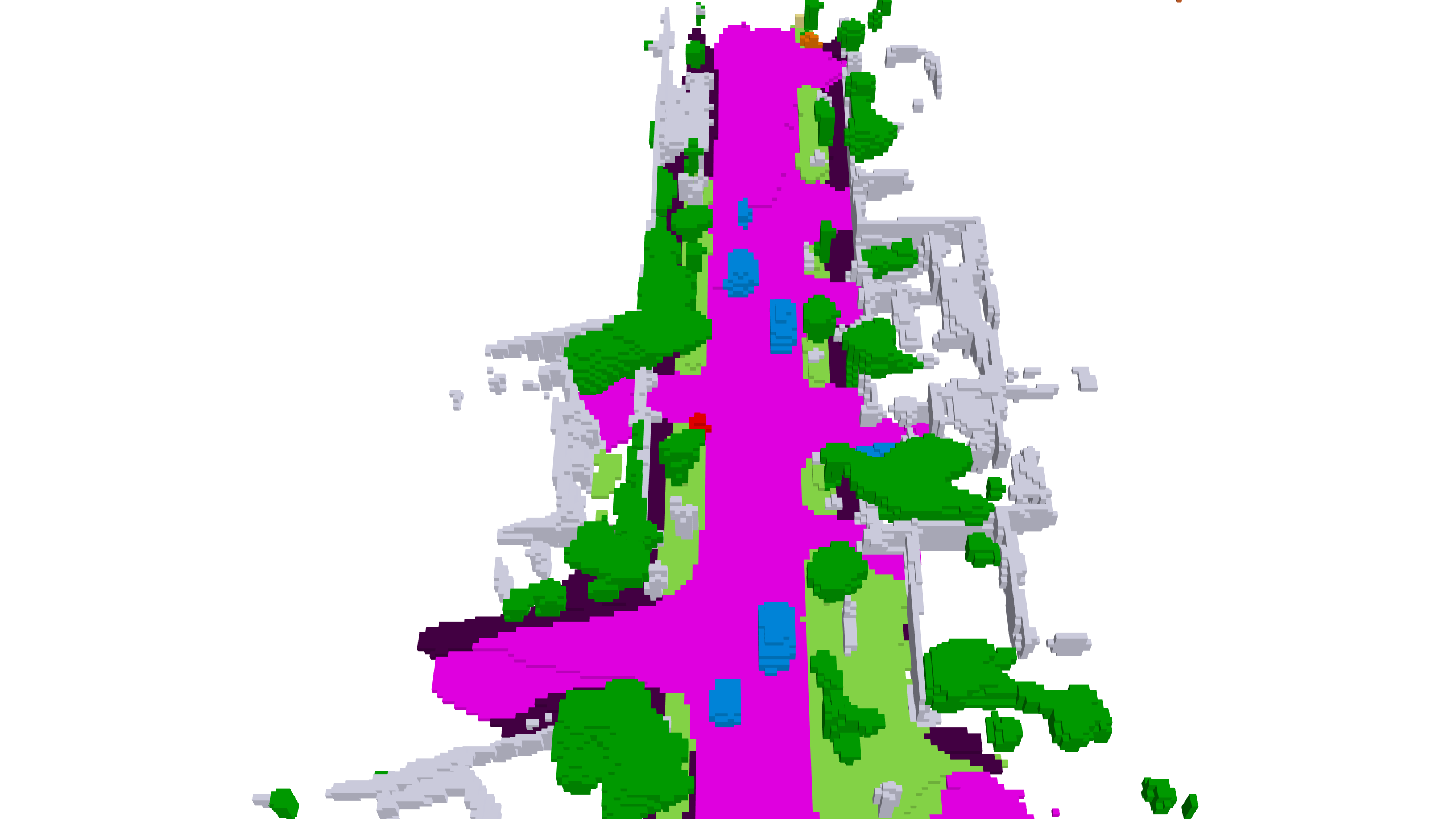}}  \\
        \fbox{\includegraphics[width=1\linewidth]{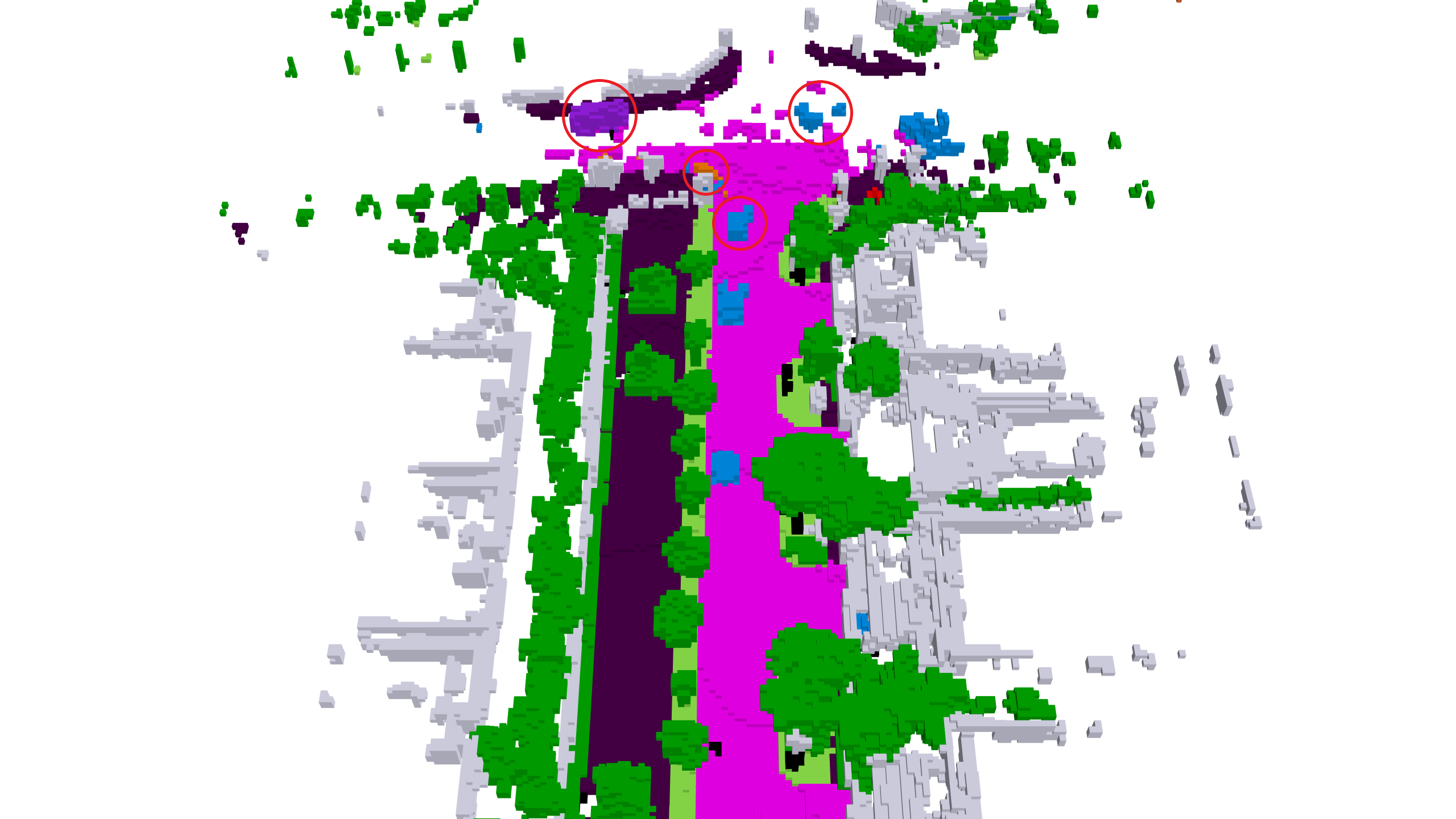}} & \fbox{\includegraphics[width=1\linewidth]{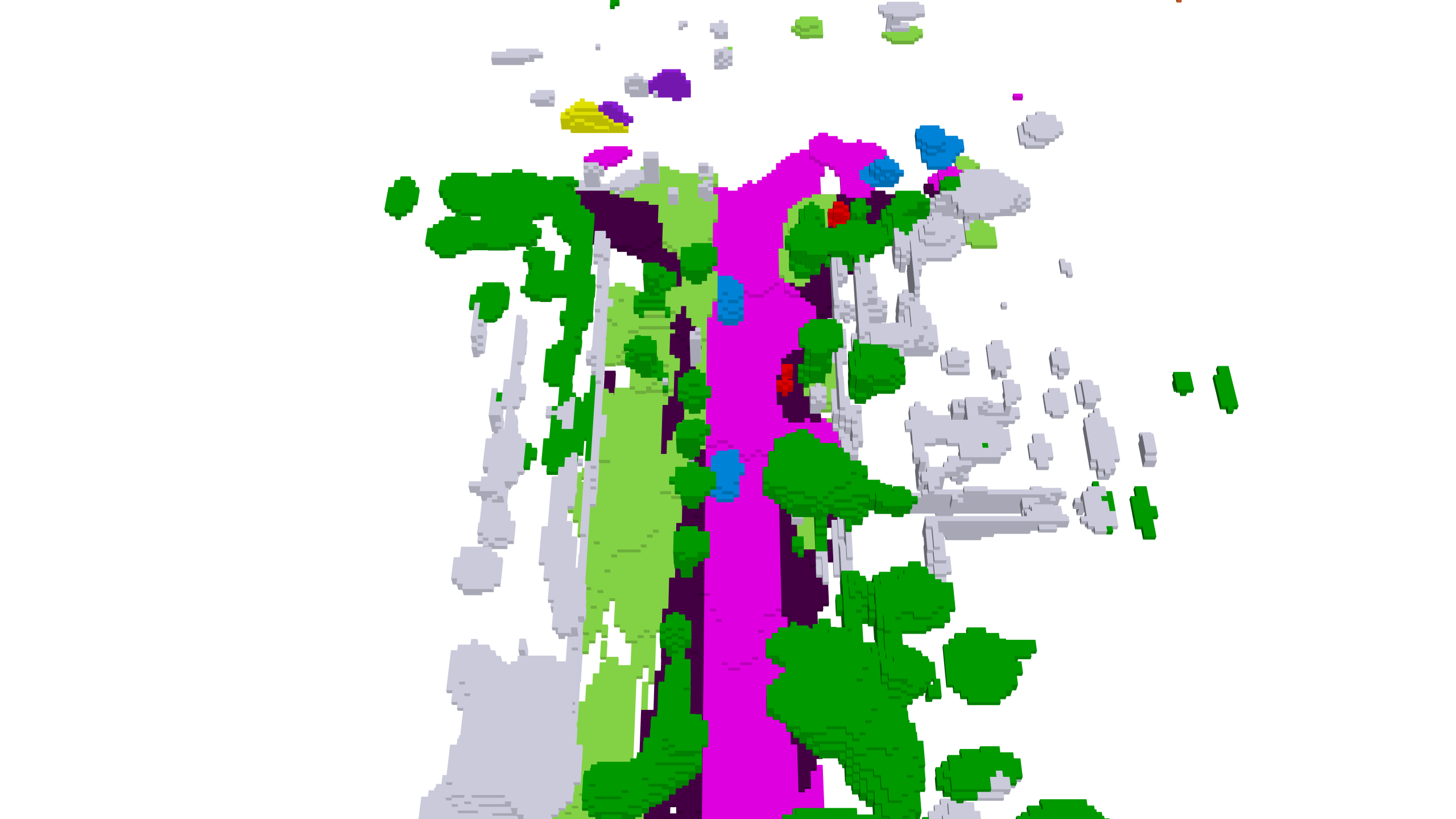}} & \fbox{\includegraphics[width=1\linewidth]{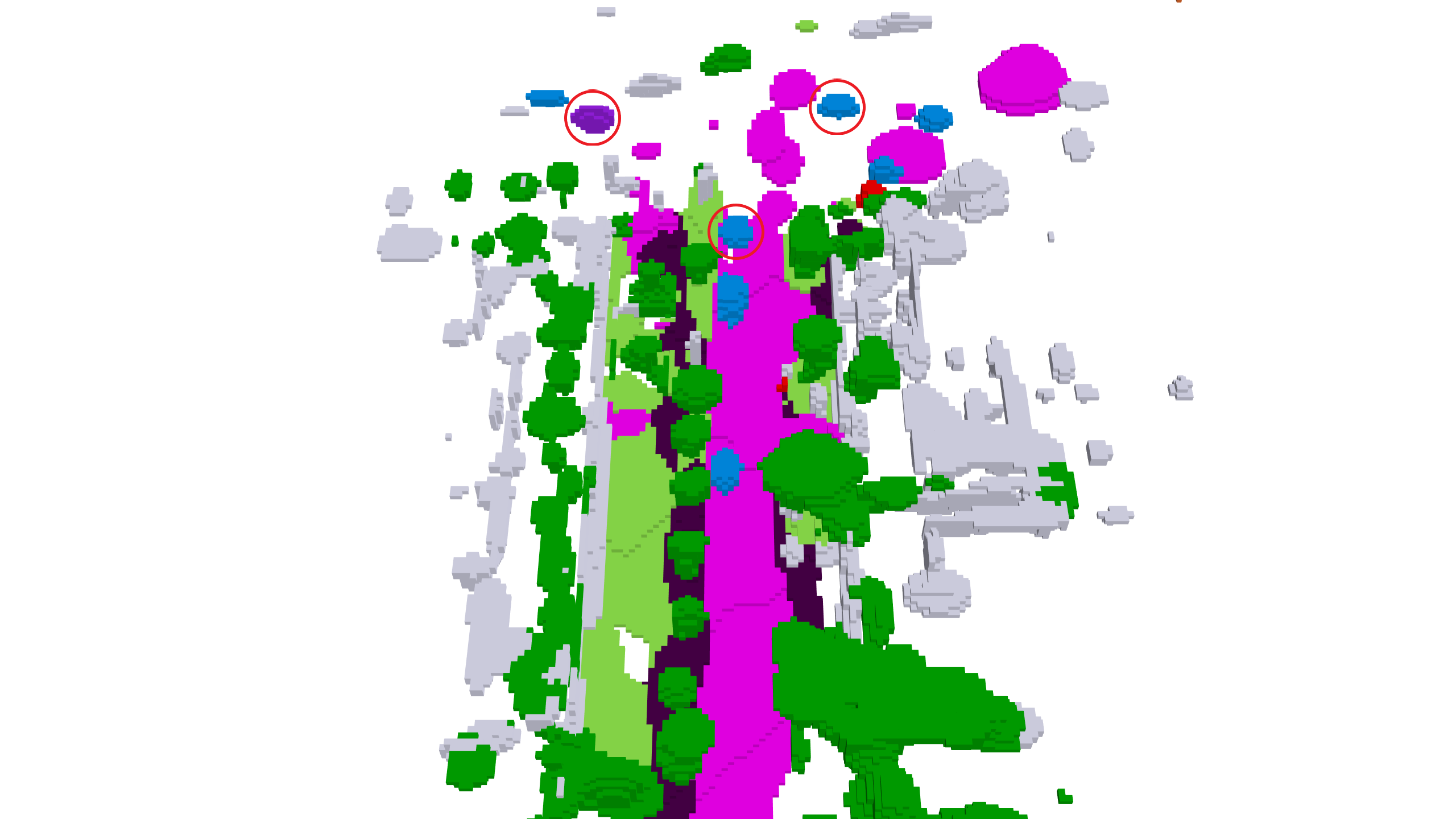}} & \fbox{\includegraphics[width=1\linewidth]{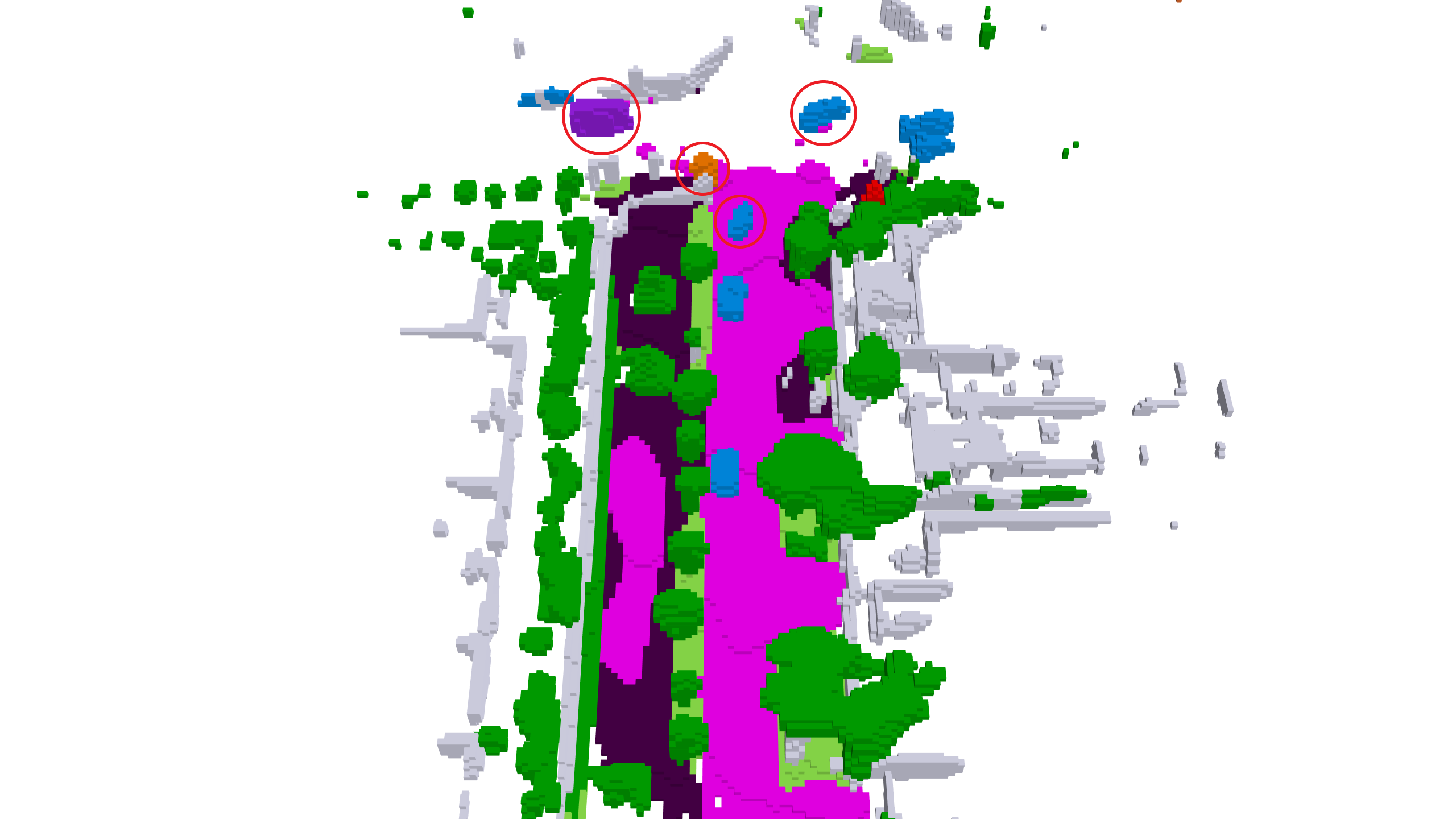}}  \\
        \fbox{\includegraphics[width=1\linewidth]{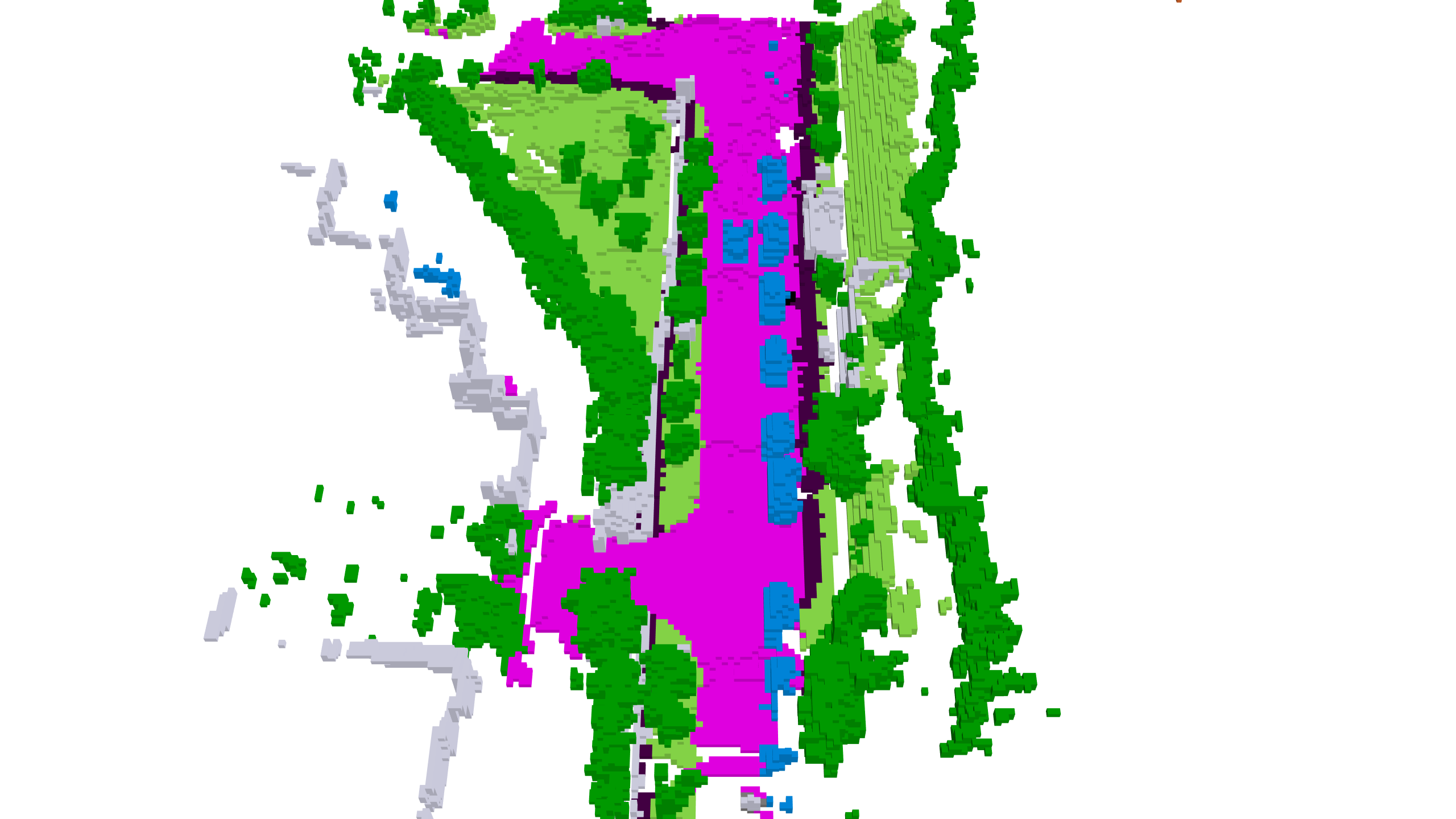}} & \fbox{\includegraphics[width=1\linewidth]{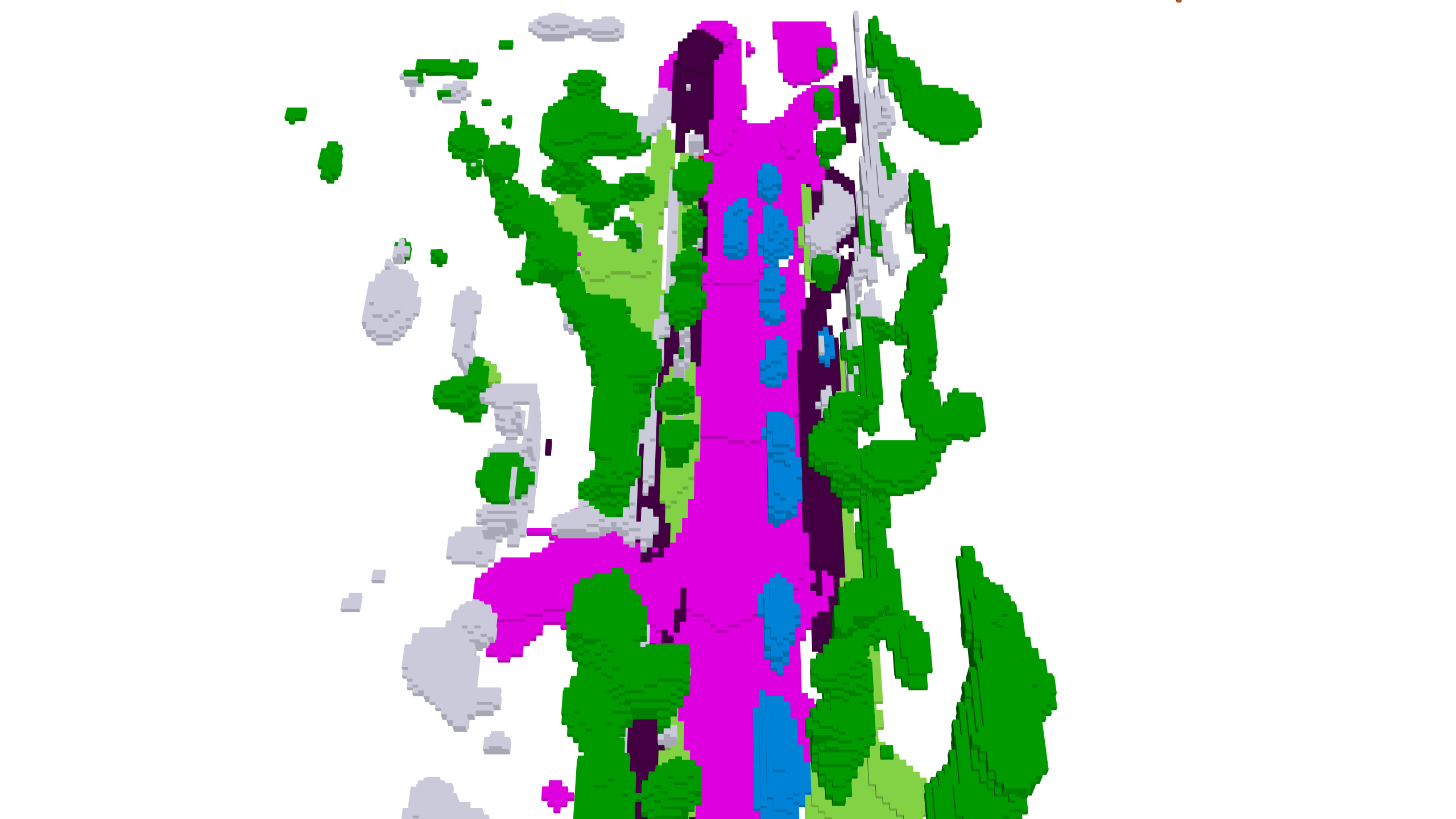}} & \fbox{\includegraphics[width=1\linewidth]{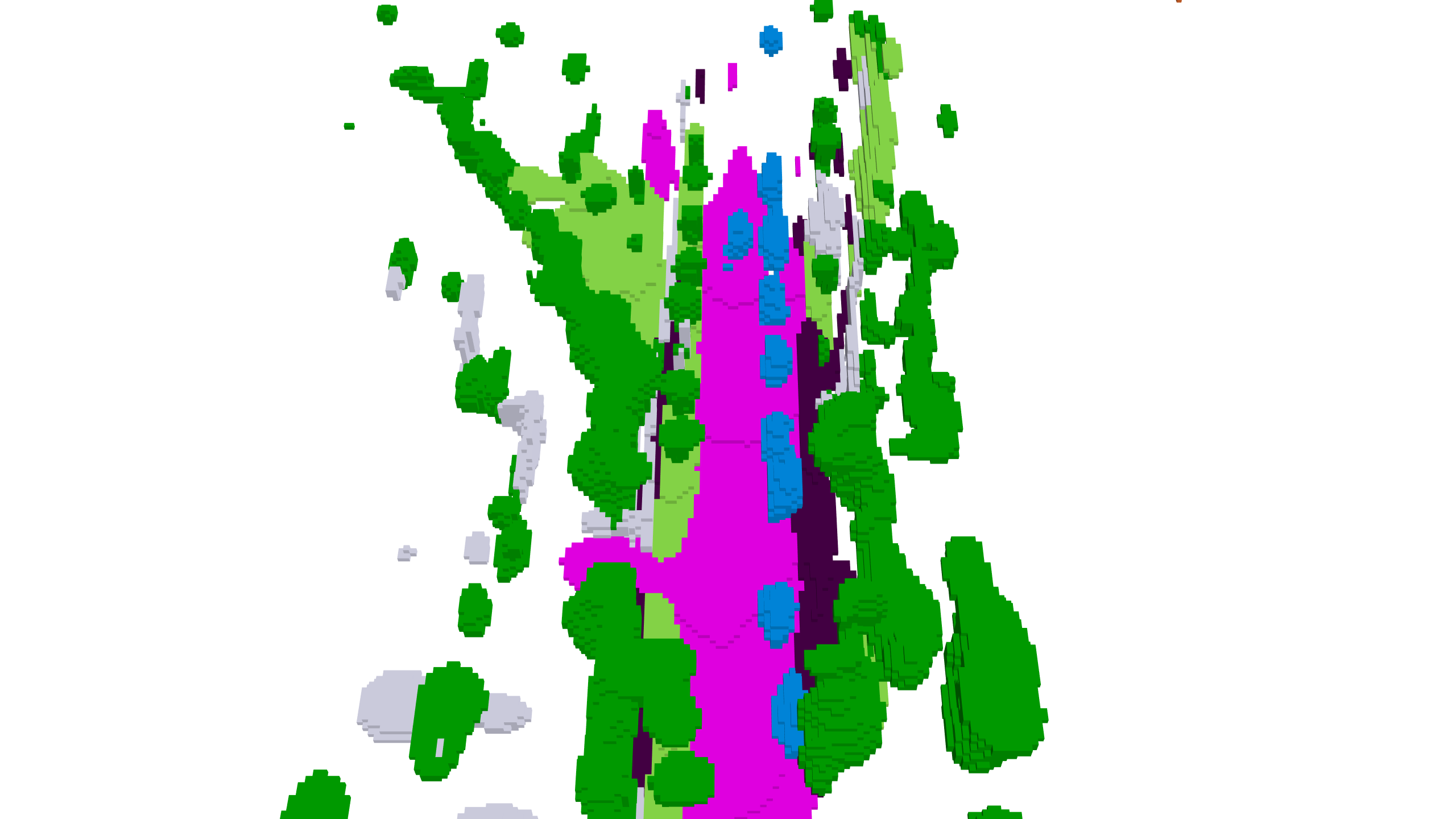}} & \fbox{\includegraphics[width=1\linewidth]{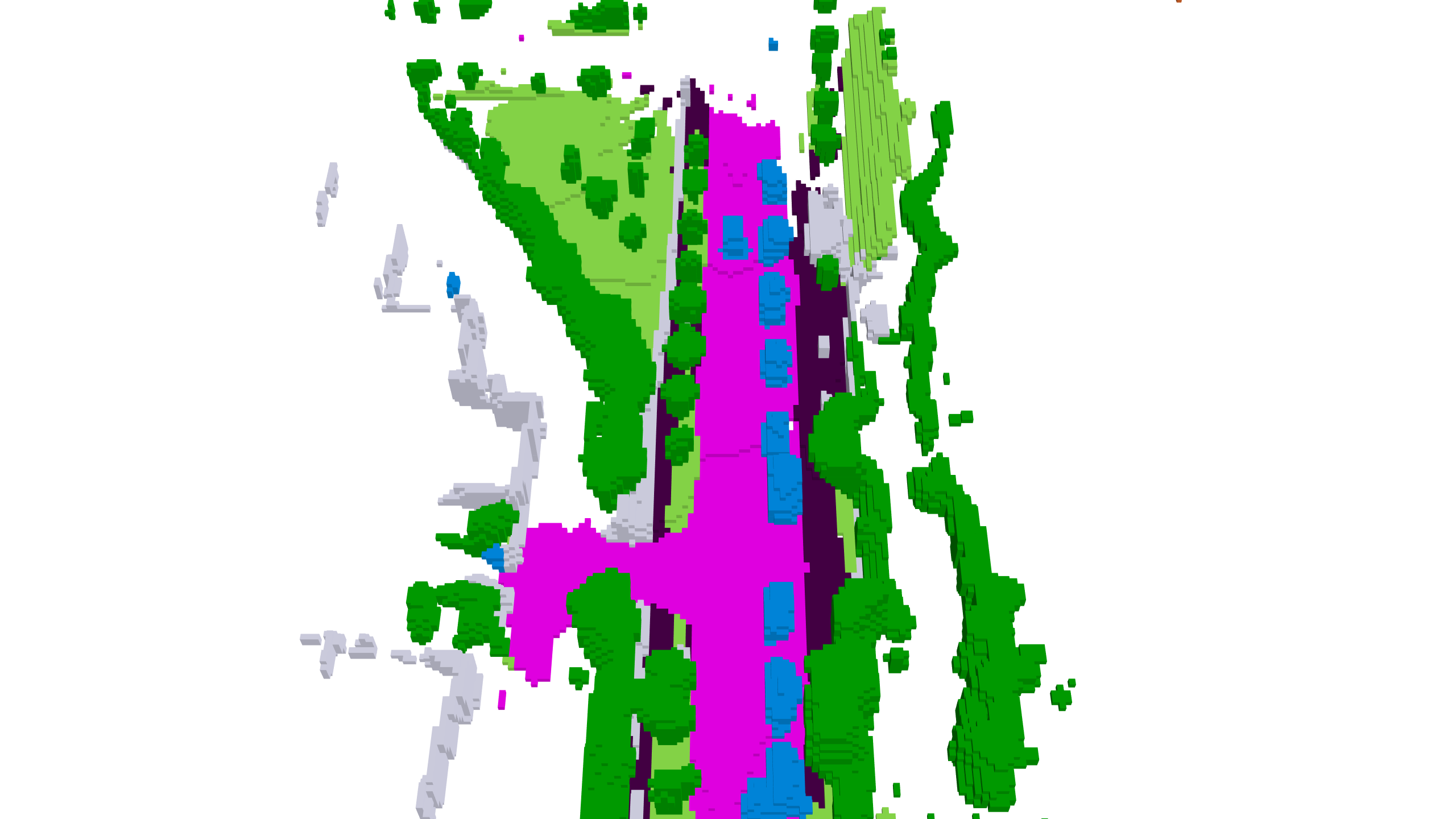}} \\
        \fbox{\includegraphics[width=1\linewidth]{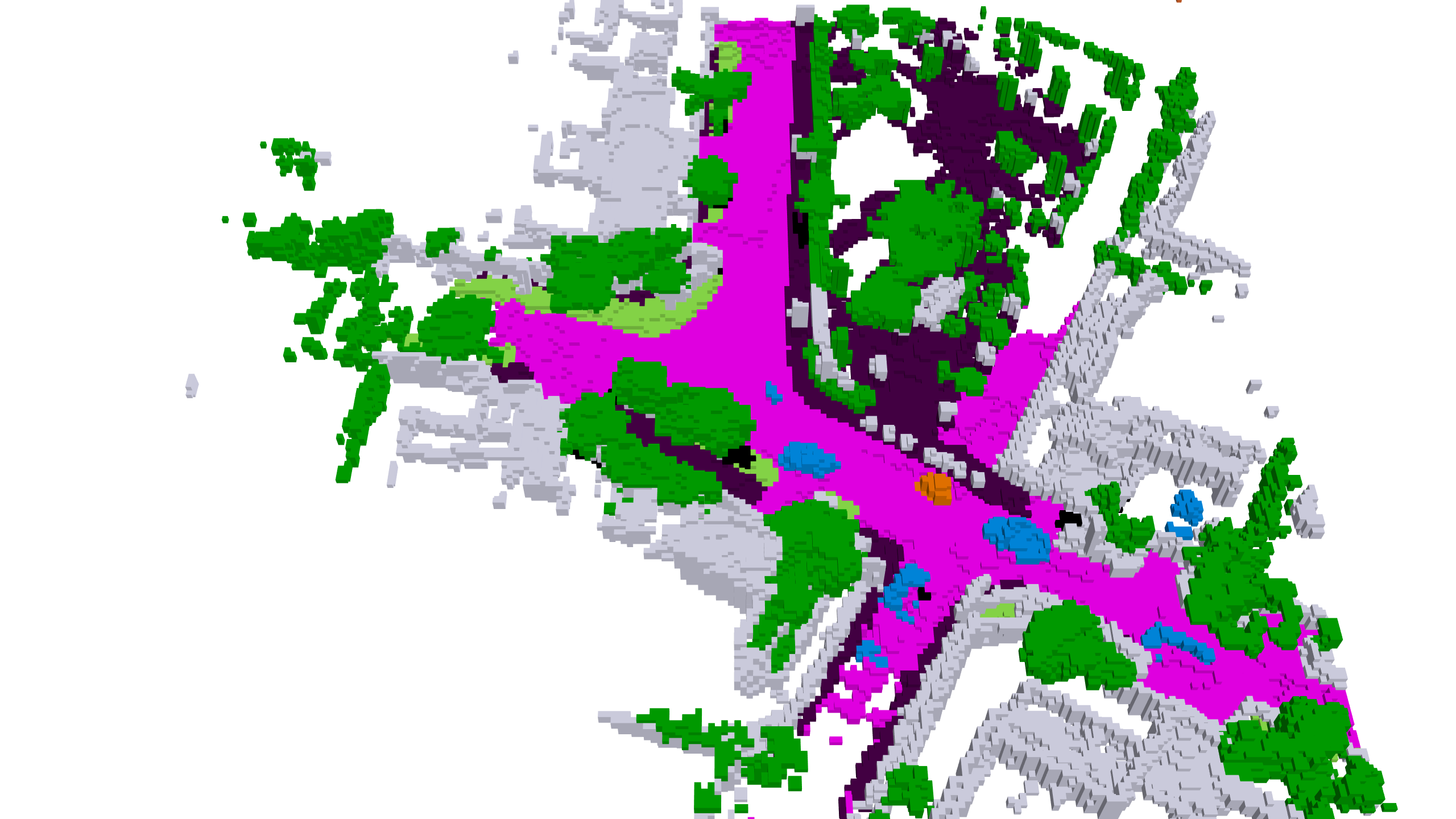}} & \fbox{\includegraphics[width=1\linewidth]{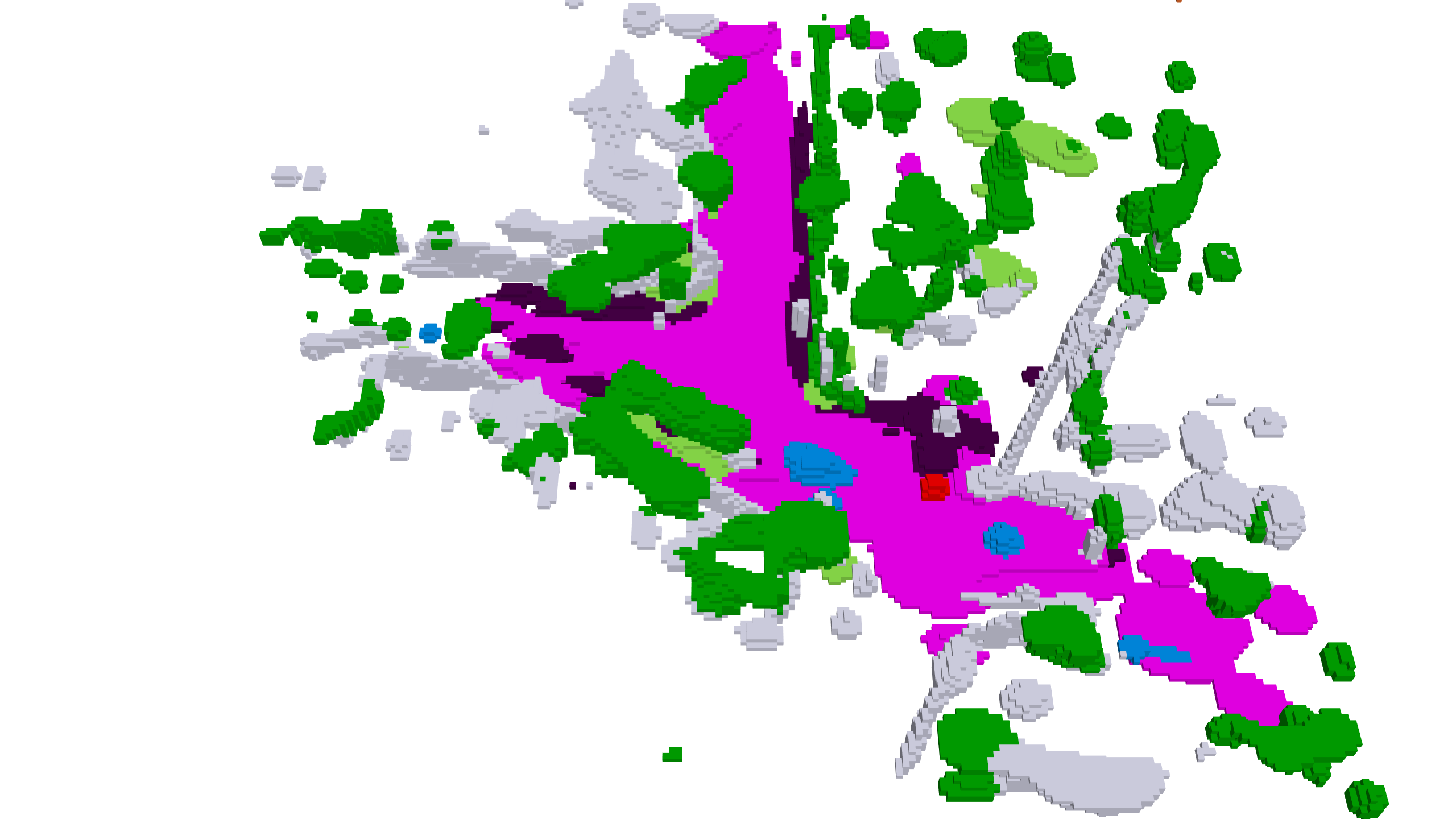}} & \fbox{\includegraphics[width=1\linewidth]{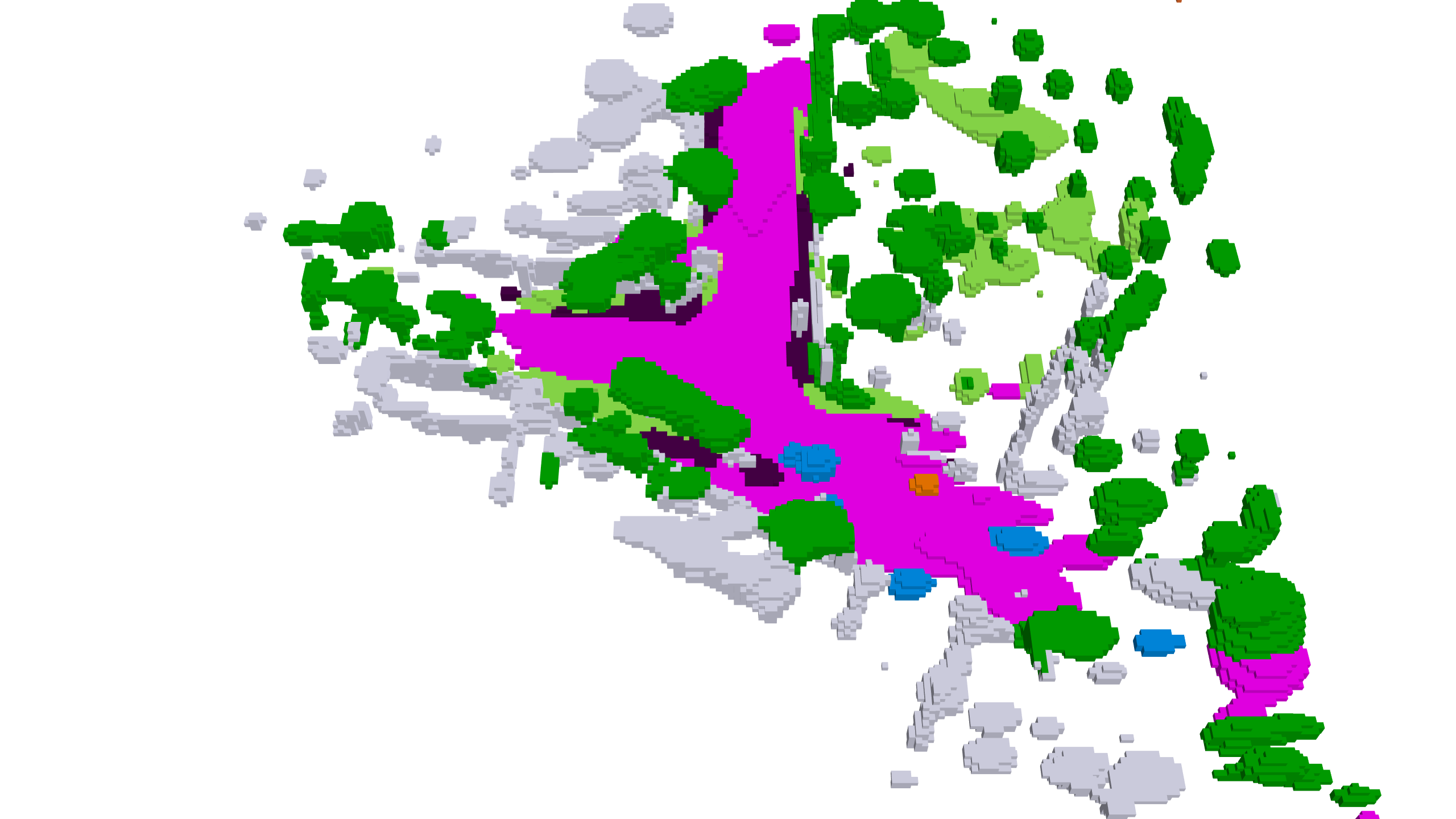}} & \fbox{\includegraphics[width=1\linewidth]{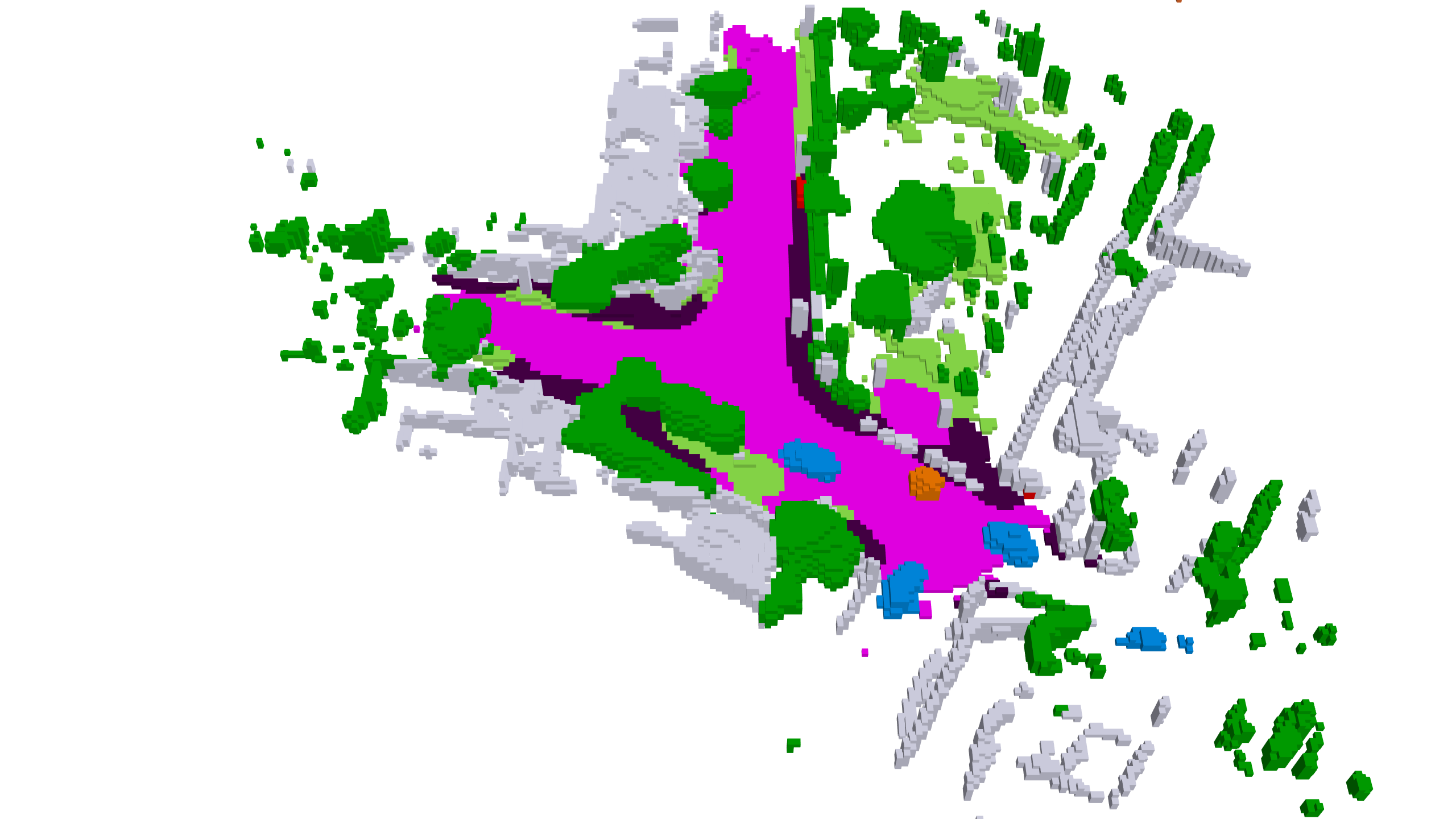}} \\
        \fbox{\includegraphics[width=1\linewidth]{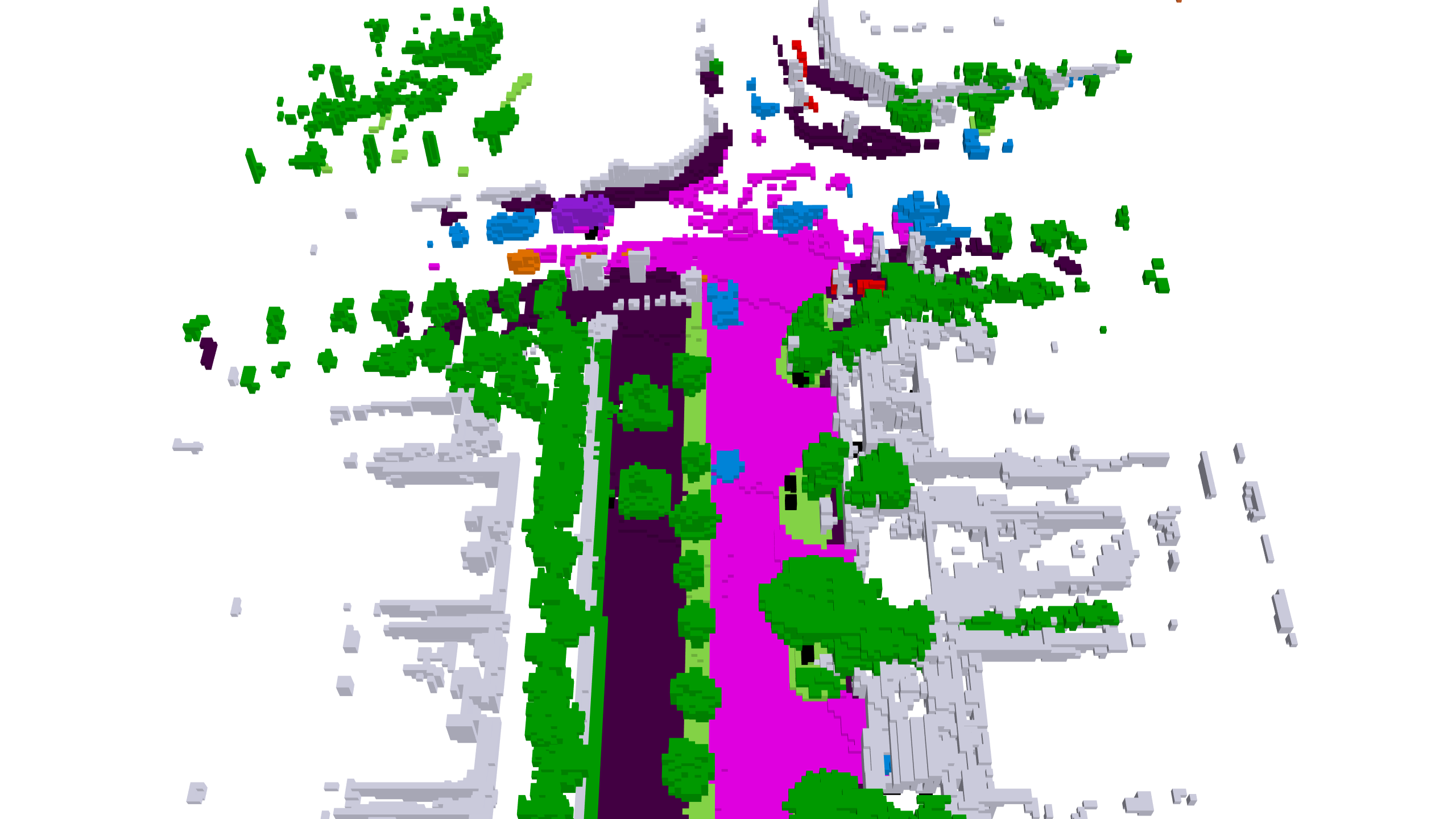}} & \fbox{\includegraphics[width=1\linewidth]{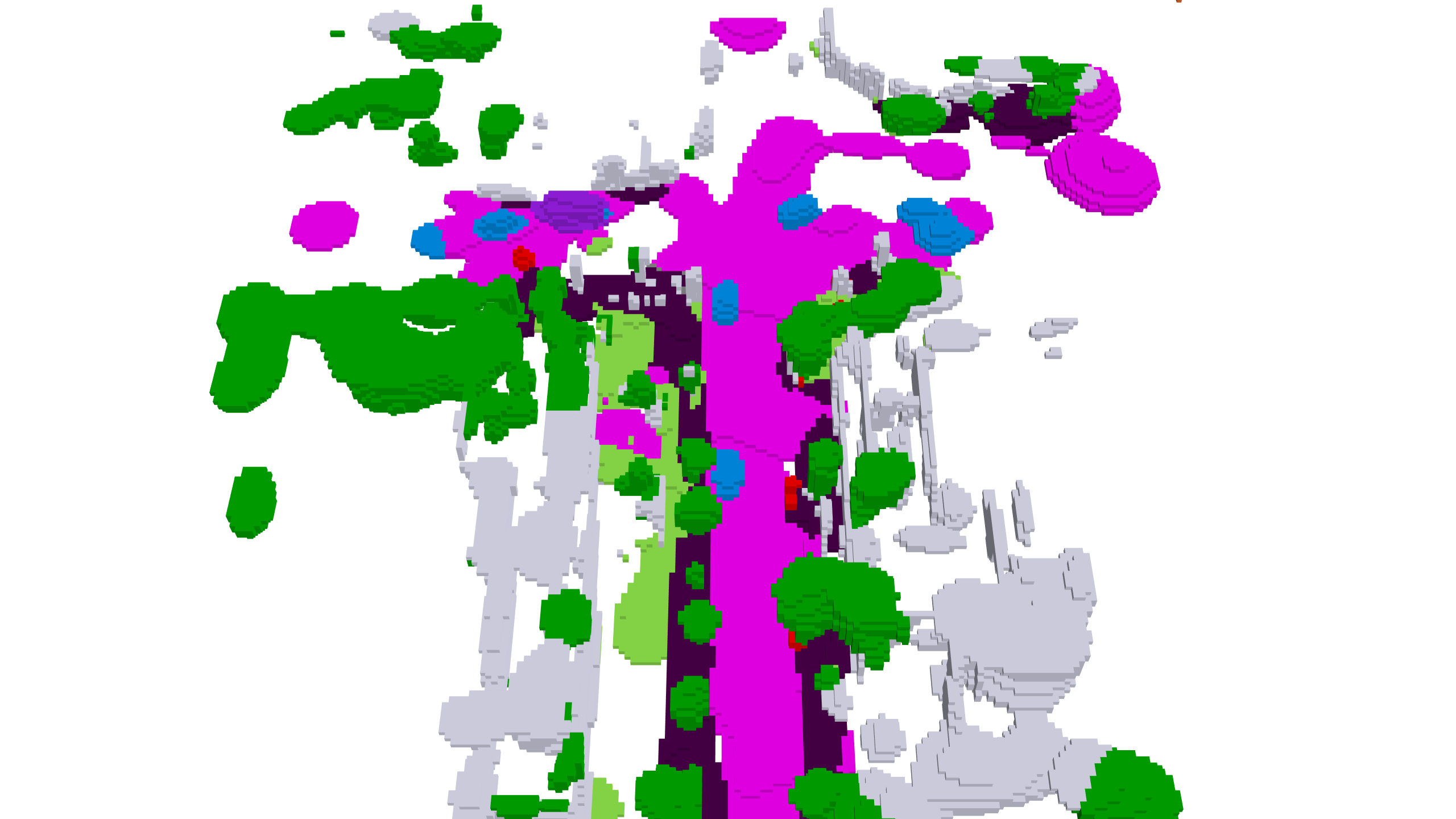}} & \fbox{\includegraphics[width=1\linewidth]{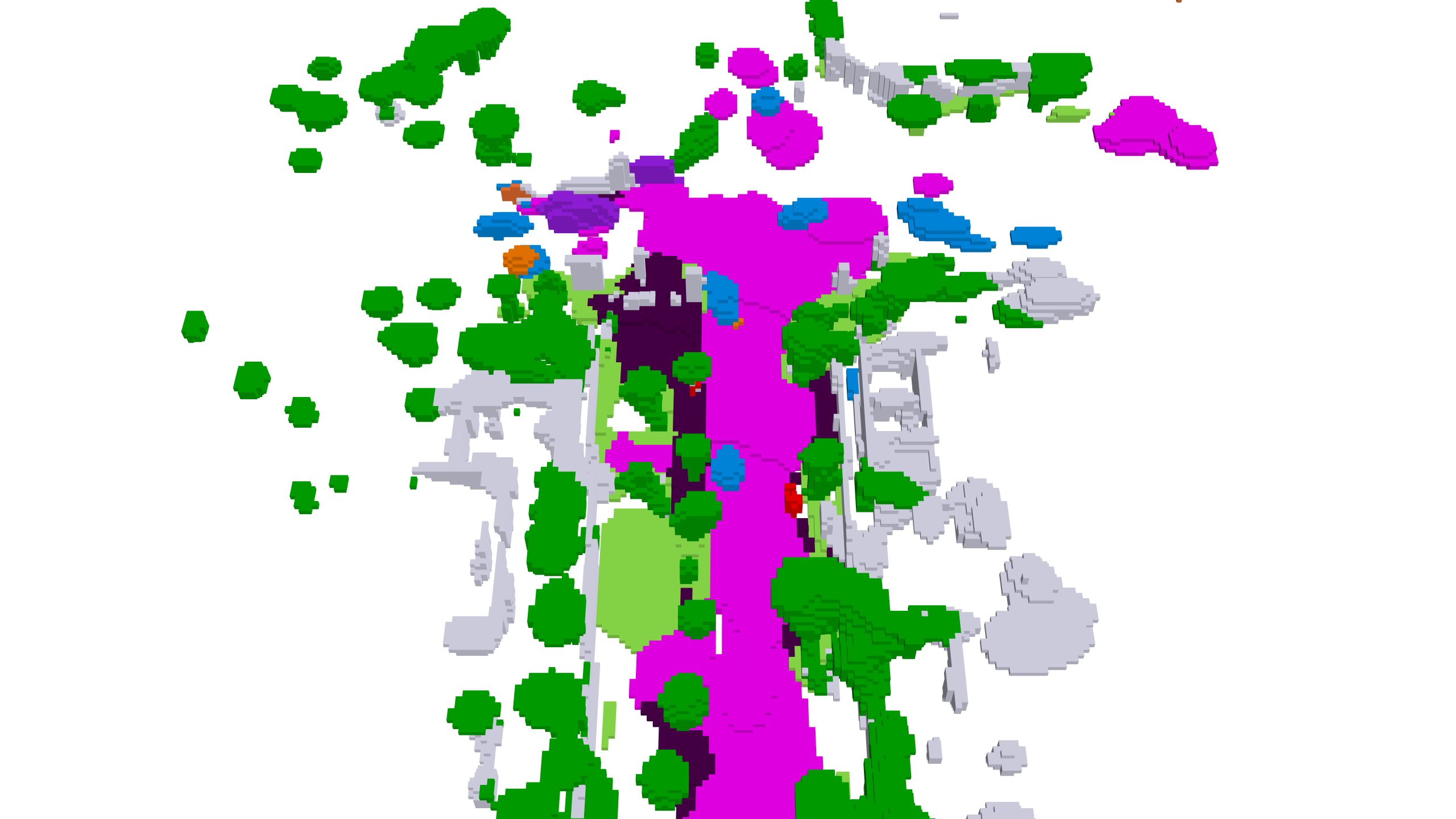}} & \fbox{\includegraphics[width=1\linewidth]{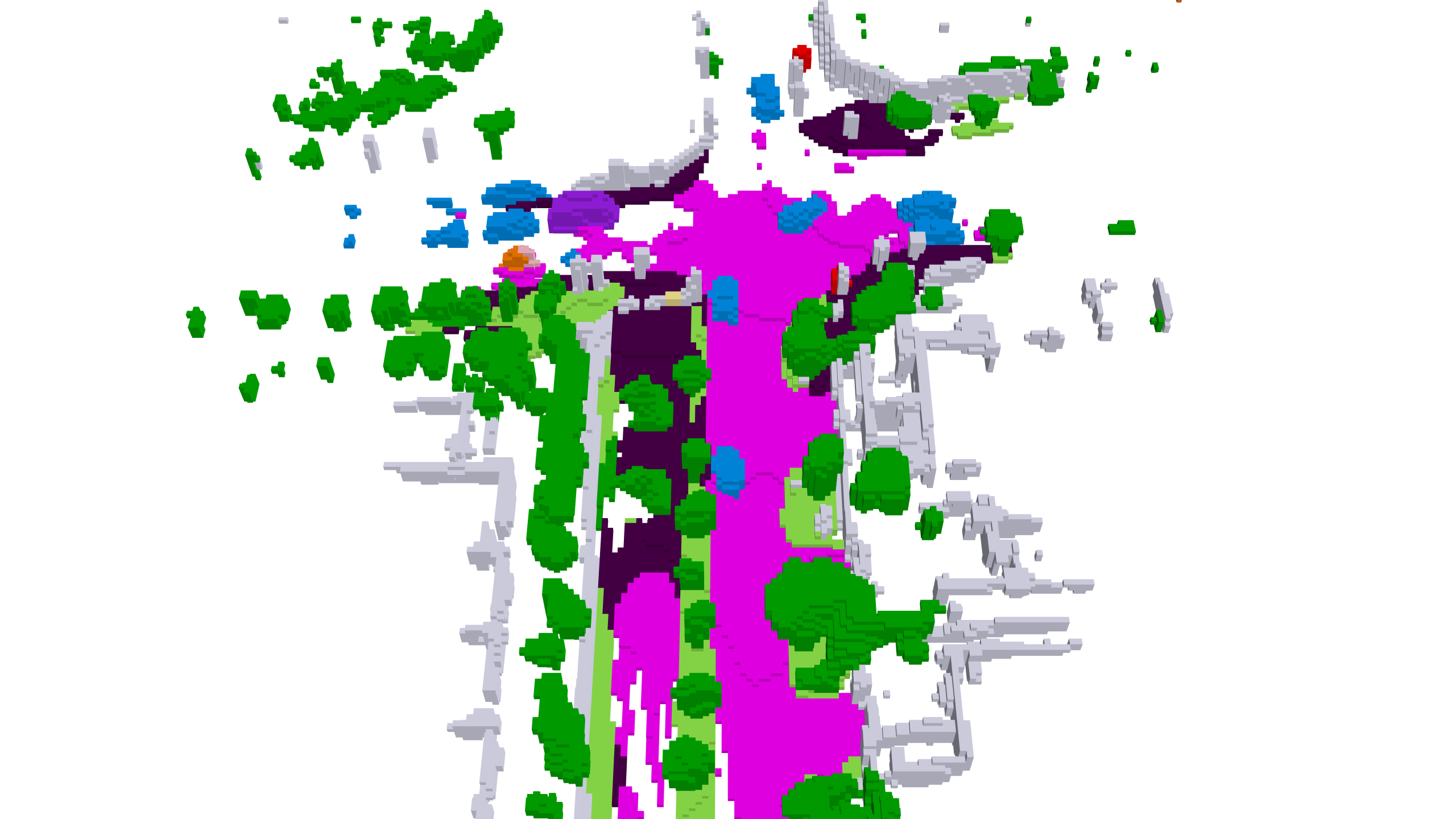}} \\
        \fbox{\includegraphics[width=1\linewidth]{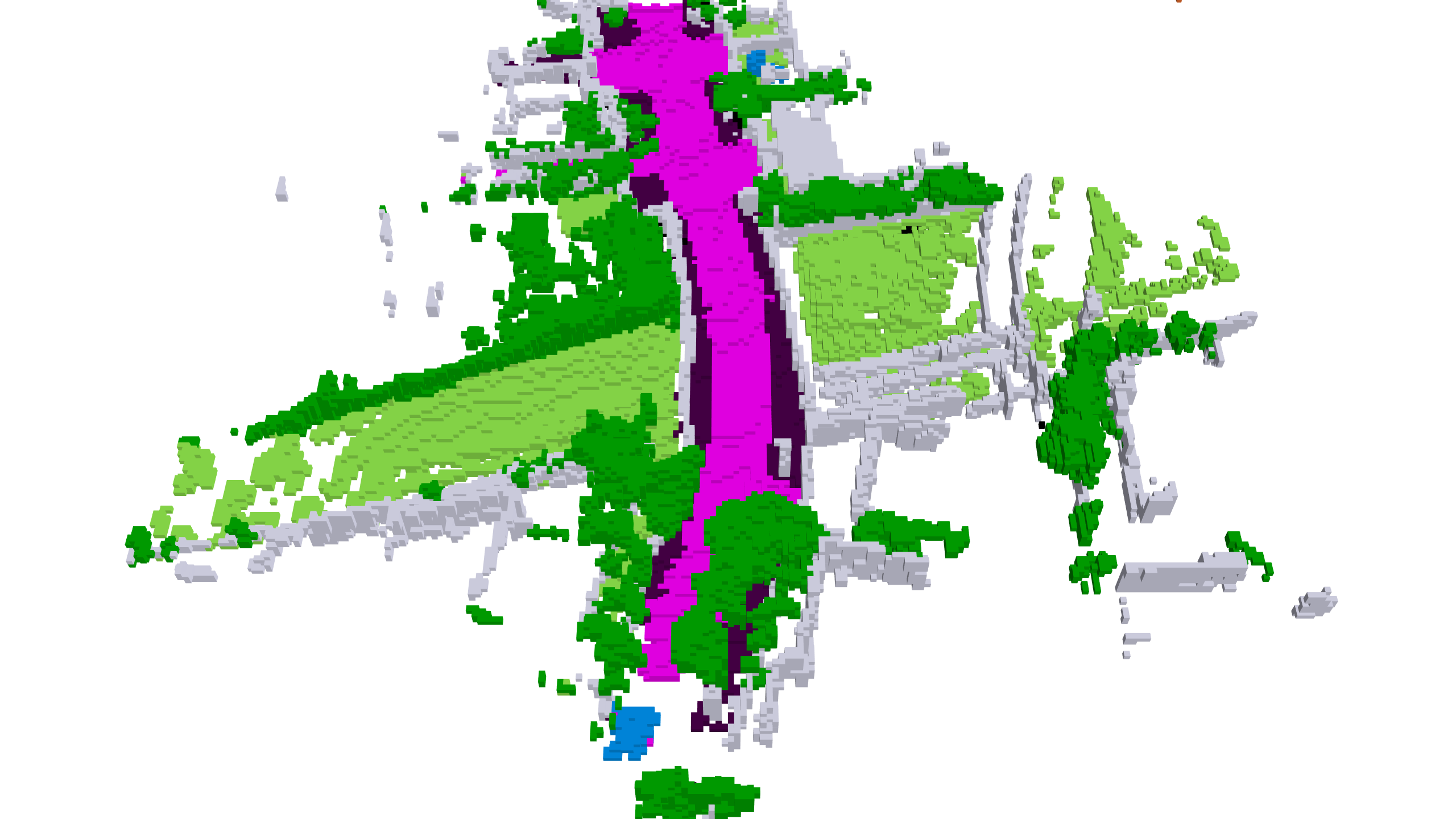}} & \fbox{\includegraphics[width=1\linewidth]{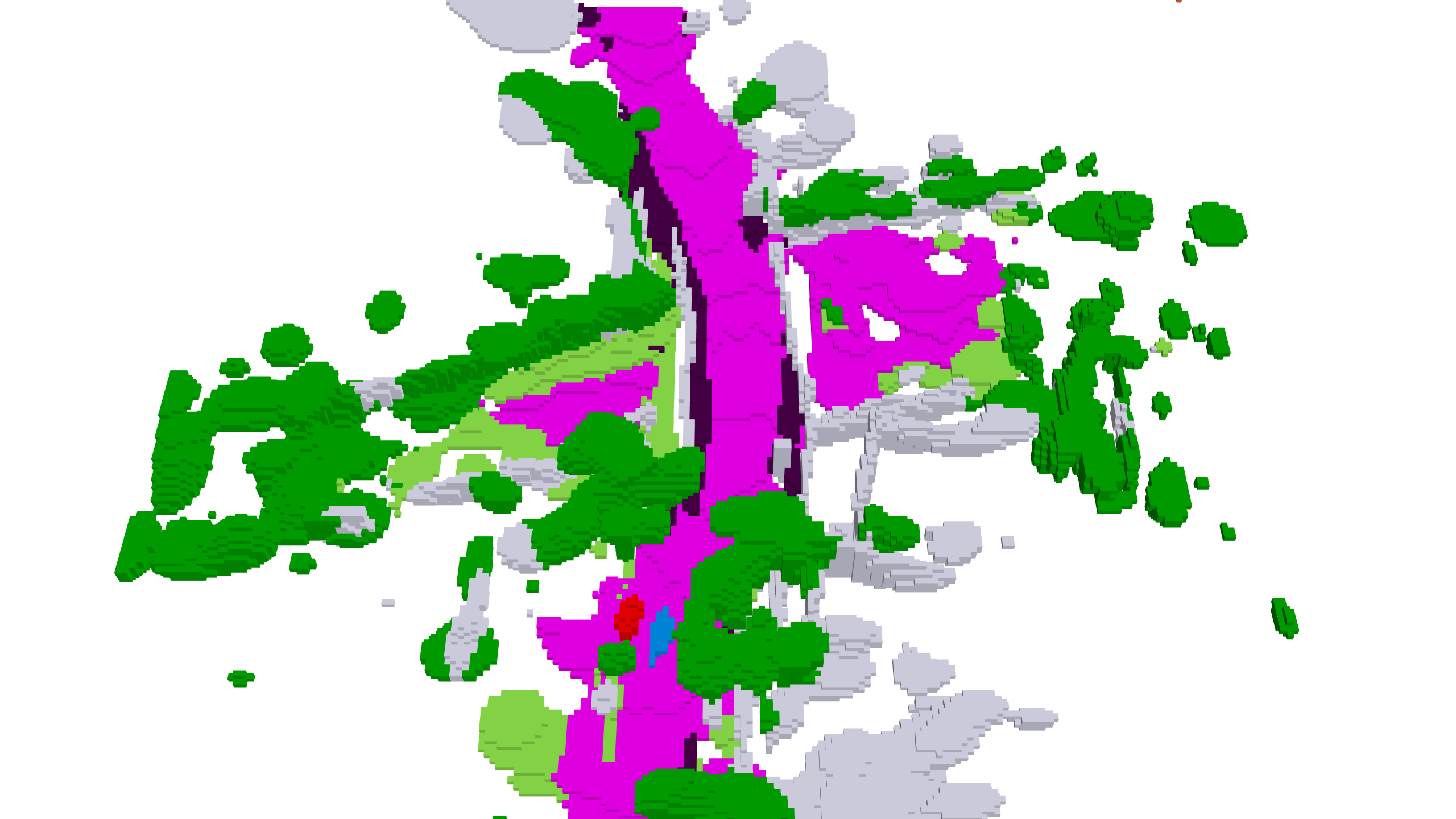}} & \fbox{\includegraphics[width=1\linewidth]{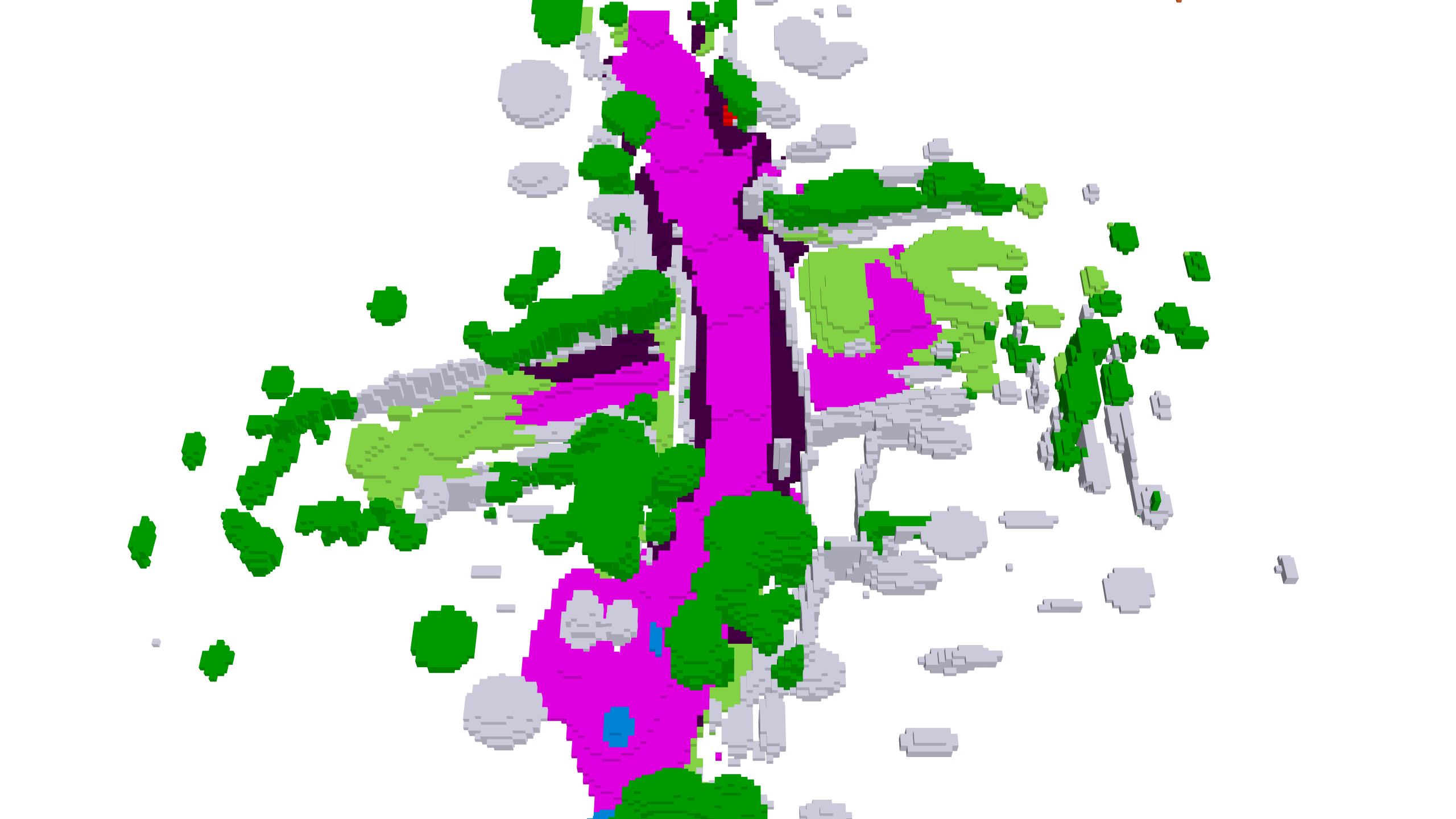}} & \fbox{\includegraphics[width=1\linewidth]{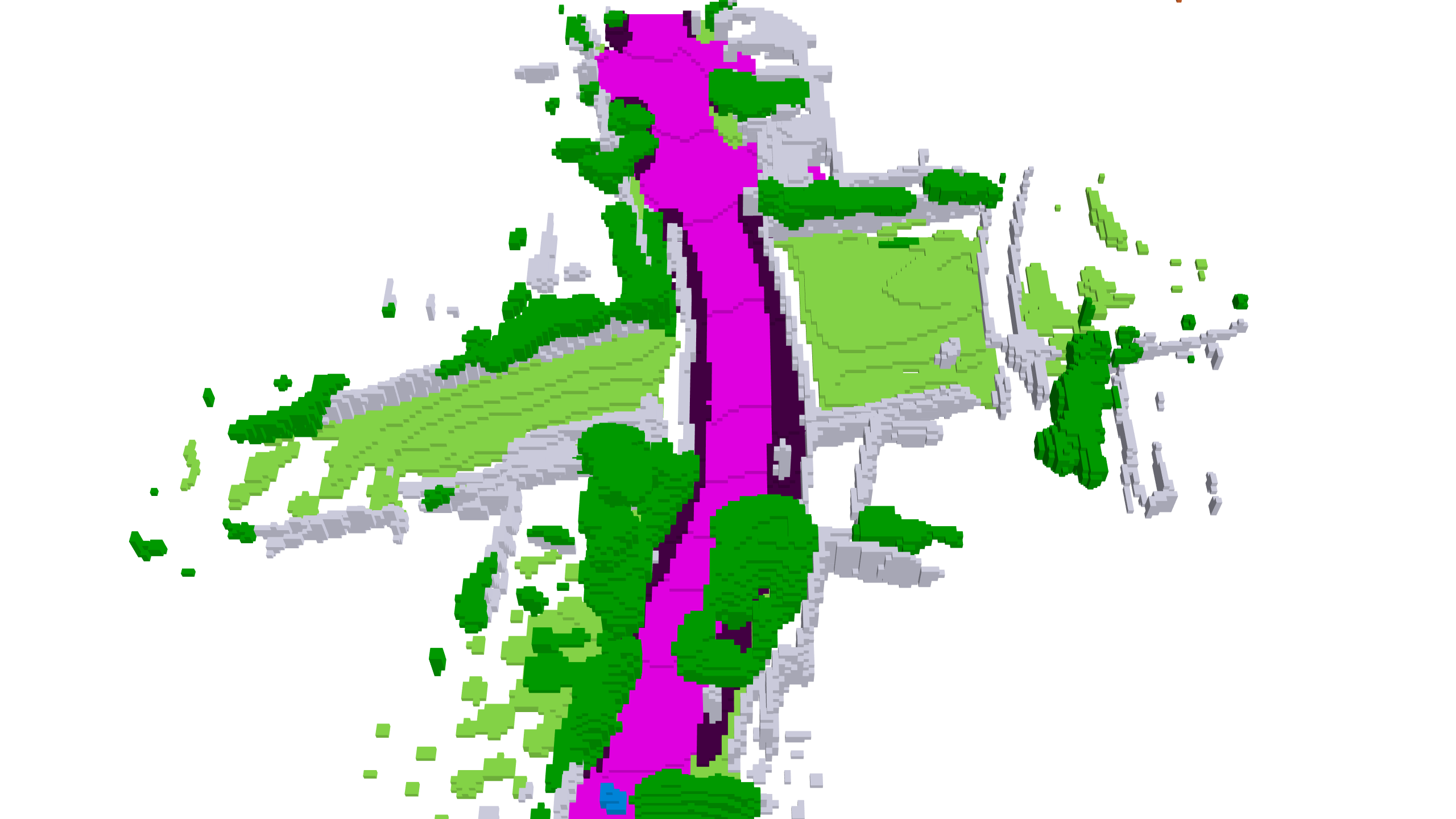}} \\
        \fbox{\includegraphics[width=1\linewidth]{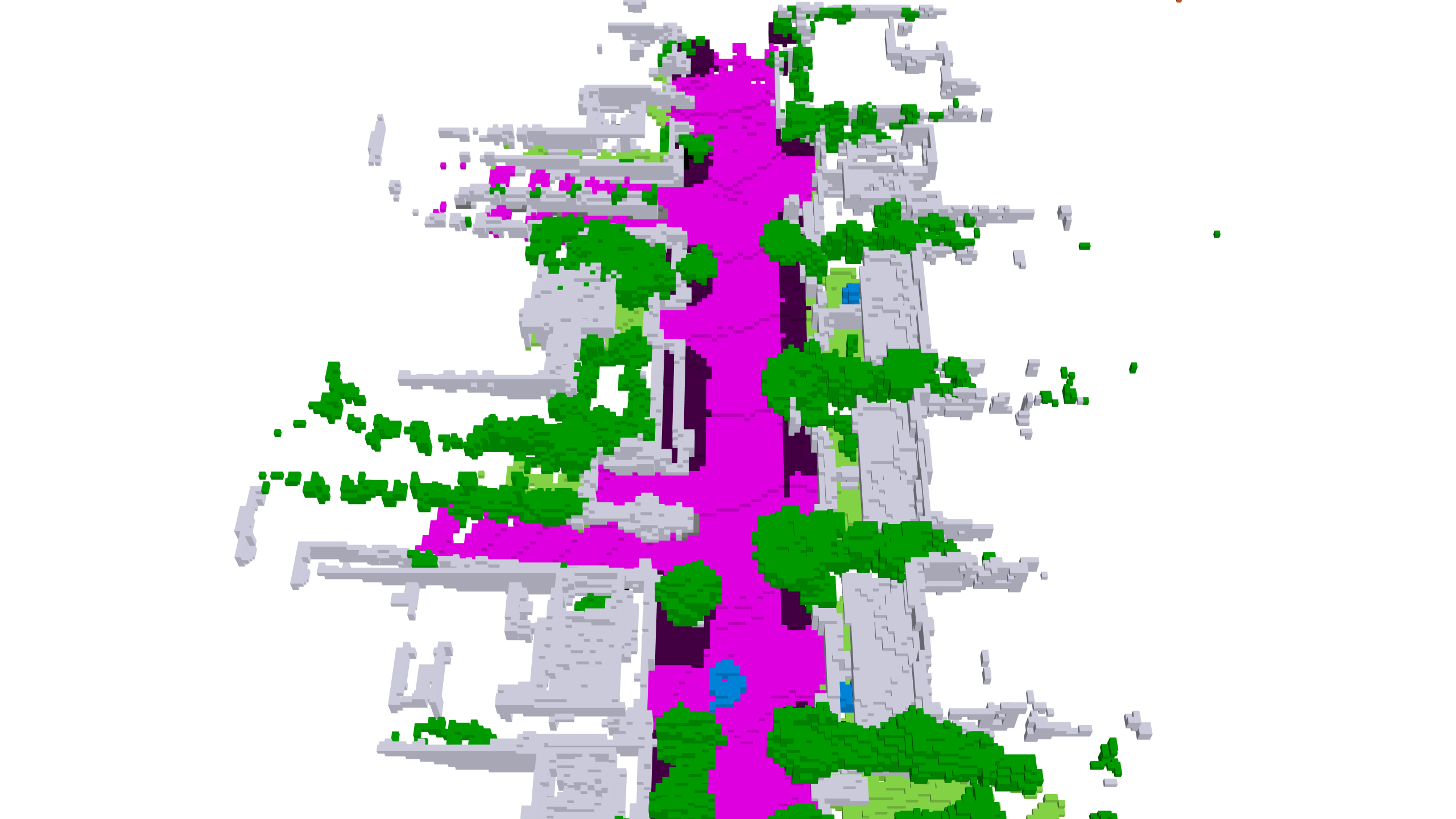}} & \fbox{\includegraphics[width=1\linewidth]{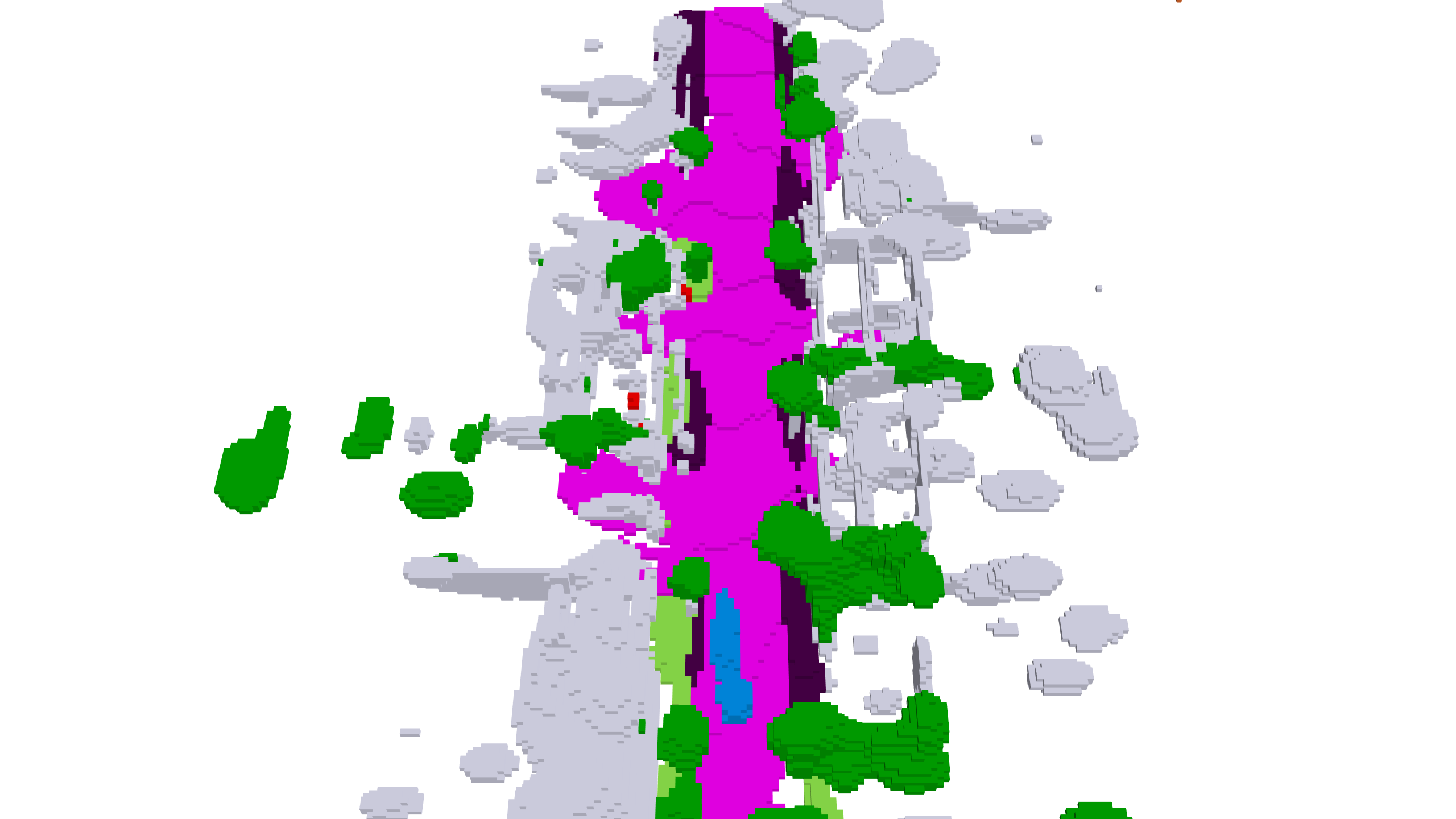}} & \fbox{\includegraphics[width=1\linewidth]{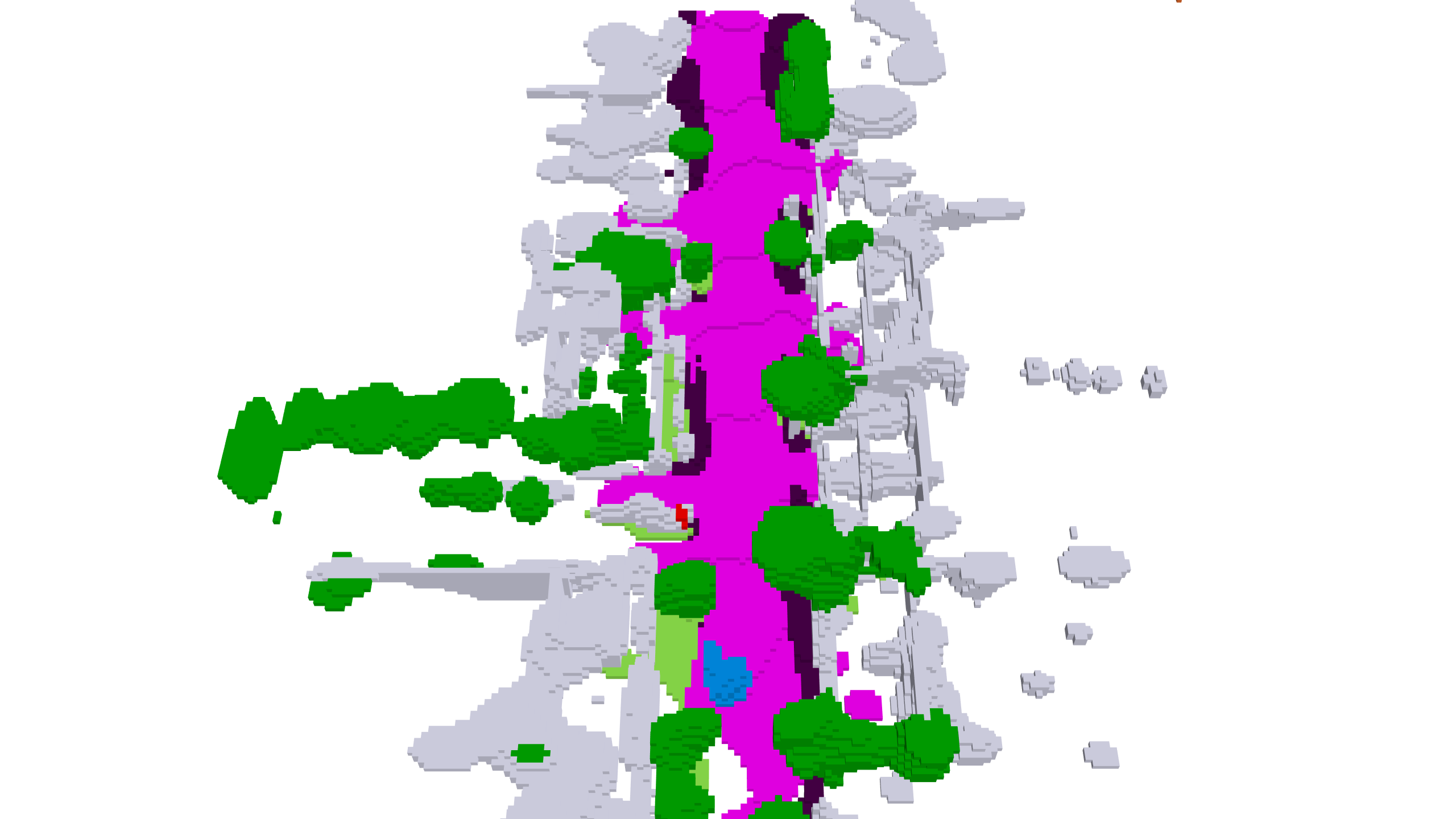}} & \fbox{\includegraphics[width=1\linewidth]{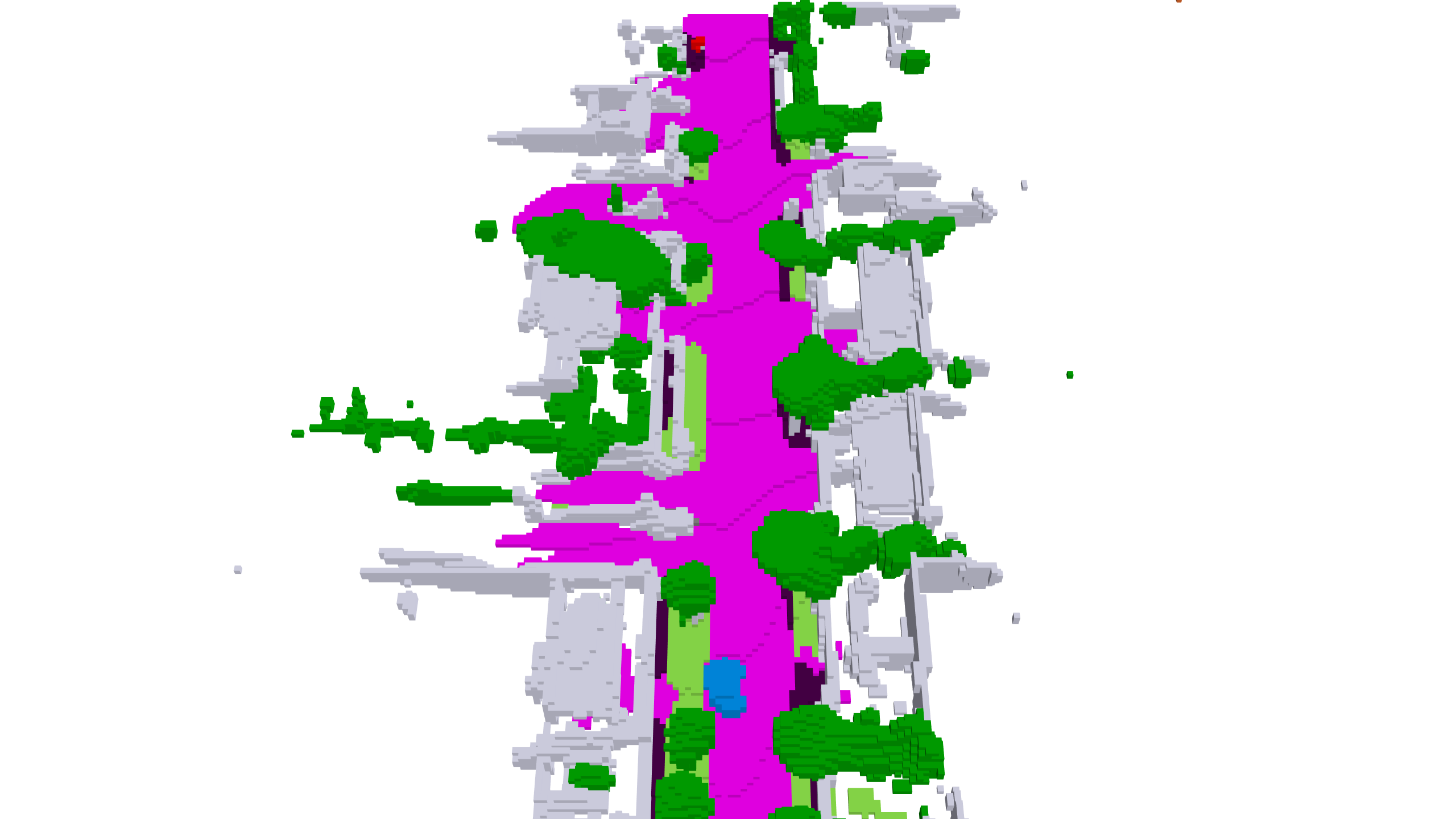}} \\
        
    \end{tabular}
    \caption{\textbf{Occupancy visualizations on nuScenes night scenes.} Visualization shows the importance of additional sensors in low-light conditions.}
    \label{fig:occupancy_visualization_night}
\end{figure*}

\begin{figure*}[htb!]
    \centering
    \footnotesize
    \setlength{\tabcolsep}{0.5pt}
    \newcommand{\imgw}{0.17\linewidth}
    \newcommand{\imgh}{3.2cm}
    \begin{tabular}{cccccc}
        \footnotesize FRONT LEFT & \footnotesize FRONT & \footnotesize FRONT RIGHT & \footnotesize BACK RIGHT & \footnotesize BACK & \footnotesize BACK LEFT \\
        \includegraphics[width=\imgw]{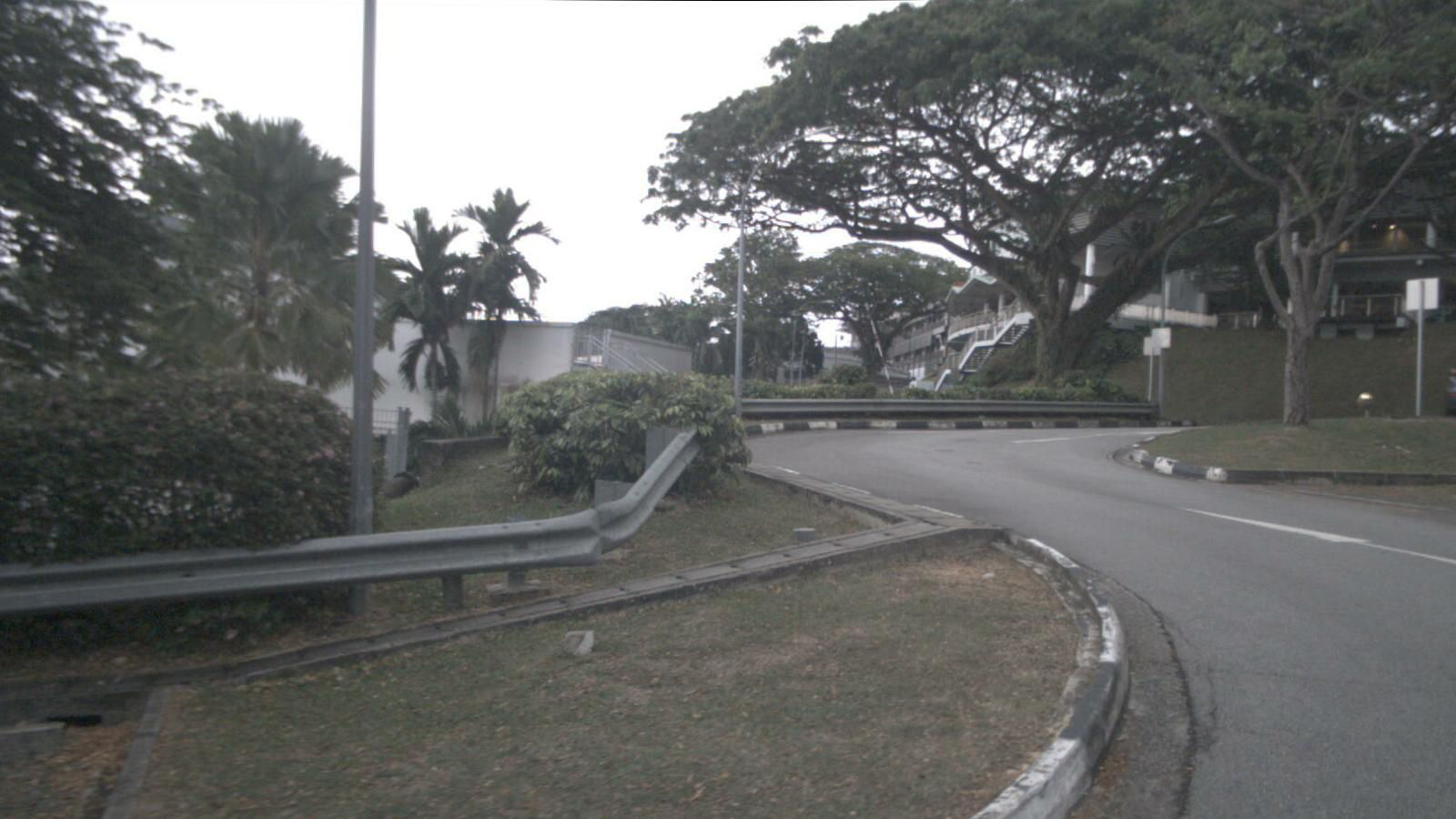} &
        \includegraphics[width=\imgw]{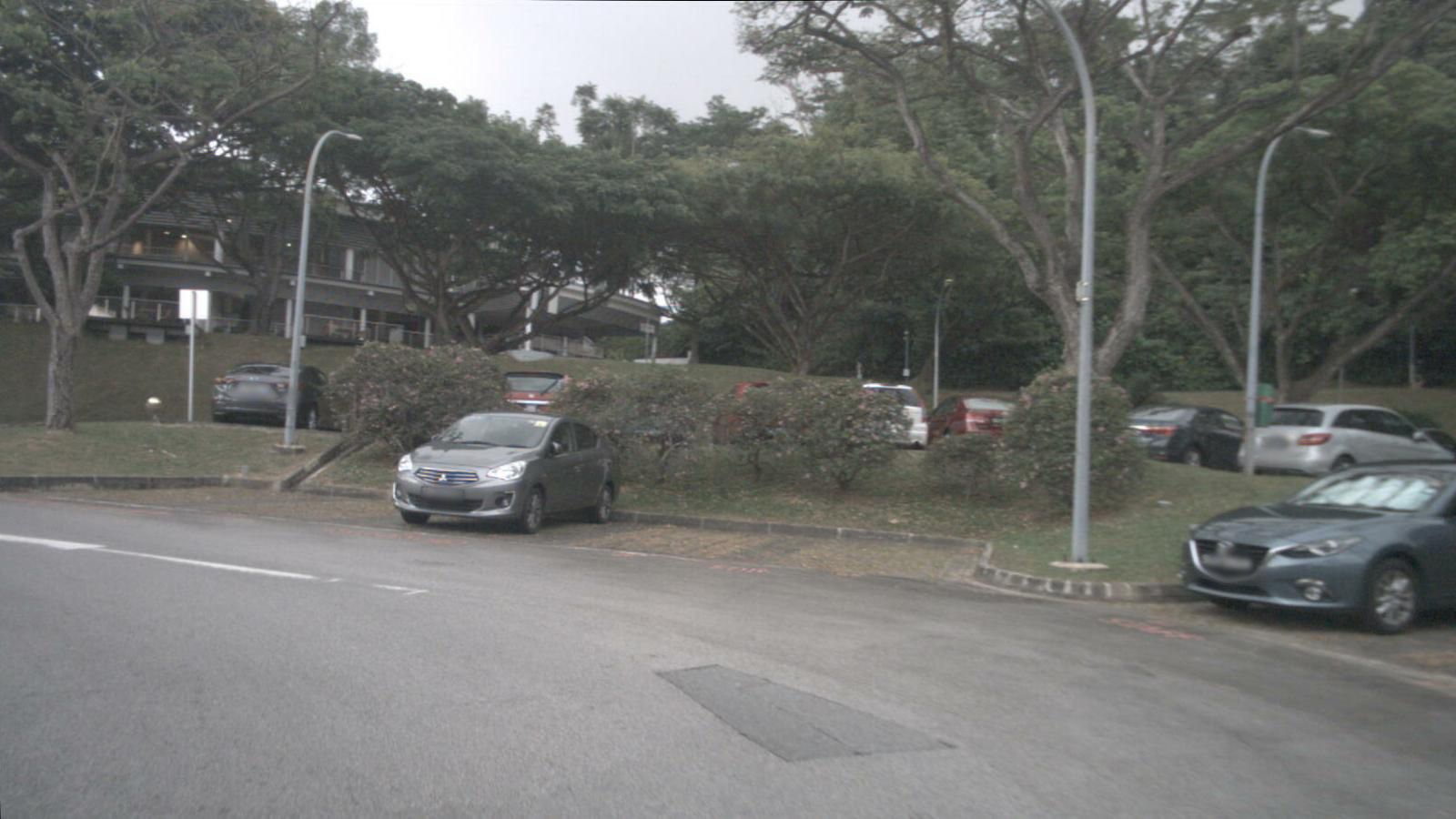} &
        \includegraphics[width=\imgw]{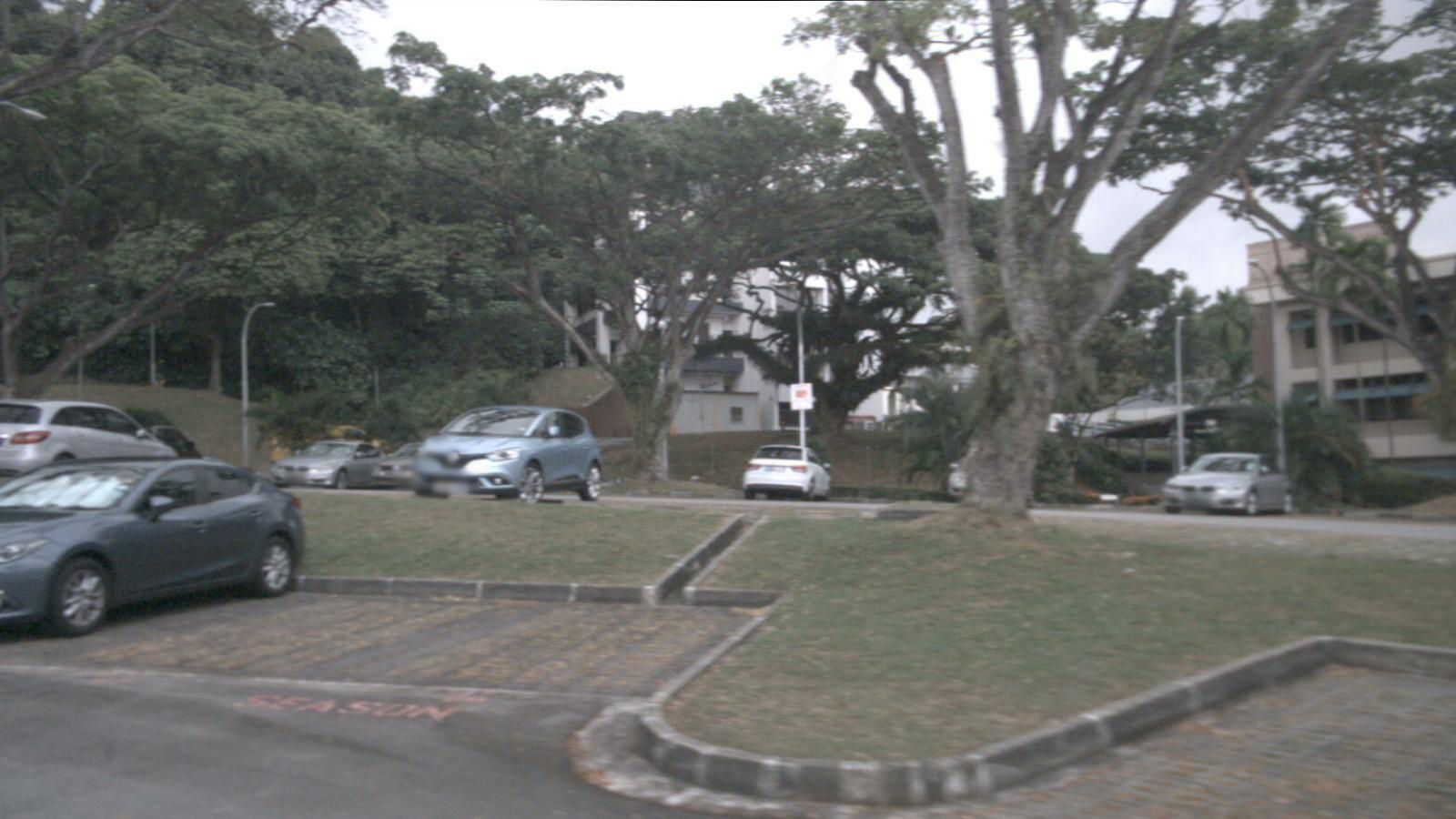} &
        \includegraphics[width=\imgw]{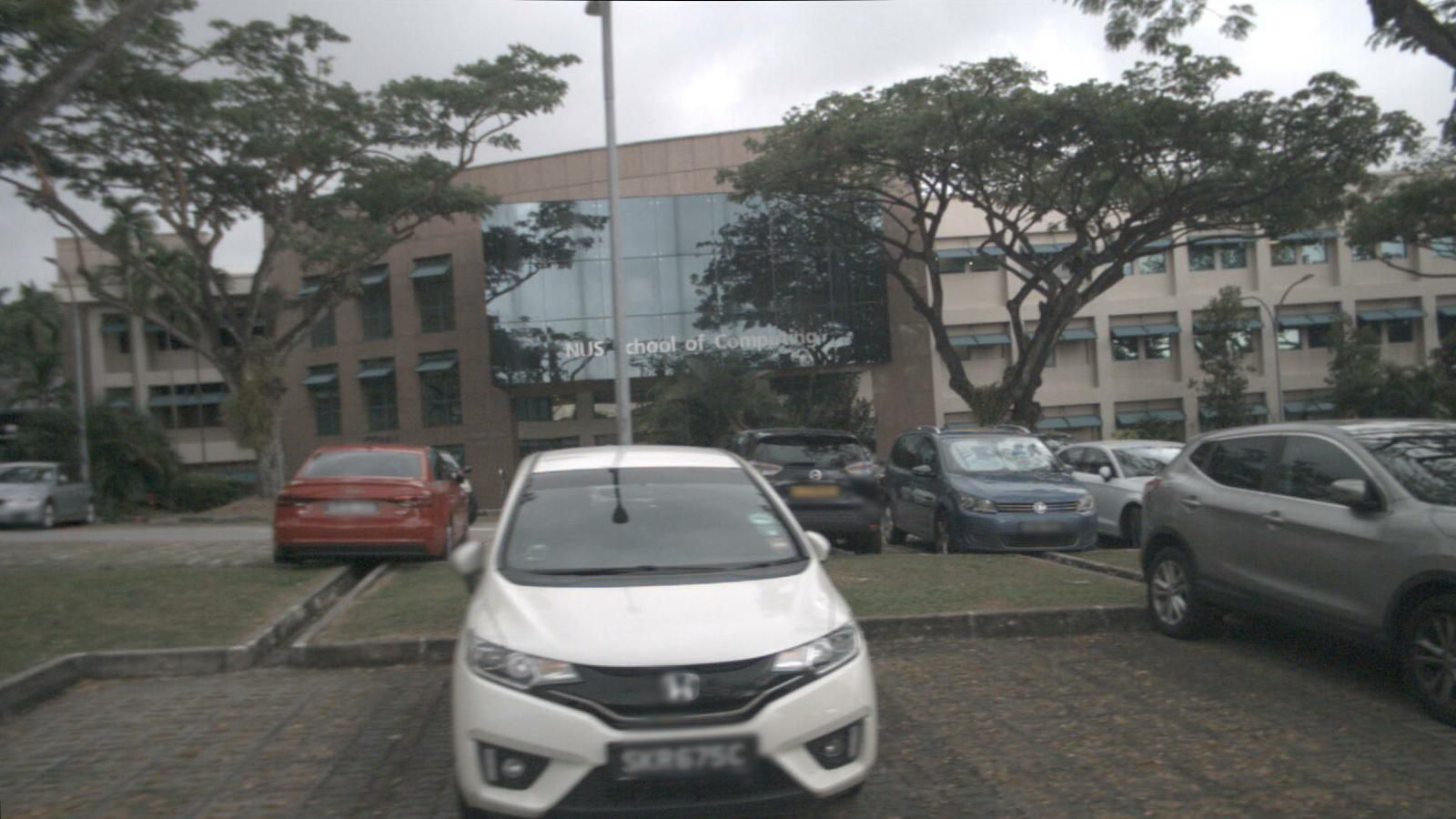} &
        \includegraphics[width=\imgw]{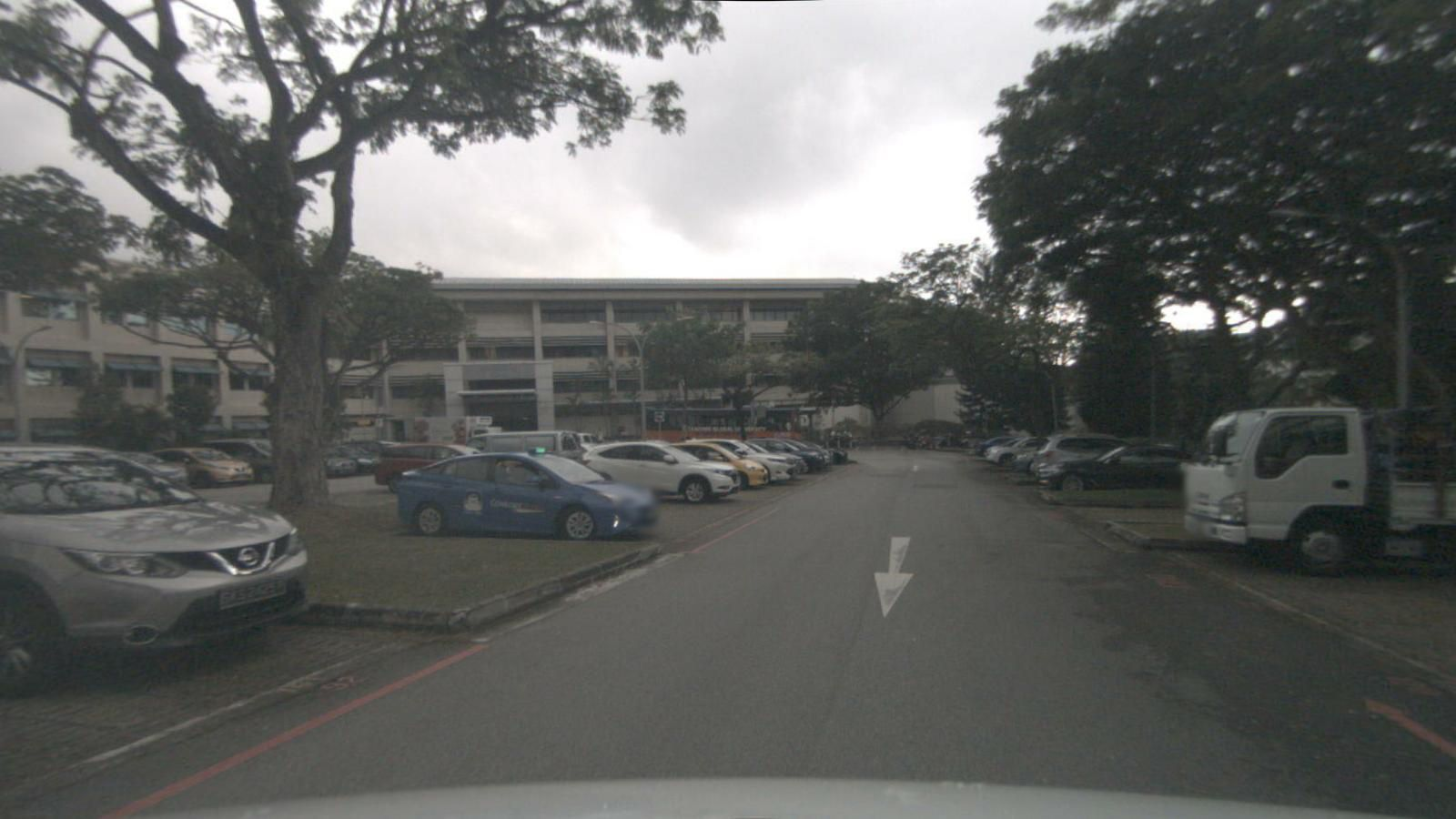} &
        \includegraphics[width=\imgw]{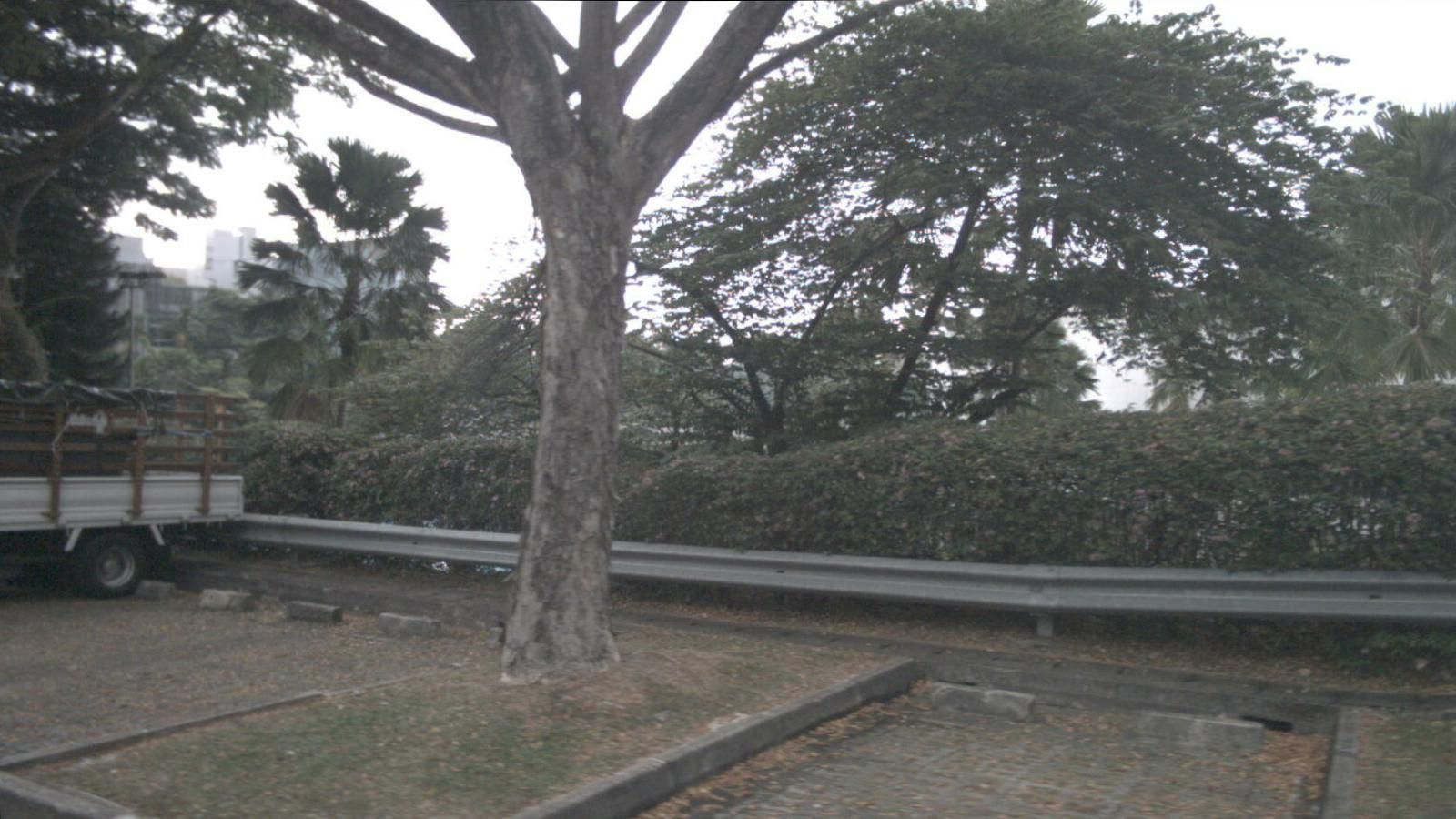} \\
        \multicolumn{2}{c}{\includegraphics[height=\imgh,trim=100 90 90 90,clip]{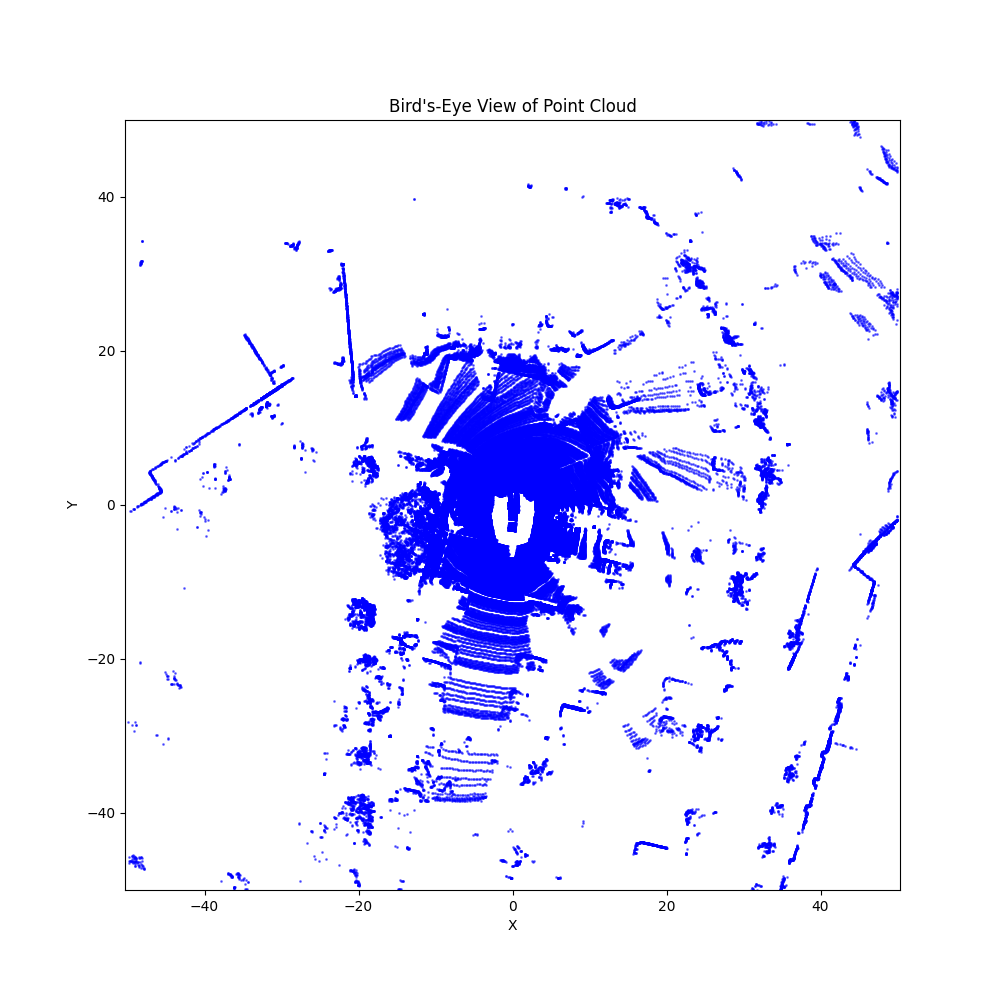}} &
        \multicolumn{2}{c}{\includegraphics[height=\imgh,trim=175 220 175 220,clip]{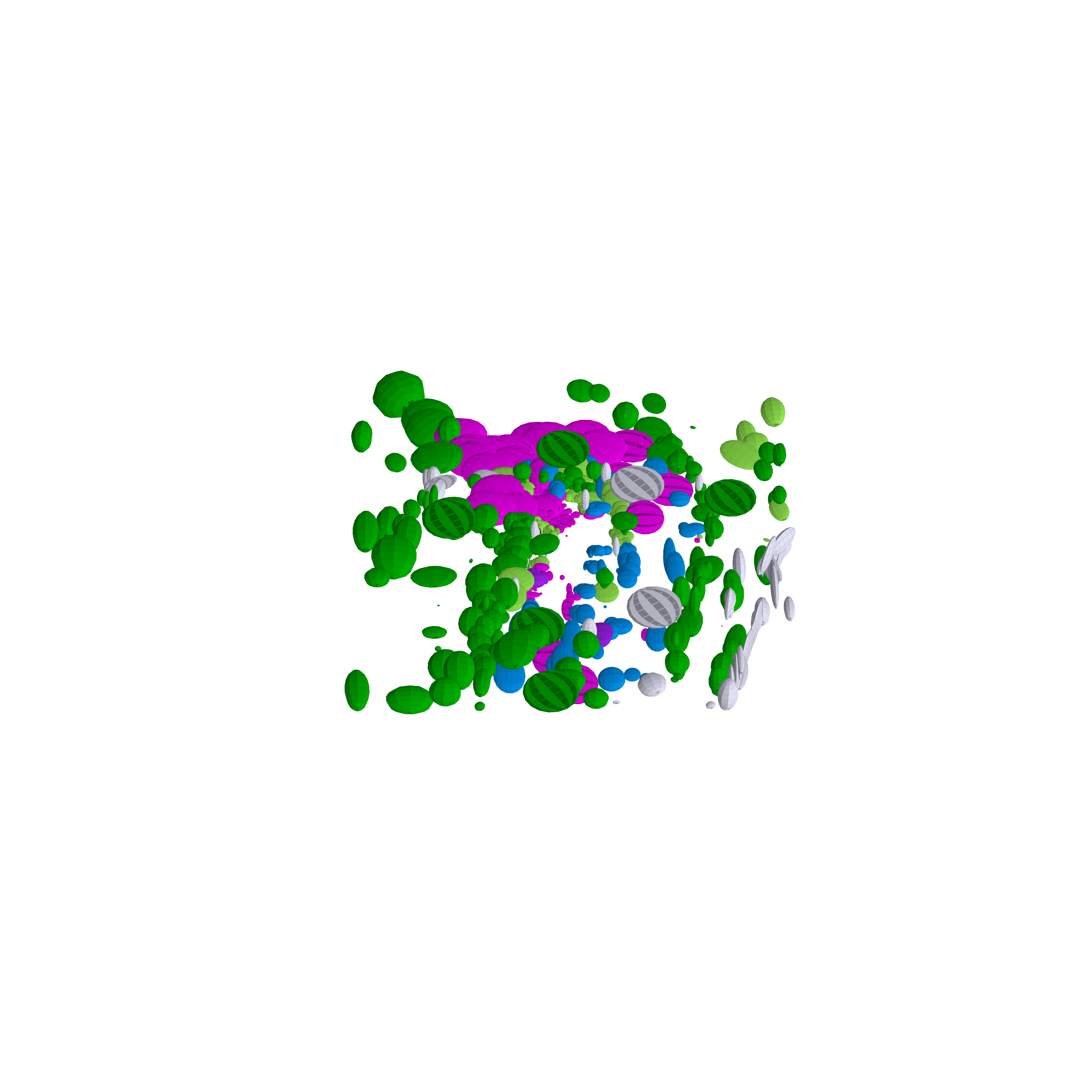}} &
        \multicolumn{2}{c}{\includegraphics[height=\imgh,trim=175 220 175 220,clip]{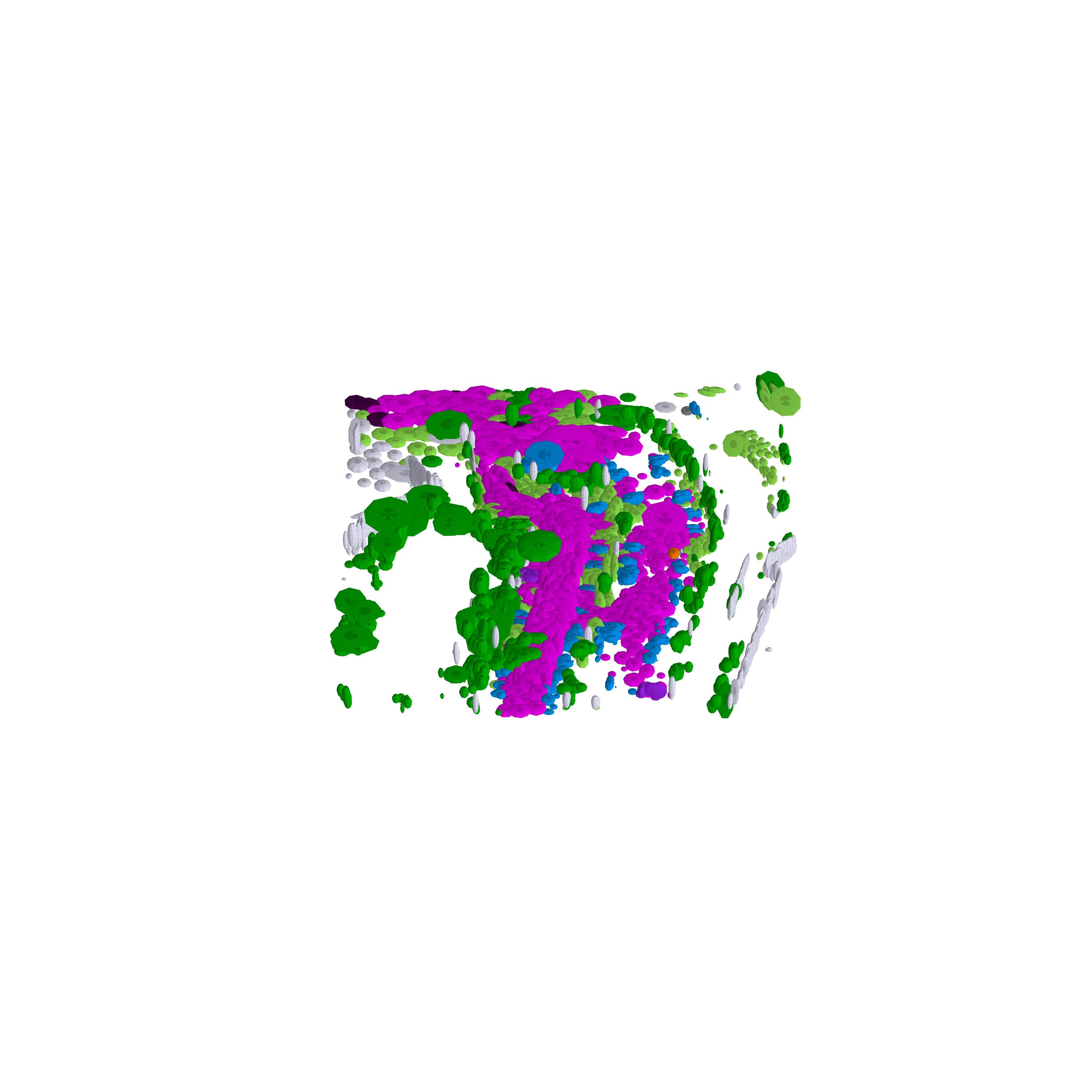}} \\
        \multicolumn{2}{c}{\footnotesize BEV LiDAR} &
        \multicolumn{2}{c}{\footnotesize Pred. Gaussians (C)} &
        \multicolumn{2}{c}{\footnotesize Pred. Gaussians (C+L+R)} \\
        \multicolumn{2}{c}{\includegraphics[width=0.33\linewidth]{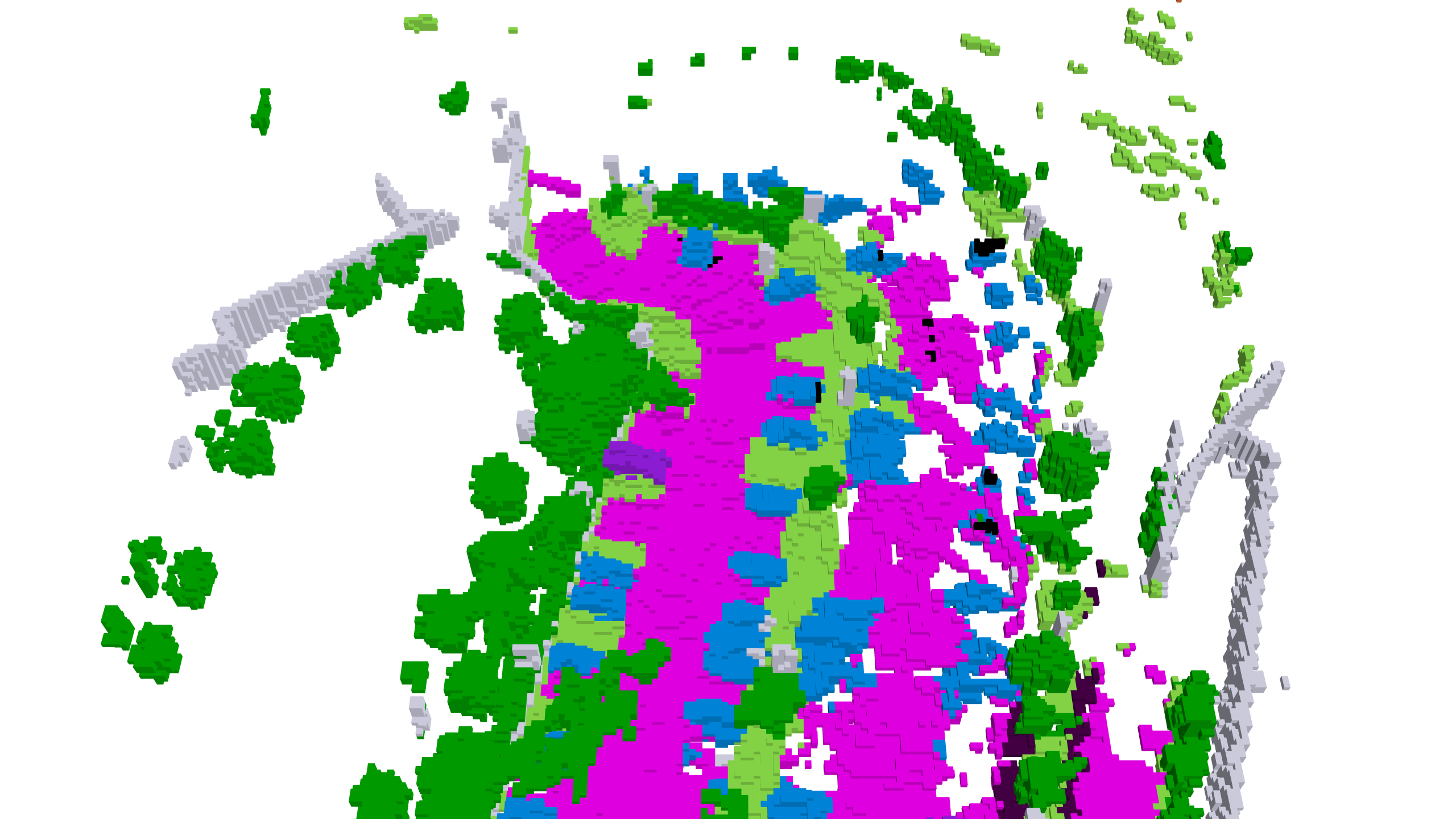}} &
        \multicolumn{2}{c}{\includegraphics[width=0.33\linewidth]{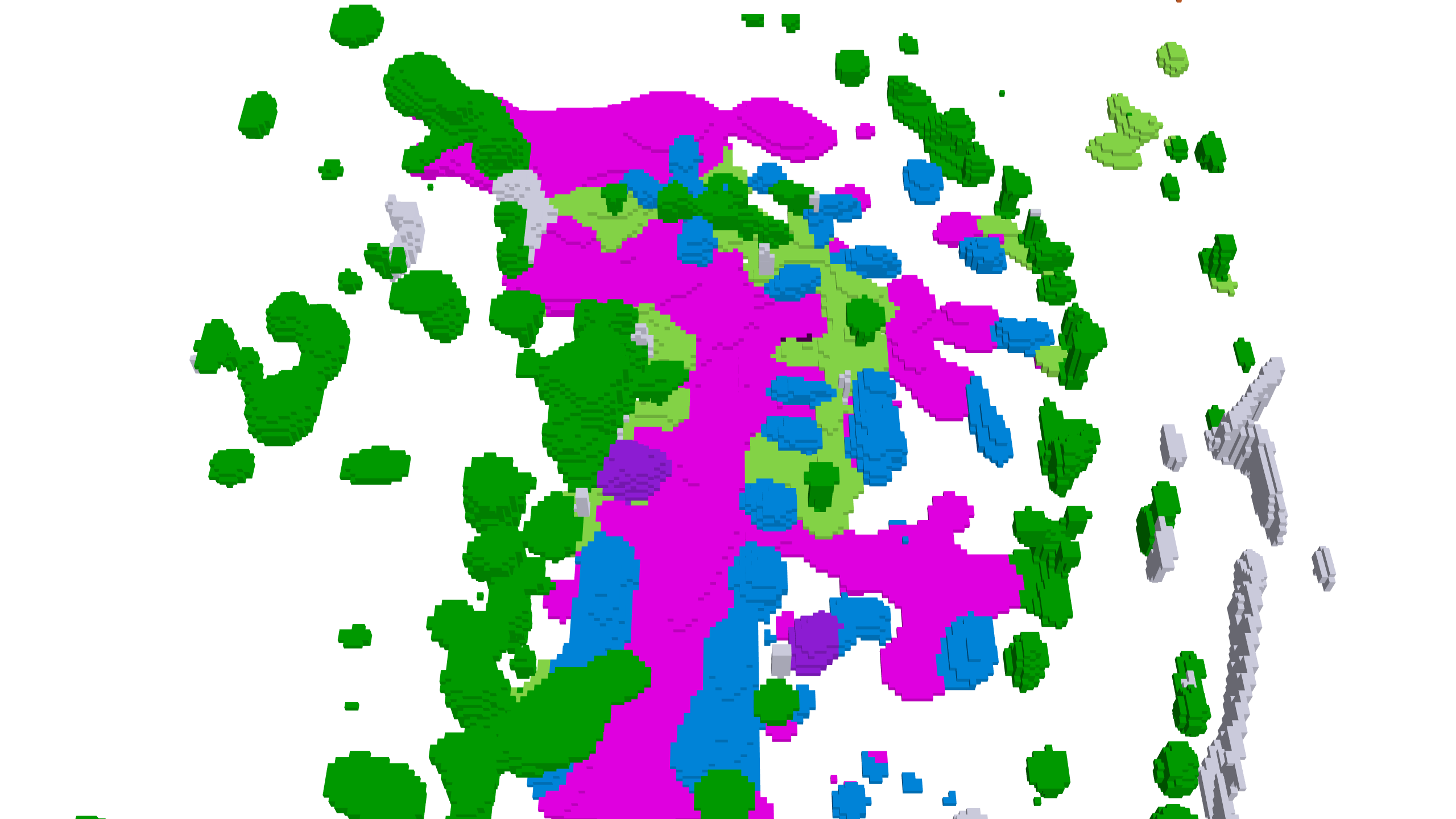}} &
        \multicolumn{2}{c}{\includegraphics[width=0.33\linewidth]{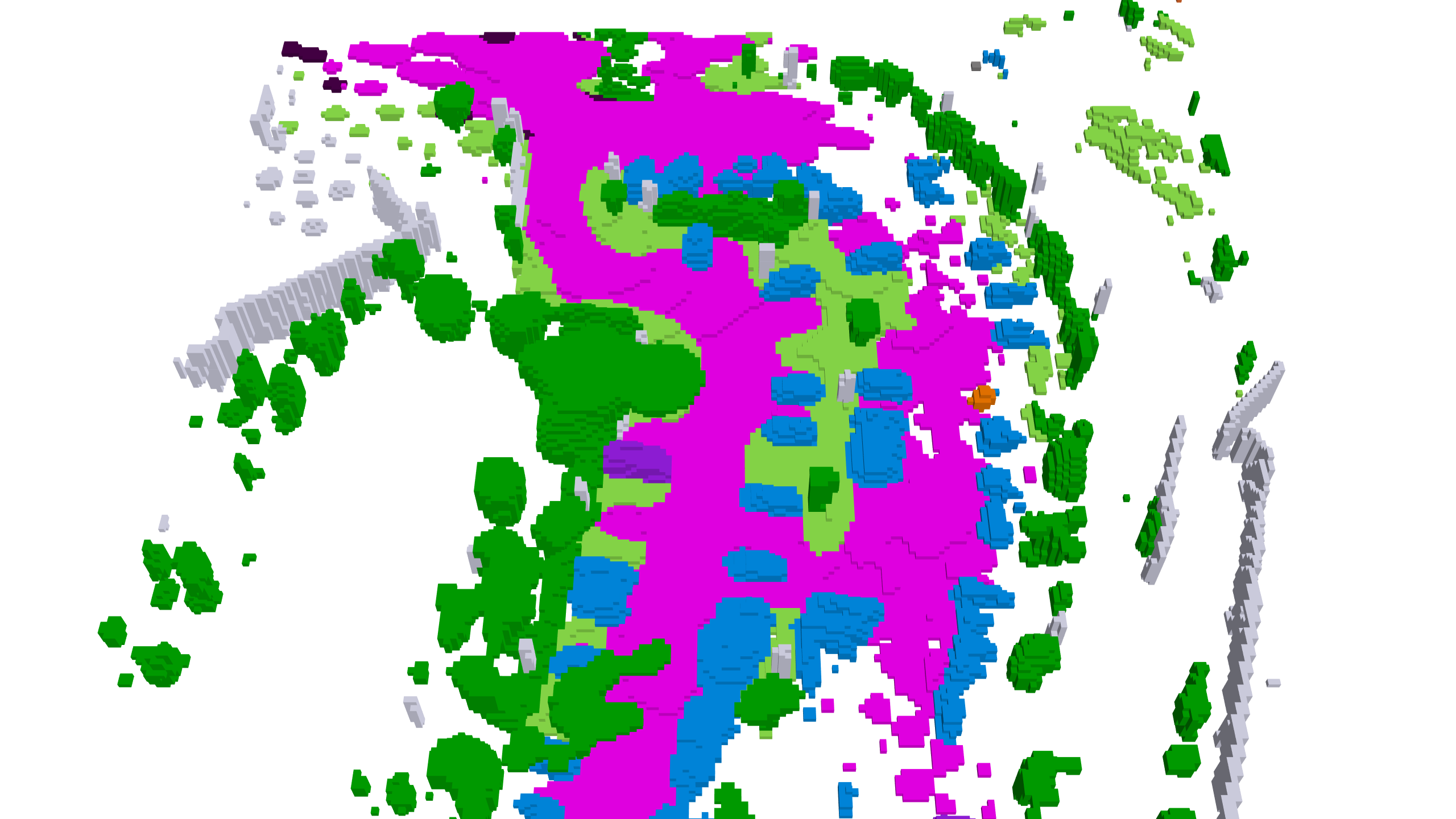}} \\ 
        \multicolumn{2}{c}{\footnotesize Ground Truth} &
        \multicolumn{2}{c}{\footnotesize Pred. Occupancy (C)} &
        \multicolumn{2}{c}{\footnotesize Pred. Occupancy (C+L+R)} \\

        \footnotesize FRONT LEFT & \footnotesize FRONT & \footnotesize FRONT RIGHT & \footnotesize BACK RIGHT & \footnotesize BACK & \footnotesize BACK LEFT \\
        \includegraphics[width=\imgw]{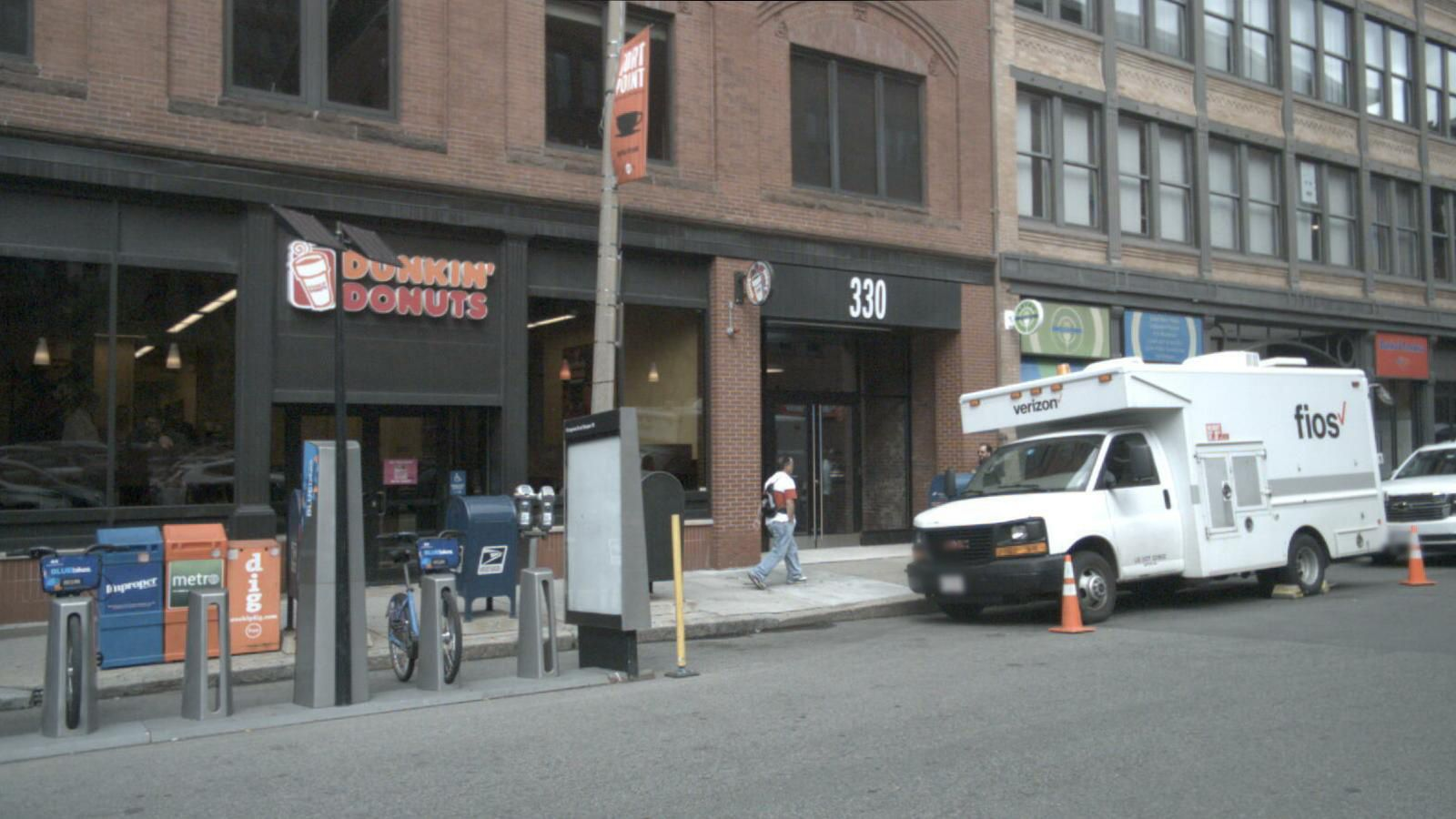} &
        \includegraphics[width=\imgw]{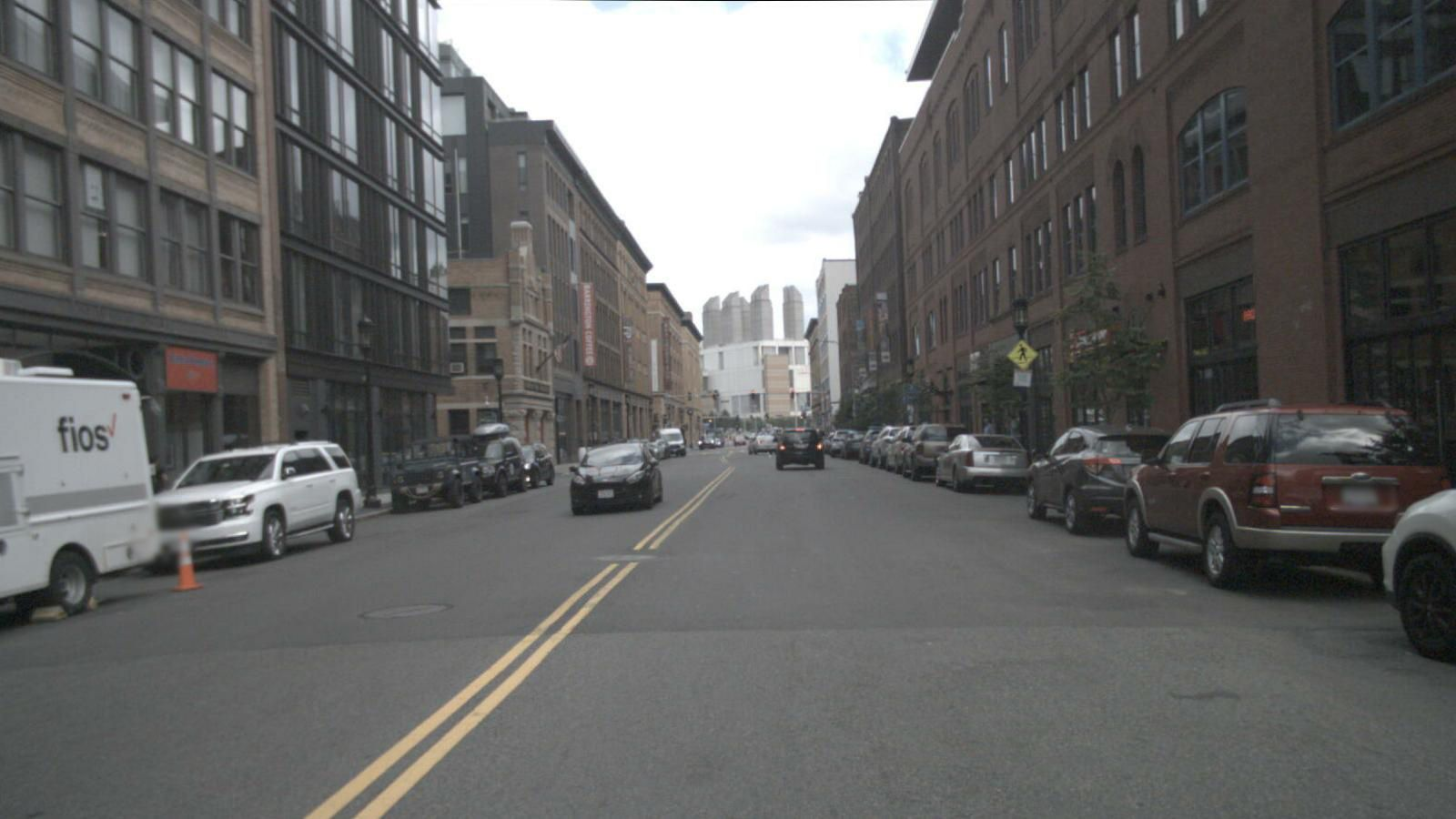} &
        \includegraphics[width=\imgw]{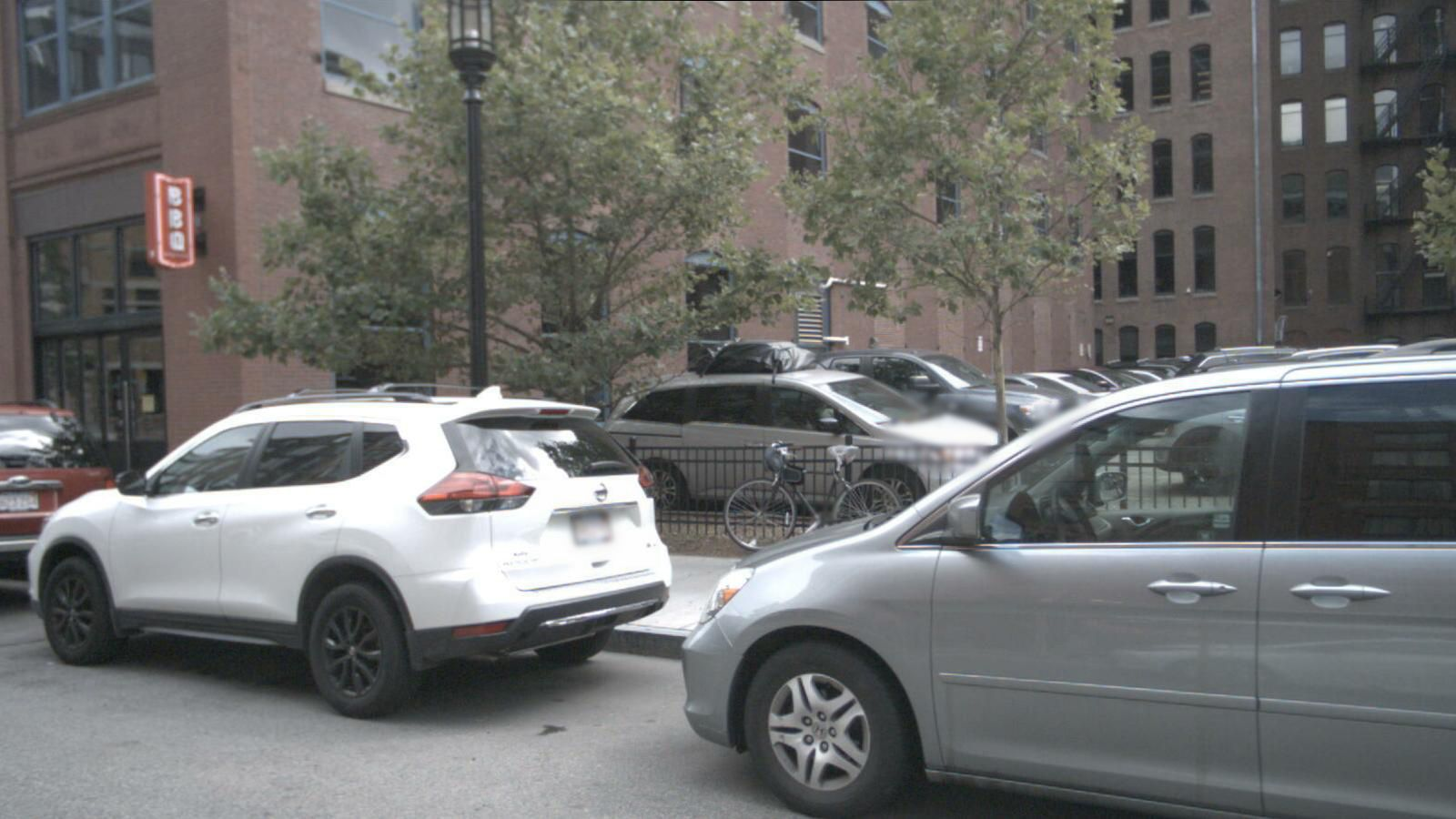} &
        \includegraphics[width=\imgw]{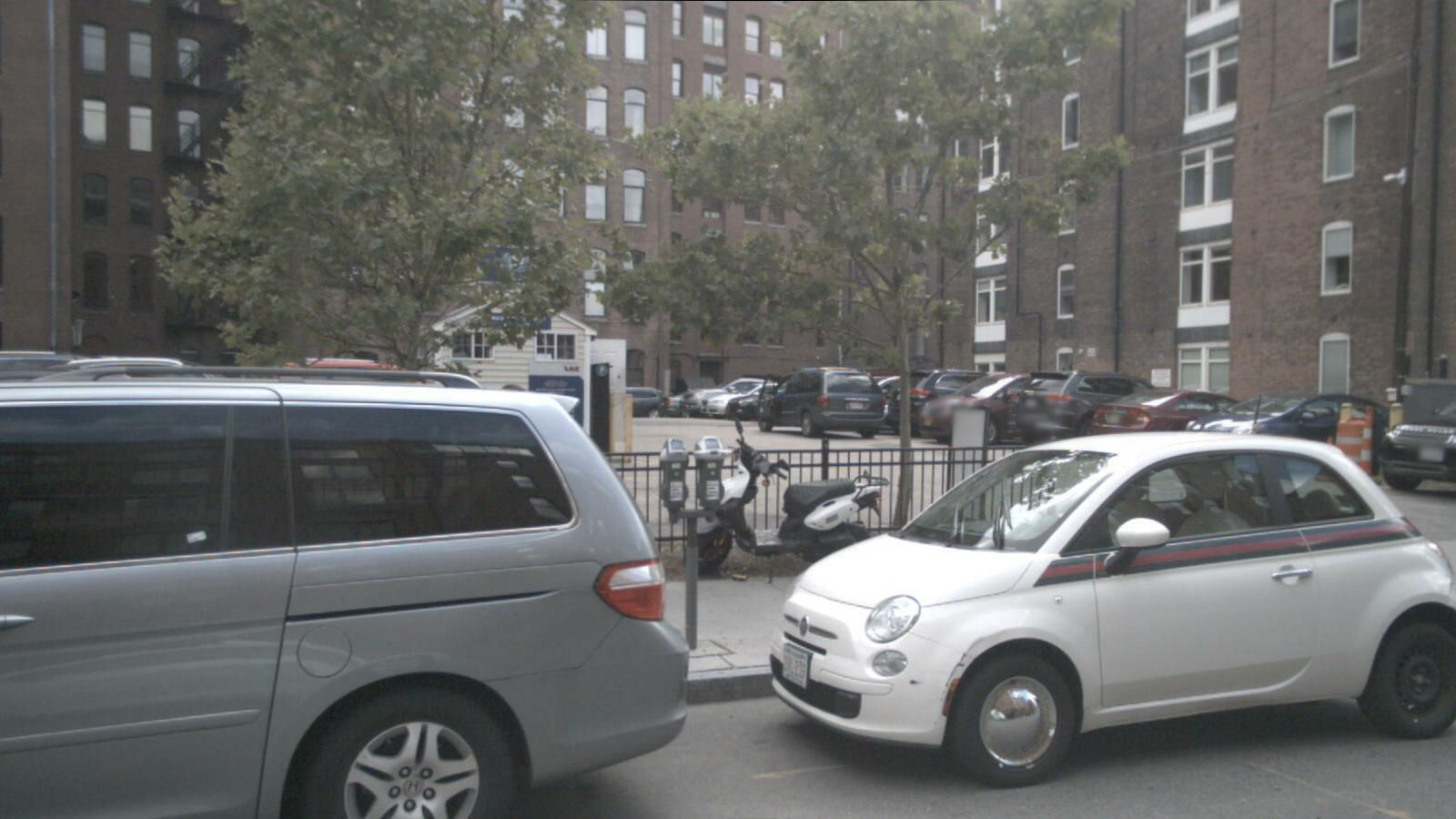} &
        \includegraphics[width=\imgw]{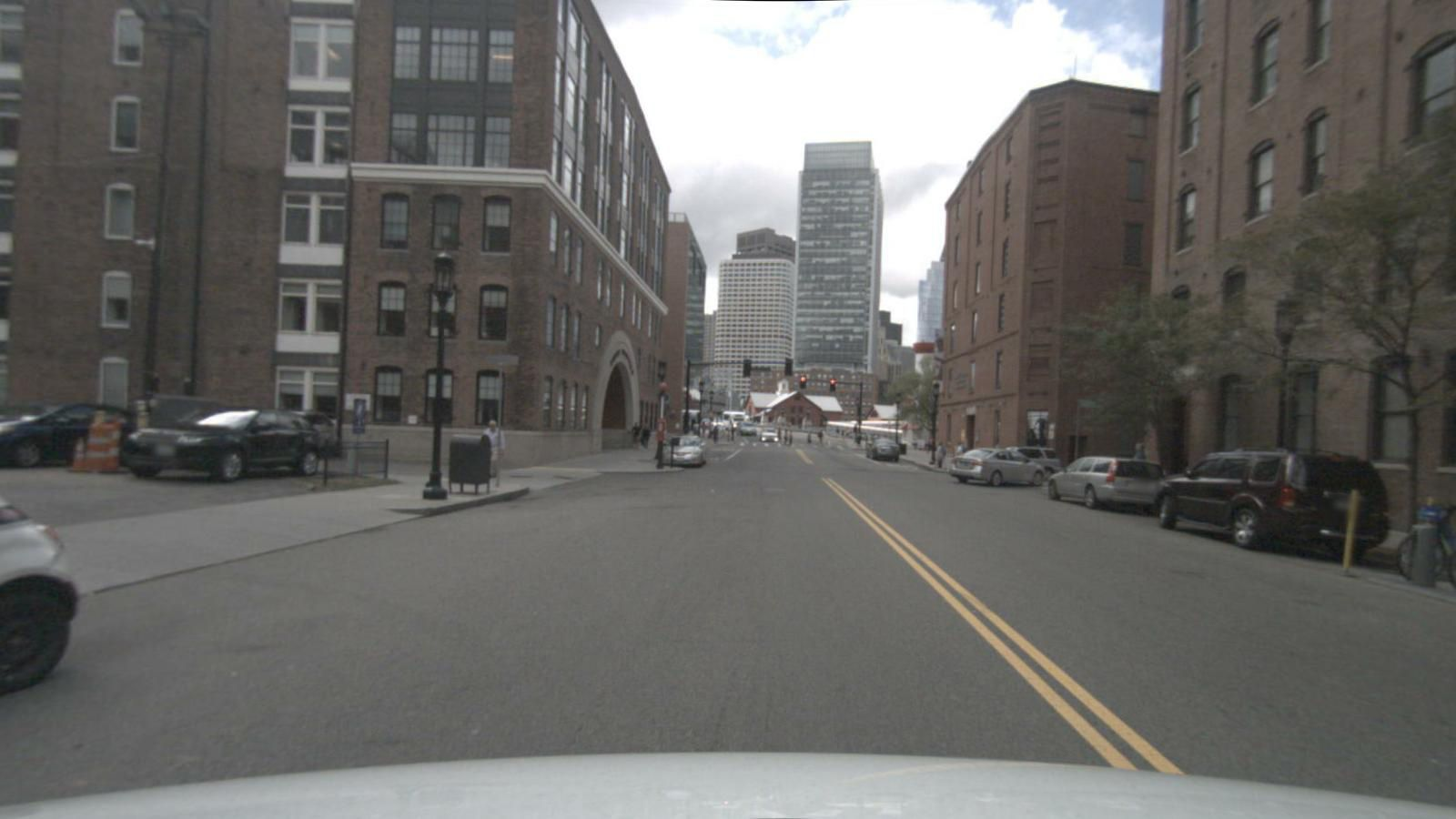} &
        \includegraphics[width=\imgw]{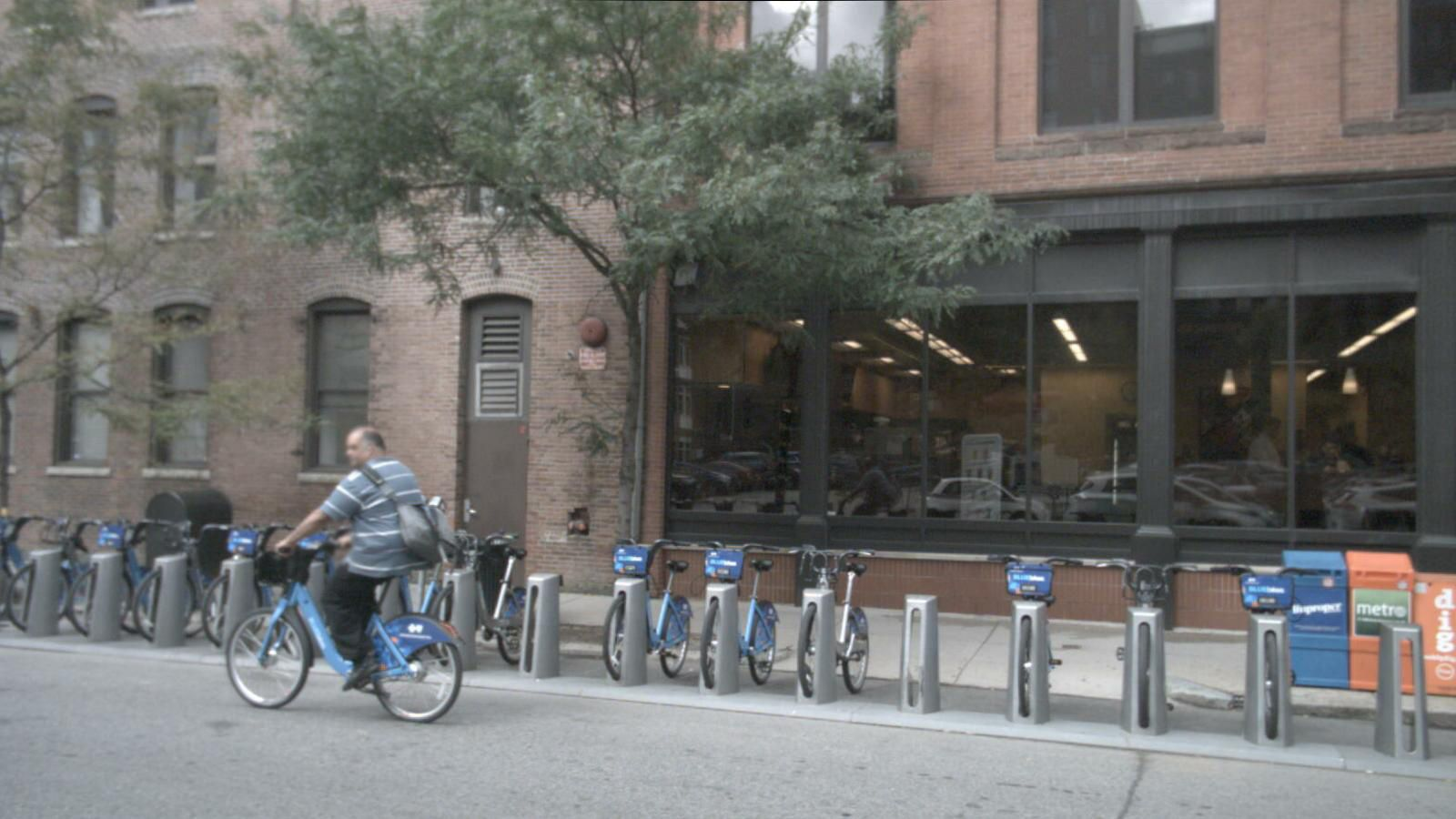} \\
        \multicolumn{2}{c}{\includegraphics[height=\imgh,trim=100 90 90 90,clip]{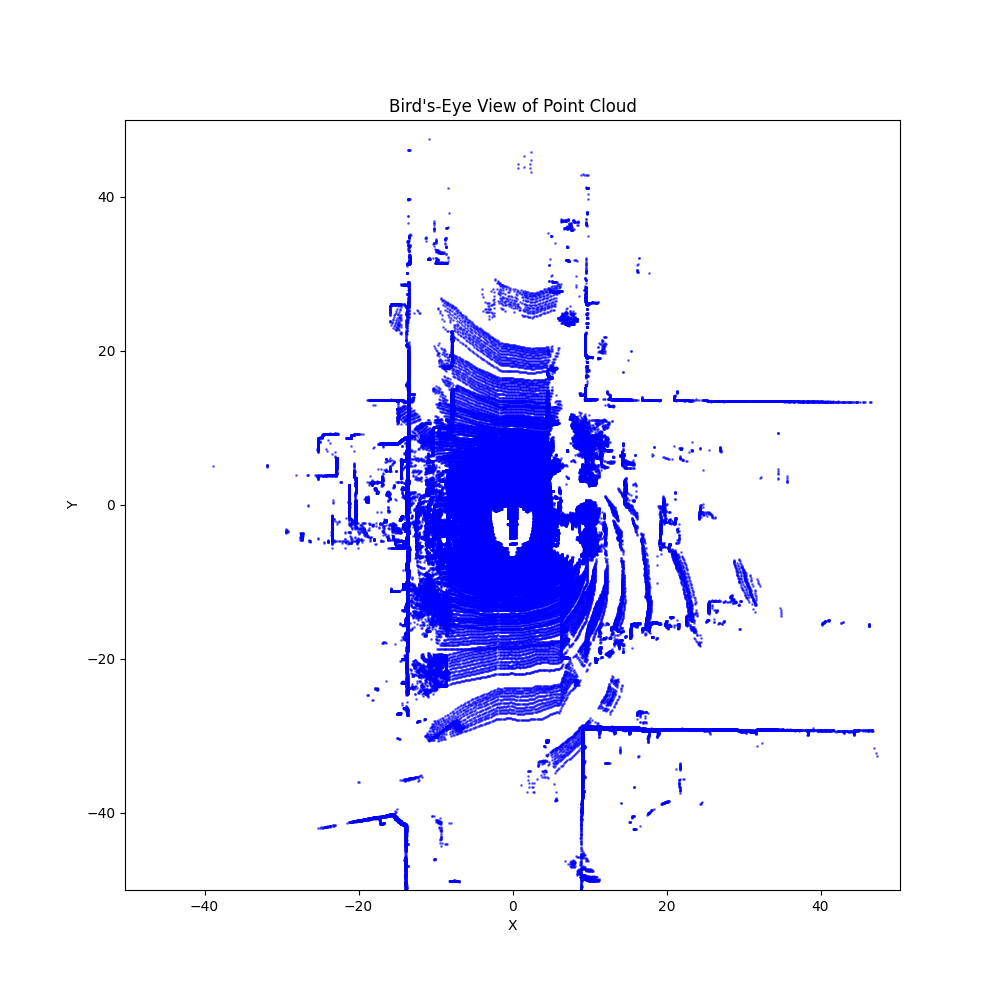}} &
        \multicolumn{2}{c}{\includegraphics[height=\imgh,trim=175 220 175 220,clip]{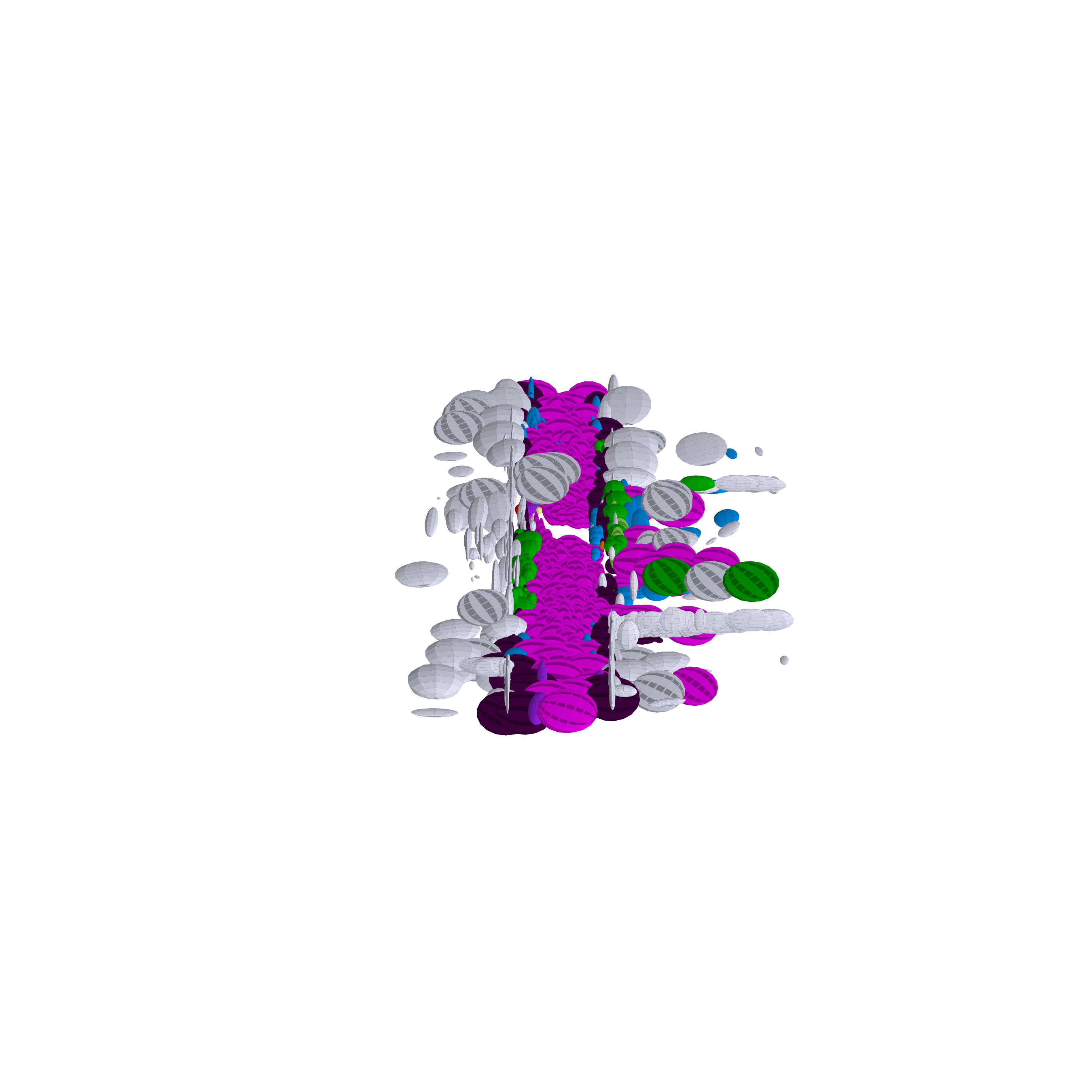}} &
        \multicolumn{2}{c}{\includegraphics[height=\imgh,trim=175 220 175 220,clip]{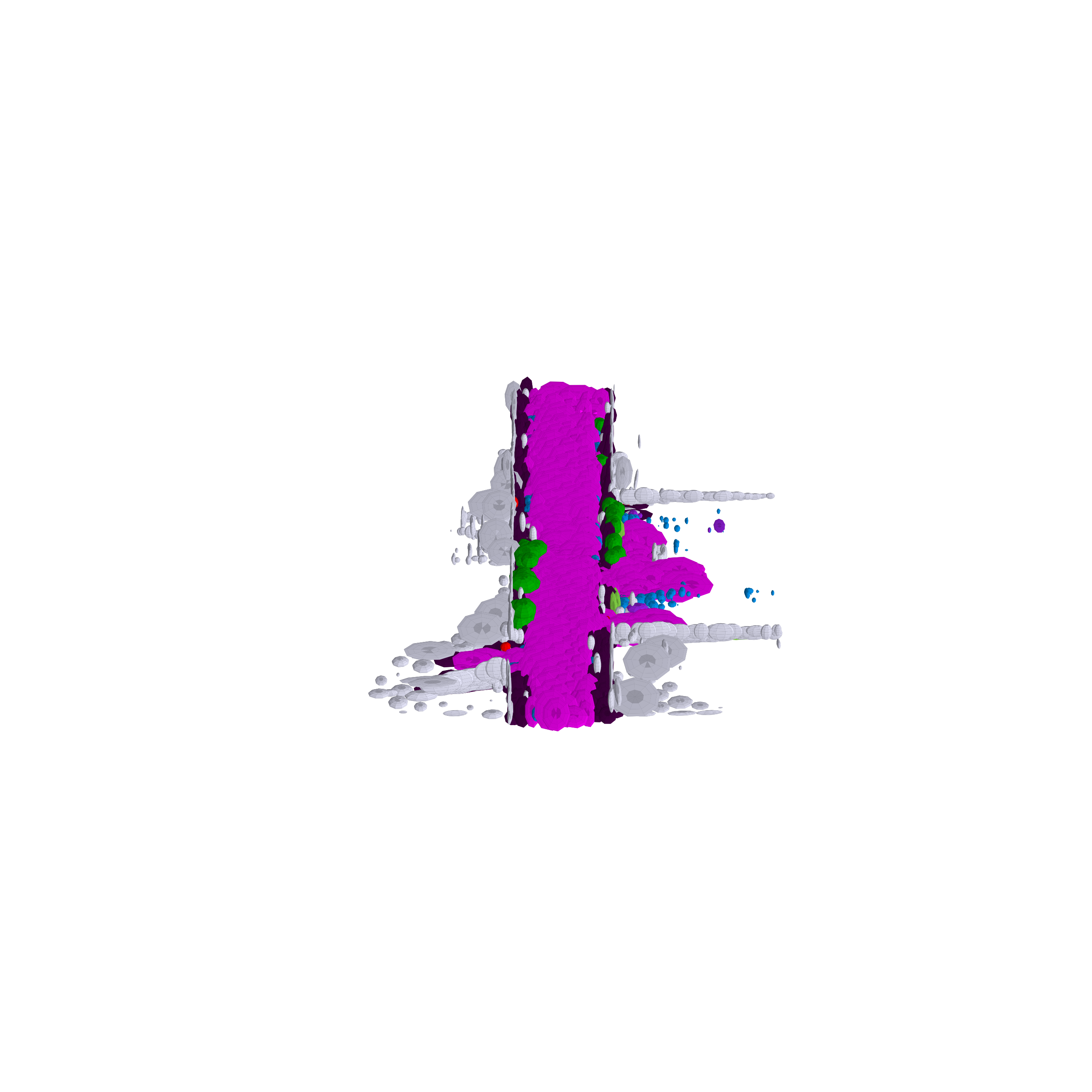}} \\
        \multicolumn{2}{c}{\footnotesize BEV LiDAR} &
        \multicolumn{2}{c}{\footnotesize Pred. Gaussians (C)} &
        \multicolumn{2}{c}{\footnotesize Pred. Gaussians (C+L+R)} \\
        \multicolumn{2}{c}{\includegraphics[width=0.33\linewidth]{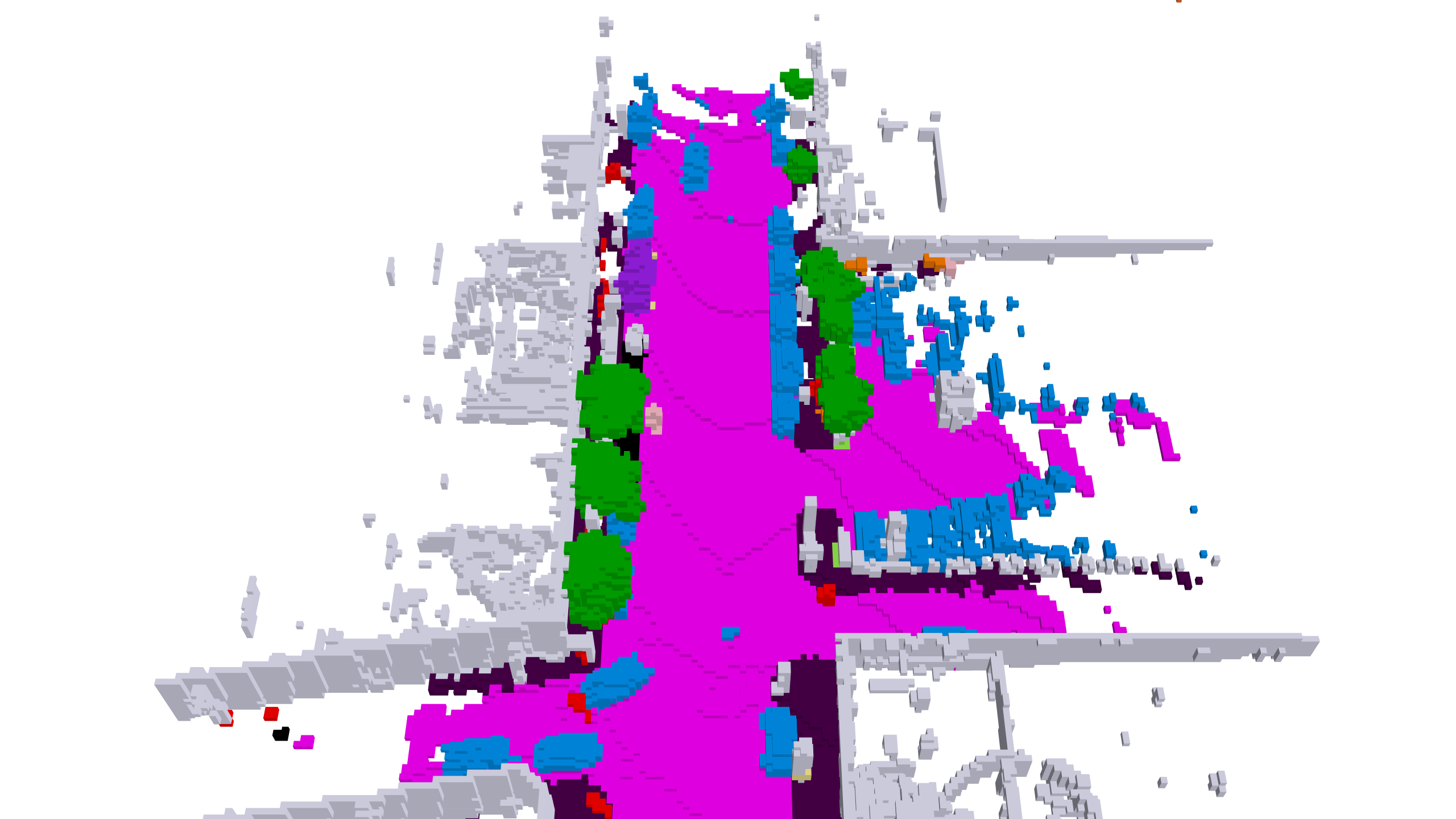}} &
        \multicolumn{2}{c}{\includegraphics[width=0.33\linewidth]{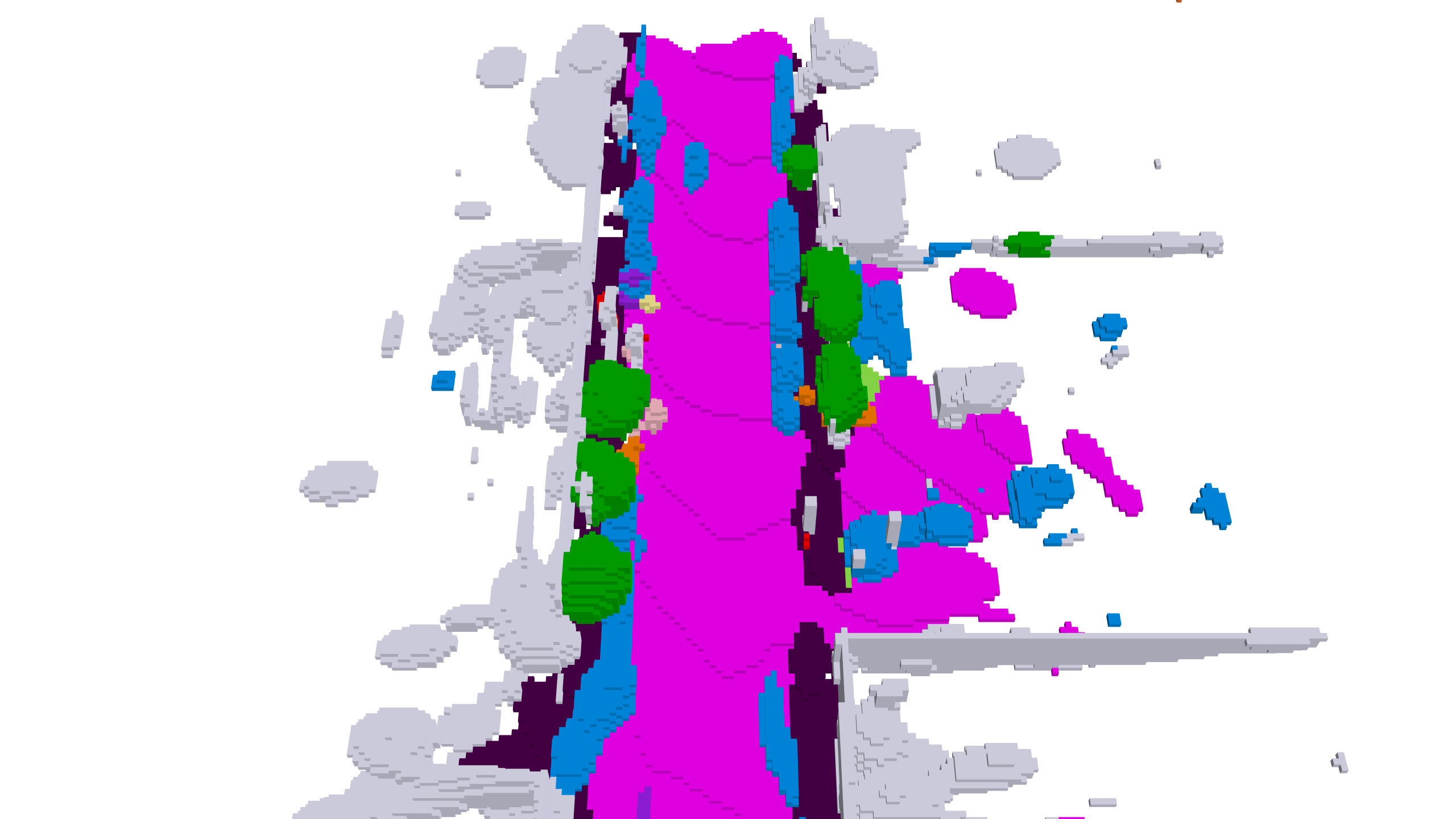}} &
        \multicolumn{2}{c}{\includegraphics[width=0.33\linewidth]{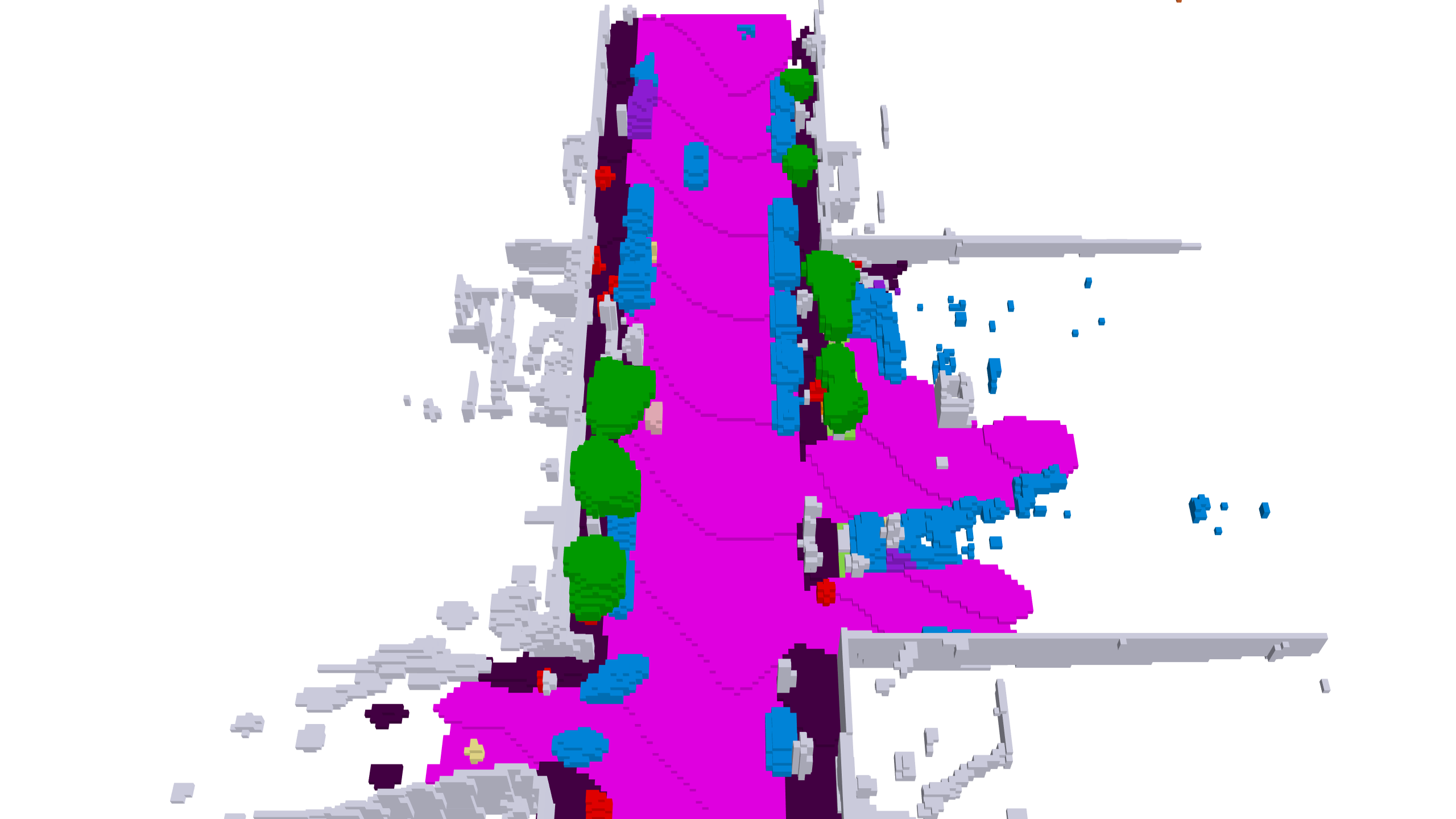}} \\ 
        \multicolumn{2}{c}{\footnotesize Ground Truth} &
        \multicolumn{2}{c}{\footnotesize Pred. Occupancy (C)} &
        \multicolumn{2}{c}{\footnotesize Pred. Occupancy (C+L+R)} \\

    \end{tabular}
    \caption{\textbf{Occupancy and Gaussian representation visualizations on nuScenes dataset.} Visualizations demonstrate the importance of additional sensors for the allocation of the Gaussians and final occupancy prediction.}
    \label{fig:occupancy_visualization_all}
\end{figure*}

The visualizations in Figure \ref{fig:occupancy_visualization_night} and Figure \ref{fig:occupancy_visualization_all} demonstrate the importance of sensor fusion for effective allocation of the semantic Gaussians, especially in low-light scenarios.

\section{Ablation study} \label{sec:ablation}

We conduct ablation studies to analyze the impact of key design choices and components of our GaussianFusionOcc framework. The latency and memory are tested on an NVIDIA RTX A6000 GPU for all the experiments.

\begin{table*}[htb!]
\centering
\caption{Ablation on number of Gaussians. The increased number of Gaussians improves the performance, without significantly increasing memory usage and the number of parameters, but it increases the latency.}
\resizebox{\textwidth}{!}{%
\begin{tabular}{l|c|c|c|c|c|c|c|c}
\toprule
Method & Modality & IoU & mIoU & Gaussians & Params & Memory (GB) & Latency (ms) \\
\midrule
GaussianFusionOcc & C+L & 45.16 & 30.21 & 6400 & 79.63M & 2.61 & 460 \\
GaussianFusionOcc & C+L & 45.74 & 30.83 & 25600 & 80M & 2.62 & 547 \\
\bottomrule
\end{tabular}%
}
\label{tab:nuscenes_val_gaussians}
\end{table*}

\begin{table*}[htb!]
\centering
\caption{Ablation on number of channels for feature representation. The increased number of channels improves the performance, but significantly increases the number of parameters and memory usage.}
\resizebox{\textwidth}{!}{%
\begin{tabular}{l|c|c|c|c|c|c|c|c}
\toprule
Method & Modality & IoU & mIoU & Channels & Params & Memory (GB) & Latency (ms) \\
\midrule
GaussianFusionOcc & C+L+R & 45.20 & 30.37 & 128 & 79.96M & 2.90 & 480 \\
GaussianFusionOcc & C+L+R & 45.69 & 30.85 & 192 & 115M & 5.41 & 486 \\
\bottomrule
\end{tabular}%
}
\label{tab:nuscenes_val_channels}
\end{table*}

\begin{table*}[htb!]
\centering
\caption{Ablation on initialization strategy. Probabilistic initialization is taken from \cite{huang2024probabilistic}.}
\resizebox{\textwidth}{!}{%
\begin{tabular}{l|c|c|c|c|c|c|c|c}
\toprule
Method & Modality & IoU & mIoU & Initinalization & Params & Memory (GB) & Latency (ms) \\
\midrule
GaussianFusionOcc & C+L+R & 45.31 & 30.07 & Random & 79.78M & 2.62 & 468 \\
GaussianFusionOcc & C+L+R & 45.20 & 30.37 & Learnable & 79.96M & 2.90 & 480 \\
GaussianFusionOcc & C+L+R & 44.52 & 30.13 & Probabilistic & 79.85M & 3.05 & 844 \\
\bottomrule
\end{tabular}%
}
\label{tab:nuscenes_val_init}
\end{table*}

\textbf{Number of Gaussians:} Table \ref{tab:nuscenes_val_gaussians} shows the influence of the number of Gaussians on efficiency and performance. We observe improvement in performance as the number of Gaussians increases. This is due to the enhanced ability to represent finer details with more Gaussians. The number of parameters and memory usage are not significantly increased because they are mostly influenced by sensor-specific encoders, which are not influenced by the number of Gaussians.

\textbf{Number of channels:} The influence of the number of channels used for extracted sensor features and per-Gaussian features is demonstrated in Table \ref{tab:nuscenes_val_channels}. An Increased number of channels improves the prediction performance with a significant efficiency degradation. The increase in memory and parameter number can be attributed to the increased size of sensor feature extractors, as they are also influenced by the number of channels.

\textbf{Initialization strategy:} We report the influence of initialization strategy on the performance and efficiency in Table \ref{tab:nuscenes_val_init}. Learnable initialization shows the highest mIoU with a slightly higher number of parameters, memory usage, and latency, compared to random initialization. Probabilistic initialization, proposed by GaussianFormer-2 \citep{huang2024probabilistic}, degrades the performance of the model while significantly slowing down the inference.

\subsection{Evaluation metrics}
To evaluate our method and compare the results with other state-of-the-art methods, we use Intersection over Union (IoU) and mean Intersection over Union (mIoU) metrics:
\begin{equation} \label{eq:iou}
    IoU = \frac{TP_{\neg c_0}}{TP_{\neg c_0} + FP_{\neg c_0} + FN_{\neg c_0}}
\end{equation}
\begin{equation} \label{eq:miou}
    mIoU = \frac{1}{|C|} \sum_{c \in C} \frac{TP_c}{TP_c + FP_c + FN_c}
\end{equation}
where TP, FP, FN denote the number of true positive, false positive, and false negative predictions, and  C, $c_0$ denote the set of classes without the empty class, and the empty class, respectively.

\end{document}